\pdfoutput=1

\documentclass[11pt]{article}

\usepackage[final]{acl}

\usepackage{times}
\usepackage{latexsym}
\usepackage{booktabs}
\usepackage{multirow}
\usepackage{float}
\usepackage{placeins}
\usepackage{booktabs}
\usepackage{microtype}
\usepackage[utf8]{inputenc}
\usepackage{textgreek}
\usepackage{wrapfig}
\usepackage{amsmath}
\usepackage{soul}
\usepackage{hyperref}

\usepackage{longtable}
\usepackage{xcolor}
\usepackage{tabularx}

\usepackage{pdfpages}

\usepackage{booktabs}
\usepackage{multirow}
\usepackage{makecell}
\usepackage[table]{xcolor}
\usepackage{graphicx}

\usepackage{caption}
\usepackage{float}
\usepackage[table]{xcolor}
\usepackage{graphicx,xcolor,colortbl,booktabs,multirow,makecell,tabularx}
\definecolor{BrickRed}{rgb}{0.8,0.25,0.33}
\definecolor{RoyalBlue}{rgb}{0.25,0.41,0.88}
\definecolor{ForestGreen}{rgb}{0.13,0.55,0.13}
\definecolor{Goldenrod}{rgb}{0.85,0.65,0.13}
\definecolor{OliveGreen}{rgb}{0.33,0.42,0.18}
\definecolor{Teal}{rgb}{0.0,0.5,0.5}
\definecolor{Purple}{rgb}{0.5,0.0,0.5}
\definecolor{LightGreen}{rgb}{0.93,0.98,0.93}
\definecolor{DeepGreen}{rgb}{0.1,0.39,0.1}
\definecolor{LightRed}{rgb}{0.99,0.93,0.93}
\definecolor{Red}{rgb}{1.0,0.0,0.0}
\definecolor{DeepRed}{rgb}{0.68,0.1,0.1}
\colorlet{posG}{DeepGreen!60!LightGreen}
\colorlet{negR}{DeepRed!60!LightRed}
\definecolor{DodgerBlue}{rgb}{0.12,0.56,1.0}

\usepackage{array}
\usepackage{pdflscape}

\usepackage{url}       

\usepackage[T1]{fontenc}

\usepackage[utf8]{inputenc}

\usepackage{microtype}

\definecolor{thedarkblue}{RGB}{0,0,120}
\definecolor{mydarkblue}{rgb}{0,0.08,0.45} 

\usepackage{hyperref}
\hypersetup{
    colorlinks=true,
    linkcolor=mydarkblue,
    citecolor=mydarkblue,
    filecolor=mydarkblue,
    urlcolor=mydarkblue}

\newcommand{\supertiny}{\fontsize{5pt}{6pt}\selectfont}

\newcommand{\score}[2]{#1 \textcolor[HTML]{8C8C8C}{\normalfont \scriptsize\supertiny(#2)}}

\usepackage[most]{tcolorbox}
\usepackage{fontawesome5} 
\usepackage{caption} 

\usepackage{graphicx}

\newcommand{\methead}[1]{\textbf{#1}}
\renewcommand{\arraystretch}{1.0}

\newtcolorbox{promptbox}[2][]{
  enhanced,
  breakable,
  colback=white!98!blue!2, 
  colframe=blue!70!black, 
  coltitle=white,         
  fonttitle=\bfseries\sffamily,
  title={\faTasks[regular]\hspace{1mm}~\parbox[t]{0.8\linewidth}{#2}},
  left=2mm, right=0mm, top=1mm, bottom=1mm,
  arc=3mm,               
  attach boxed title to top left={xshift=3mm, yshift*=-2mm},
  boxed title style={
    colback=blue!80!black,
    size=small,
    sharp corners=south,
    bottom=0.5mm, top=0.5mm,
    left=1mm, right=1mm,
    fontupper=\bfseries\sffamily,
    boxrule=0pt,
    drop shadow
  },
  boxrule=0.9pt,
  drop shadow southeast,
}

\usepackage{booktabs}     
\usepackage{multirow}     
\usepackage{makecell}     
\usepackage{array}        
\usepackage[table]{xcolor} 
\usepackage{graphicx}      
\definecolor{GoogleRed}{HTML}{EA4335}
\definecolor{GoogleBlue}{HTML}{4285F4}
\definecolor{GoogleGreen}{HTML}{34A853}

\usepackage{xcolor}
\definecolor{thedarkblue}{RGB}{0,0,120} 
\definecolor{mydarkblue}{rgb}{0,0.08,0.45} 
\definecolor{darkblue}{rgb}{0,0.08,180}
\colorlet{TufteRed}{red!80!black}
\definecolor{theblue}{RGB}{0,0,180}
\colorlet{thered}{TufteRed}

\usepackage{microtype}
\usepackage{balance}

\usepackage{booktabs}
\usepackage{tabularx}

\usepackage{amsmath,amssymb,amsthm}

\newcommand{\eat}[1]{\ignorespaces}
\usepackage{comment}

\usepackage{tikz}
\usepackage{verbatim}
\usetikzlibrary{arrows}
\usetikzlibrary{shapes,snakes}
\usetikzlibrary{decorations.pathmorphing} 
\usetikzlibrary{fit}					
\usetikzlibrary{backgrounds}	

\usepackage{ragged2e}
\usepackage{multirow}
\usepackage{microtype}
\usepackage{balance}
\usepackage{setspace}

\graphicspath{{./}{./graphics/}}
\newcolumntype{H}{>{\setbox0=\hbox\bgroup}c<{\egroup}@{}}

\newcolumntype{R}[1]{>{\RaggedLeft\arraybackslash}} 
\newcolumntype{L}[1]{>{\RaggedRight\arraybackslash}} 

\AtBeginEnvironment{pmatrix}{\setlength{\arraycolsep}{2pt}}

\usepackage{booktabs,tabularx,array,multirow,xcolor,hyperref}
\newcolumntype{Y}{>{\raggedright\arraybackslash}X}
\definecolor{preserve}{HTML}{2E7D32}   
\definecolor{editq}{HTML}{1565C0}      
\definecolor{fidelity}{HTML}{C62828}   

\usepackage{booktabs}
\usepackage{xcolor}
\definecolor{ForestGreen}{RGB}{34,139,34}
\definecolor{RoyalBlue}{RGB}{65,105,225}
\definecolor{BrickRed}{RGB}{178,34,34}

\DeclareMathOperator{\hugeE}{\mbox{\huge\raise-0.3ex\hbox{E}}}
\DeclareMathOperator{\p}{\mathbb{P}}
\DeclareMathOperator{\hugep}{\mbox{\huge\raise-0.3ex\hbox{$\p$}}}

\usepackage{subfigure}
\usepackage{nicefrac}

\DeclareMathAlphabet{\mathbcal}{OMS}{cmsy}{b}{n}
\usepackage{amsmath}
\usepackage{mathrsfs}
\usepackage{comment}

\usepackage{bm}
\usepackage{bbm}
\usepackage{amssymb}
\usepackage{paralist}
\usepackage[dvipsnames]{xcolor}

\definecolor{googleblue}{HTML}{4285F4}
\definecolor{googlered}{HTML}{DB4437}
\definecolor{googlepurple}{HTML}{A142F4} 
\definecolor{googlegreen}{HTML}{0F9D58}

\usepackage{tabu}
\usepackage[table]{xcolor}
\usepackage{enumitem}
\usepackage{colortbl}
\usepackage{booktabs}
\usepackage{tabularx}
\usepackage{ragged2e}

\title{Human-Aligned MLLM Judges for Fine-Grained Image Editing Evaluation: A Benchmark, Framework, and Analysis}

\author{
\small
Runzhou Liu$^{1}$ \quad Hailey Weingord$^{2}$ \quad Sejal Mittal$^{2}$ \quad Prakhar Dungarwal$^{2}$ \quad Anusha Nandula$^{2}$ \\ [-4pt]
\bfseries
\small
Bo Ni$^{3}$ \quad Samyadeep Basu$^{4}$ \quad Hongjie Chen$^{5}$ \quad Nesreen K.\ Ahmed$^{6}$ \quad Li Li$^{7}$ \quad Jiayi Zhang$^{8}$ \\ [-4pt]
\bfseries
\small
Koustava Goswami$^{4}$ \quad Subhojyoti Mukherjee$^{4}$ \quad Branislav Kveton$^{4}$ \quad Puneet Mathur$^{4}$ \\ [-4pt]
\bfseries
\small
Franck Dernoncourt$^{4}$ 
\bfseries
\small
\quad Yue Zhao$^{7}$ \quad Yu Wang$^{9}$ \qquad Ryan A.\ Rossi$^{4}$ \quad Zhengzhong Tu$^{10}$ \quad Hongru Du$^{1}$ \\
\\ [-4pt]
\small
$^{1}$University of Virginia \quad
$^{2}$Columbia University \quad
$^{3}$Vanderbilt University \quad
$^{4}$Adobe Research \quad \\ [-4pt]
\small
$^{5}$Dolby Laboratories \quad 
$^{6}$Cisco Research \quad
$^{7}$University of Southern California \quad \\ [-4pt]
\small
$^{8}$University of Wisconsin-Madison \quad 
$^{9}$University of Oregon \quad
$^{10}$Texas A\&M University
}

\usepackage{graphicx,xcolor,colortbl,booktabs,multirow,makecell}

\definecolor{BrickRed}{rgb}{0.8,0.25,0.33}
\definecolor{RoyalBlue}{rgb}{0.25,0.41,0.88}
\definecolor{ForestGreen}{rgb}{0.13,0.55,0.13}
\definecolor{LightGreen}{rgb}{0.93,0.98,0.93}
\definecolor{DeepGreen}{rgb}{0.1,0.39,0.1}
\colorlet{posG}{DeepGreen!60!LightGreen}

\begin{document}
\maketitle

\begin{abstract}
Evaluating image editing models remains challenging due to the coarse granularity and limited interpretability of traditional metrics, which often fail to capture aspects important to human perception and intent. Such metrics frequently reward visually plausible outputs while overlooking controllability, edit localization, and faithfulness to user instructions.
In this work, we introduce a fine-grained Multimodal Large Language Model (MLLM)–as–a–Judge framework for image editing that decomposes common evaluation notions into twelve fine-grained interpretable factors spanning image preservation, edit quality, and instruction fidelity. Building on this formulation, we present a new human-validated benchmark that integrates human judgments, MLLM-based evaluations, model outputs, and traditional metrics across diverse image editing tasks.
Through extensive human studies, we show that the proposed MLLM judges align closely with human evaluations at a fine granularity, supporting their use as reliable and scalable evaluators. We further demonstrate that traditional image editing metrics are often poor proxies for these factors, failing to distinguish over-edited or semantically imprecise outputs, whereas our judges provide more intuitive and informative assessments in both offline and online settings.
Together, this work introduces a benchmark, a principled factorization, and empirical evidence positioning fine-grained MLLM judges as a practical foundation for studying, comparing, and improving image editing approaches.
\end{abstract}

\section{Introduction}

\begin{figure*}[t]
\centering
\includegraphics[width=\textwidth]{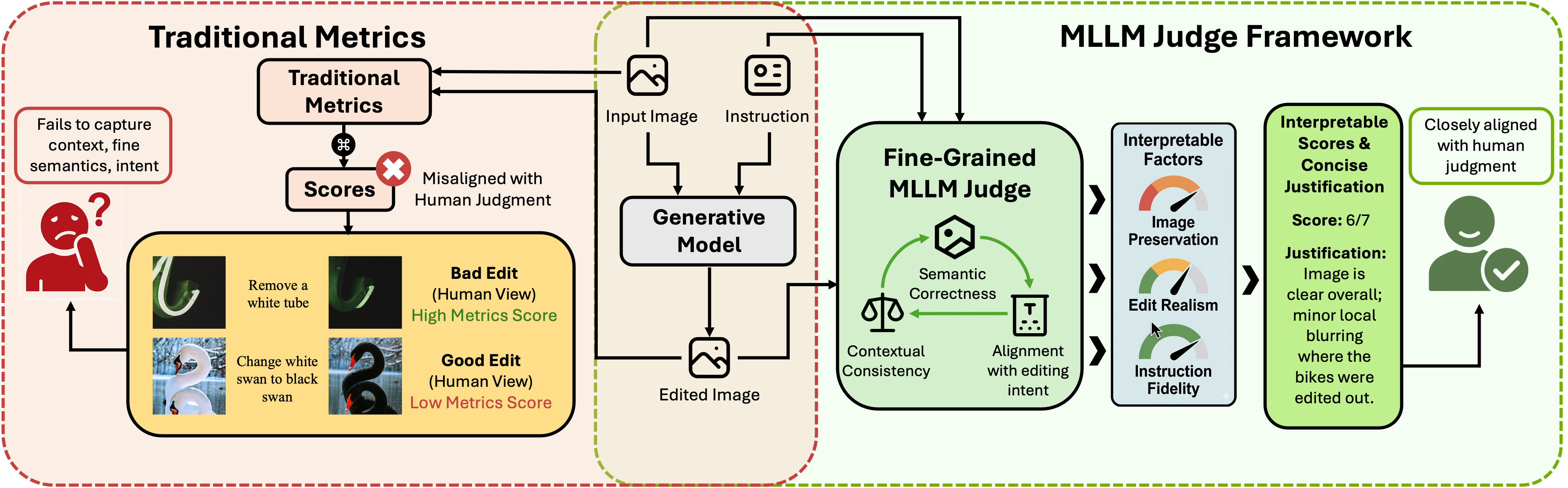}
\vspace{-0.1in}
\caption{
Motivation for fine-grained MLLM-based evaluation. The same image-editing example is assessed using two approaches: (left) traditional metrics, which collapse diverse editing behaviors into a single, potentially misleading score, and (right) our MLLM judge, which decomposes the edit into interpretable factors that explicitly explain why the edit succeeds or fails. This decomposition makes the motivation, methodology, and benefits of our approach immediately apparent.
}
\label{fig:figure_1}
\vspace{-0.4cm}
\end{figure*}

The rapid advancement of generative image editing models, with a highly impactful use-case being the ability to edit images via natural language prompts~\citep{hertz2022prompttoprompt, kawar2022imagic, mokady2022nulltext, zhang2022sine, ruiz2022dreambooth, shi2023instantbooth, couairon2022diffedit, brooks2023instructpix2pix}, has introduced a critical challenge: how to reliably evaluate the quality of these edits. The traditional metrics relied on to assess image edit quality are fundamentally misaligned with human judgment in real-world applications. Addressing this problem requires a new evaluation system that captures semantic reasoning, contextual understanding, and alignment with human intent, rather than relying solely on pixel-level similarity.

Existing metrics such as Peak Signal-to-Noise Ratio (PSNR)~\cite{jain1989digitalimage}, Structural Similarity Index Measure (SSIM)~\cite{wang2004ssim}, and Learned Perceptual Image Patch Similarity (LPIPS)~\cite{zhang2018lpips}usually focus on only one aspect of edit quality, such as low-level fidelity. Commonly used CLIP-based scores~\cite{hessel2021clipscore} are also shown to be not reliable~\cite{goel2022cyclip}. They often fail to capture other important factors that humans care about, including context consistency, fine-grained semantics, and alignment with editing intent. As a result, these metrics may assign high scores to edits that appear unsatisfactory to human observers, or assign low scores to edits that appear satisfactory to human observers like the case presented in Figure~\ref{fig:figure_1}. Refer to Appendix \ref{app: prob motivation} for more detailed examples of traditional metrics failing.

Moreover, most traditional metrics depend on the availability of a ground-truth (GT) image, an idealized reference for comparison. In real-world scenarios, this type of reference rarely exists. Users typically provide only an instruction and an input image, expecting the model to generate a plausible output. As a result, conventional evaluation frameworks are poorly suited to these \emph{online} (no-GT) settings, where performance must be assessed without comparison to a known correct image.

These limitations highlight the need for an evaluation framework that captures higher-level edit quality beyond pixel-level differences. Prior LLMs/MLLMs as judges approaches~\citep{fu2023judge, li2023judging, hsu2023gpt, prometheus2024} show promise for subjective evaluation but are not tailored to the unique demands of image editing. Existing MLLM-based judges for image editing approaches remain limited in that they either lack comprehensive multimodal reasoning for interpretable image edit assessment, being general on human judgments, or do not establish a unified scoring framework human evaluation and model-based judges~\citep{basu2023editval,pan2025icebench,yosef2025editinspector}.

To address these gaps, we propose a fine-grained Multimodal LLM-as-a-Judge (MLLM Judge) framework that evaluates instruction-guided image edits along interpretable semantic factors, including contextual consistency, semantic correctness, image preservation, edit realism, and instruction fidelity. Our framework supports both offline (with ground truth) and online (no ground truth) settings, and explicitly enables a three-way comparison among human evaluation, MLLM-as-a-Judge, and traditional metrics. By jointly analyzing these perspectives, the MLLM Judge more closely reflects human assessment of image editing quality and delivers evaluations better aligned with human judgment than existing approaches.

Furthermore, we release a high-quality benchmark for others to leverage for fine-tuning, and other important use cases. Beyond its role as an evaluation framework, the benchmark and its 13 fine grained factors serve as a principled curriculum for improving MLLMs. By releasing fine grained, human annotated supervision, the benchmark provides the signal required to teach models why an edit is correct or incorrect, rather than merely what the final output should resemble. This structure naturally supports supervised fine tuning toward reasoning judges that produce chain of thought style critiques, grounding each score in explicit, factor level logic. In turn, the factorized design gives rise to factor aware RLHF, where dense, semantically meaningful rewards encourage a calibrated trade off between instruction fidelity and image preservation. Finally, the benchmark enables systematic hard negative mining and curriculum learning, allowing practitioners to surface concrete failure modes such as scale realism or spatial inconsistency and to train models along a controlled progression of increasingly complex editing behaviors.

\begin{table*}[t]
\centering
\caption{
\textbf{Human and our MLLM-as-a-Judge scores for all factors and across all edit types.}
We report the average score over all image edit types in the last column and over all factors in the last row.  When the difference between our judge score and human evaluation is closer than $0.5$, its background is dark green. When the difference is closer than $1.0$, its background is light green. The gray text is the standard deviation of the scores from which the average is computed.
}
\label{tab:human_judge_v1}
\tiny
\setlength{\tabcolsep}{3pt}
\renewcommand{\arraystretch}{1.1}
\begin{tabular}{@{}c c l *{6}{c} c@{}}
\toprule
& & &
\multicolumn{6}{c}{\sc Image Edit Types} \\
\cmidrule{4-9}
&
\textbf{Factor} &
&
\makecell{\textbf{Add}} &
\makecell{\textbf{Remove}} &
\makecell{\textbf{Replace}} &
\makecell{\textbf{Action}} &
\makecell{\textbf{Counting}} &
\makecell{\textbf{Relation}} &
\textbf{All Edits} \\
\midrule
\multirow{6}{*}{\rotatebox[origin=c]{90}{\textcolor{GoogleGreen}{\textsc{\sc \bfseries Image Preserv.}}}} & \multirow{2}{*}{\centering\arraybackslash \textcolor{GoogleGreen}{\textbf{Unchanged Regions}}} & Human & \score{5.172}{0.82} & \score{5.731}{0.75} & \score{4.972}{1.02} & \score{4.352}{1.06} & \score{3.393}{0.68} & \score{4.992}{0.43} & \score{4.769}{0.74} \\
 &  & Our Judge & \score{6.444}{0.50} & \cellcolor{green!30}\textbf{\score{5.824}{0.71}} & \score{6.222}{0.63} & \score{5.826}{0.82} & \score{4.400}{1.50} & \cellcolor{green!20}\textbf{\score{5.833}{0.90}} & \cellcolor{green!20}\textbf{\score{5.758}{0.65}} \\
\cmidrule(l){2-10}
 & \multirow{2}{*}{\centering\arraybackslash \textcolor{GoogleGreen}{\textbf{Global Consistency}}} & Human & \score{5.602}{0.70} & \score{5.982}{0.61} & \score{5.551}{0.90} & \score{4.769}{0.87} & \score{5.243}{1.13} & \score{5.444}{0.39} & \score{5.432}{0.37} \\
 &  & Our Judge & \cellcolor{green!20}\textbf{\score{6.333}{0.47}} & \cellcolor{green!30}\textbf{\score{5.971}{0.82}} & \cellcolor{green!20}\textbf{\score{6.333}{0.47}} & \score{6.087}{0.65} & \cellcolor{green!30}\textbf{\score{4.800}{1.17}} & \cellcolor{green!20}\textbf{\score{6.167}{0.69}} & \cellcolor{green!20}\textbf{\score{5.948}{0.53}} \\
\cmidrule(l){2-10}
 & \multirow{2}{*}{\centering\arraybackslash \textcolor{GoogleGreen}{\textbf{Identity Preservation}}} & Human & \score{5.613}{0.62} & \score{5.913}{0.82} & \score{5.625}{0.84} & \score{4.871}{1.12} & \score{4.227}{1.12} & \score{5.714}{0.20} & \score{5.327}{0.59} \\
 &  & Our Judge & \score{6.889}{0.31} & \cellcolor{green!30}\textbf{\score{6.118}{1.02}} & \cellcolor{green!20}\textbf{\score{6.500}{0.96}} & \score{6.696}{0.46} & \score{6.400}{0.49} & \cellcolor{green!20}\textbf{\score{6.500}{0.50}} & \score{6.517}{0.24} \\
\midrule
\multirow{12}{*}{\rotatebox[origin=c]{90}{\textcolor{GoogleBlue}{\textsc{\sc \bfseries Edit Quality}}}} & \multirow{2}{*}{\centering\arraybackslash \textcolor{GoogleBlue}{\textbf{Scale Realism}}} & Human & \score{5.276}{0.95} & \score{6.286}{0.54} & \score{5.865}{0.80} & \score{5.984}{0.61} & \score{5.510}{0.68} & \score{6.033}{0.57} & \score{5.826}{0.34} \\
 &  & Our Judge & \score{6.444}{0.50} & \cellcolor{green!30}\textbf{\score{6.471}{0.74}} & \cellcolor{green!20}\textbf{\score{6.556}{0.60}} & \cellcolor{green!20}\textbf{\score{6.565}{0.92}} & \cellcolor{green!20}\textbf{\score{6.400}{0.49}} & \cellcolor{green!30}\textbf{\score{6.000}{1.41}} & \cellcolor{green!20}\textbf{\score{6.406}{0.19}} \\
\cmidrule(l){2-10}
 & \multirow{2}{*}{\centering\arraybackslash \textcolor{GoogleBlue}{\textbf{Spatial Relationship}}} & Human & \score{5.561}{0.79} & \score{6.225}{0.50} & \score{5.890}{0.63} & \score{5.948}{0.74} & \score{4.650}{1.19} & \score{5.728}{0.63} & \score{5.667}{0.50} \\
 &  & Our Judge & \cellcolor{green!20}\textbf{\score{6.556}{0.68}} & \cellcolor{green!30}\textbf{\score{6.382}{0.84}} & \cellcolor{green!20}\textbf{\score{6.778}{0.63}} & \cellcolor{green!30}\textbf{\score{6.043}{0.75}} & \cellcolor{green!30}\textbf{\score{4.800}{1.17}} & \cellcolor{green!20}\textbf{\score{6.667}{0.47}} & \cellcolor{green!20}\textbf{\score{6.204}{0.67}} \\
\cmidrule(l){2-10}
 & \multirow{2}{*}{\centering\arraybackslash \textcolor{GoogleBlue}{\textbf{Texture and Detail}}} & Human & \score{5.504}{0.56} & \score{5.844}{0.61} & \score{5.483}{0.80} & \score{5.643}{0.84} & \score{5.157}{0.90} & \score{5.639}{0.59} & \score{5.545}{0.21} \\
 &  & Our Judge & \cellcolor{green!30}\textbf{\score{5.889}{0.31}} & \cellcolor{green!30}\textbf{\score{5.676}{0.72}} & \cellcolor{green!20}\textbf{\score{6.000}{0.47}} & \cellcolor{green!30}\textbf{\score{6.000}{0.66}} & \cellcolor{green!20}\textbf{\score{5.800}{0.40}} & \cellcolor{green!30}\textbf{\score{5.667}{0.75}} & \cellcolor{green!30}\textbf{\score{5.839}{0.14}} \\
\cmidrule(l){2-10}
 & \multirow{2}{*}{\centering\arraybackslash \textcolor{GoogleBlue}{\textbf{Image Quality}}} & Human & \score{5.569}{0.54} & \score{6.048}{0.66} & \score{5.513}{0.71} & \score{5.947}{0.71} & \score{5.247}{0.53} & \score{5.683}{0.75} & \score{5.668}{0.27} \\
 &  & Our Judge & \cellcolor{green!20}\textbf{\score{6.333}{0.47}} & \cellcolor{green!30}\textbf{\score{6.059}{0.59}} & \score{6.556}{0.50} & \cellcolor{green!30}\textbf{\score{6.174}{0.48}} & \cellcolor{green!20}\textbf{\score{6.200}{0.40}} & \cellcolor{green!20}\textbf{\score{6.333}{0.75}} & \cellcolor{green!20}\textbf{\score{6.276}{0.16}} \\
\cmidrule(l){2-10}
 & \multirow{2}{*}{\centering\arraybackslash \textcolor{GoogleBlue}{\textbf{Color and Lighting}}} & Human & \score{5.515}{0.75} & \score{5.855}{0.71} & \score{5.442}{0.86} & \score{5.549}{0.72} & \score{5.403}{0.83} & \score{5.553}{0.64} & \score{5.553}{0.15} \\
 &  & Our Judge & \cellcolor{green!20}\textbf{\score{6.111}{0.57}} & \cellcolor{green!30}\textbf{\score{5.941}{0.87}} & \cellcolor{green!20}\textbf{\score{6.278}{0.56}} & \cellcolor{green!30}\textbf{\score{5.870}{0.80}} & \cellcolor{green!20}\textbf{\score{6.200}{0.75}} & \cellcolor{green!20}\textbf{\score{6.167}{0.69}} & \cellcolor{green!20}\textbf{\score{6.094}{0.14}} \\
\cmidrule(l){2-10}
 & \multirow{2}{*}{\centering\arraybackslash \textcolor{GoogleBlue}{\textbf{Seamlessness}}} & Human & \score{5.722}{0.64} & \score{6.101}{0.58} & \score{5.767}{0.74} & \score{5.598}{0.83} & \score{5.357}{1.00} & \score{5.578}{0.73} & \score{5.687}{0.23} \\
 &  & Our Judge & \cellcolor{green!30}\textbf{\score{6.000}{0.47}} & \cellcolor{green!30}\textbf{\score{5.706}{0.89}} & \cellcolor{green!20}\textbf{\score{6.333}{0.58}} & \cellcolor{green!30}\textbf{\score{5.913}{0.83}} & \cellcolor{green!30}\textbf{\score{5.600}{0.80}} & \cellcolor{green!30}\textbf{\score{5.667}{0.75}} & \cellcolor{green!30}\textbf{\score{5.870}{0.25}} \\
\midrule
\multirow{6}{*}{\rotatebox[origin=c]{90}{\textcolor{GoogleRed}{\textsc{\sc \bfseries Instruct. Fidel.}}}} & \multirow{2}{*}{\centering\arraybackslash \textcolor{GoogleRed}{\textbf{Alignment}}} & Human & \score{5.556}{0.49} & \score{5.927}{0.97} & \score{5.681}{0.65} & \score{5.666}{1.13} & \score{3.437}{1.40} & \score{5.178}{0.99} & \score{5.241}{0.84} \\
 &  & Our Judge & \score{6.667}{0.94} & \cellcolor{green!20}\textbf{\score{6.471}{1.01}} & \cellcolor{green!20}\textbf{\score{6.500}{0.90}} & \score{6.957}{0.20} & \score{5.200}{2.23} & \cellcolor{green!20}\textbf{\score{6.167}{1.86}} & \score{6.327}{0.56} \\
\cmidrule(l){2-10}
 & \multirow{2}{*}{\centering\arraybackslash \textcolor{GoogleRed}{\textbf{Completeness}}} & Human & \score{5.693}{0.72} & \score{5.966}{1.17} & \score{5.789}{0.71} & \score{5.719}{1.14} & \score{3.537}{1.74} & \score{5.556}{0.66} & \score{5.376}{0.83} \\
 &  & Our Judge & \score{6.778}{0.63} & \cellcolor{green!30}\textbf{\score{6.294}{1.30}} & \cellcolor{green!20}\textbf{\score{6.389}{1.06}} & \score{6.870}{0.45} & \score{5.200}{2.23} & \cellcolor{green!20}\textbf{\score{6.167}{1.86}} & \cellcolor{green!20}\textbf{\score{6.283}{0.55}} \\
\cmidrule(l){2-10}
 & \multirow{2}{*}{\centering\arraybackslash \textcolor{GoogleRed}{\textbf{Plausibility}}} & Human & \score{5.209}{1.03} & \score{6.023}{0.78} & \score{5.743}{0.80} & \score{5.692}{1.17} & \score{4.917}{1.08} & \score{5.586}{0.89} & \score{5.528}{0.36} \\
 &  & Our Judge & \score{6.667}{0.47} & \cellcolor{green!20}\textbf{\score{6.529}{0.78}} & \score{6.889}{0.31} & \score{6.826}{0.64} & \score{6.800}{0.40} & \cellcolor{green!20}\textbf{\score{6.500}{0.76}} & \score{6.702}{0.15} \\
\midrule
  & \multirow{2}{*}{\textbf{Overall Average}} 
 & Human & \score{5.499}{0.17} & \score{5.992}{0.15} & \score{5.610}{0.24} & \score{5.478}{0.50} & \score{4.673}{0.78} & \score{5.557}{0.26} & \score{5.652}{0.47} \\
 &  & Our Judge & \cellcolor{green!20}\textbf{\score{6.426}{0.30}} & \cellcolor{green!30}\textbf{\score{6.120}{0.29}} & \cellcolor{green!20}\textbf{\score{6.444}{0.23}} & \cellcolor{green!20}\textbf{\score{6.319}{0.41}} & \cellcolor{green!20}\textbf{\score{5.650}{0.74}} & \cellcolor{green!20}\textbf{\score{6.153}{0.31}} & \cellcolor{green!20}\textbf{\score{6.236}{0.40}} \\
\bottomrule
\end{tabular}
\end{table*}

\medskip\noindent\textbf{Summary of Main Contributions.} 
The key contributions of this work are as follows:

\begin{compactitem}
    \item \textbf{Benchmark.}
    We introduce a new image-editing benchmark that combines human evaluations over fine-grained factors with scores from models and traditional metrics, with representative use cases discussed in Figure~\ref{fig:benchmark_flowchart} and Appendix~\ref{sec:appendix-benchmark}.
    
    \item \textbf{Novel MLLM Judge Factors for Image Editing.}     
    We propose an MLLM judge with 12 fine-grained factors spanning image preservation, edit quality, and instruction fidelity (Table~\ref{table:Factors}), enabling diagnostic evaluation in both offline and online settings.
    
    \item \textbf{Human Alignment of Proposed Image Editing Judges}.
    We show strong alignment between human judgments and our MLLM judges across all factors and edit types, indicating reliable, human-consistent evaluation as shown in Table~\ref{tab:human_judge_v1}.

    \item \textbf{Traditional vs. Our MLLM Image Editing Judges.}
    We demonstrate that traditional metrics correlate with only a limited subset of our judge factors, while MLLM judges as detailed in Section~\ref{sec:exp} provide more intuitive and reliable assessments of edit quality.
    
    \item \textbf{Extensive Experiments \& Findings.} We conduct comprehensive experiments and analyses, and release our code and data to support future research and practical adoption. The code and data are made available.\footnote{Our code and data: \href{https://github.com/mllmasajudge-anonymous/MLLM-as-a-Judge}
    {\nolinkurl{https://github.com/mllmasajudge-anonymous/MLLM-as-a-Judge}}}
\end{compactitem}

\begin{table*}[t!]
\vspace{4mm}
\centering
\setlength{\tabcolsep}{4pt}
\renewcommand{\arraystretch}{1.1}
\caption{Results comparing the overall scores from our MLLM judges to humans across the higher-order categories of \textcolor{googlegreen}{image preservation}, \textcolor{googleblue}{edit quality}, and \textcolor{googlered}{instruction fidelity}.
}
\vspace{-2mm}
\label{tab:results-summary}
\tiny
\setlength{\tabcolsep}{3pt}
\renewcommand{\arraystretch}{1.1}
\begin{tabular}{@{}c l *{6}{c} c@{}}
\toprule
& & \multicolumn{6}{c}{\sc Image Edit Types} \\
\cmidrule{3-8}
\textbf{Category} & &
\makecell{\textbf{Add}} &
\makecell{\textbf{Remove}} &
\makecell{\textbf{Replace}} &
\makecell{\textbf{Action}} &
\makecell{\textbf{Counting}} &
\makecell{\textbf{Relation}} &
\textbf{All Edits} \\
\midrule
\multirow{2}{*}{\rotatebox[origin=c]{0}{\textcolor{GoogleGreen}{\textsc{\sc \bfseries Image Preserv.}}}} & Human & \score{5.462}{0.71} & \score{5.875}{0.73} & \score{5.383}{0.92} & \score{4.664}{1.02} & \score{4.288}{0.98} & \score{5.383}{0.34} & \score{5.176}{0.57} \\
 & Our Judge & \score{6.555}{0.43} & \cellcolor{green!30}\textbf{\score{5.971}{0.85}} & \cellcolor{green!20}\textbf{\score{6.352}{0.69}} & \score{6.203}{0.64} & \cellcolor{green!20}\textbf{\score{5.200}{1.05}} & \cellcolor{green!20}\textbf{\score{6.167}{0.70}} & \cellcolor{green!20}\textbf{\score{6.074}{0.47}} \\
\midrule
\multirow{2}{*}{\rotatebox[origin=c]{0}{\textcolor{GoogleBlue}{\textsc{\sc \bfseries Edit Quality}}}} & Human & \score{5.524}{0.70} & \score{6.060}{0.60} & \score{5.660}{0.76} & \score{5.778}{0.74} & \score{5.221}{0.85} & \score{5.702}{0.65} & \score{5.658}{0.28} \\
 & Our Judge & \cellcolor{green!20}\textbf{\score{6.222}{0.50}} & \cellcolor{green!30}\textbf{\score{6.039}{0.77}} & \cellcolor{green!20}\textbf{\score{6.417}{0.56}} & \cellcolor{green!30}\textbf{\score{6.094}{0.74}} & \cellcolor{green!20}\textbf{\score{5.833}{0.67}} & \cellcolor{green!30}\textbf{\score{6.084}{0.80}} & \cellcolor{green!30}\textbf{\score{6.115}{0.26}} \\
\midrule
\multirow{2}{*}{\rotatebox[origin=c]{0}{\textcolor{GoogleRed}{\textsc{\sc \bfseries Instruct. Fidel.}}}} & Human & \score{5.486}{0.75} & \score{5.972}{0.97} & \score{5.738}{0.72} & \score{5.692}{1.15} & \score{3.964}{1.41} & \score{5.440}{0.85} & \score{5.382}{0.68} \\
 & Our Judge & \score{6.704}{0.68} & \cellcolor{green!30}\textbf{\score{6.431}{1.03}} & \cellcolor{green!20}\textbf{\score{6.593}{0.76}} & \score{6.884}{0.43} & \score{5.733}{1.62} & \cellcolor{green!20}\textbf{\score{6.278}{1.49}} & \score{6.437}{0.42} \\
\midrule
 \multirow{2}{*}{\textbf{Overall Average}} & Human & \score{5.499}{0.17} & \score{5.992}{0.15} & \score{5.610}{0.24} & \score{5.478}{0.50} & \score{4.673}{0.78} & \score{5.557}{0.26} & \score{5.652}{0.47} \\
   & Our Judge & \cellcolor{green!20}\textbf{\score{6.426}{0.30}} & \cellcolor{green!30}\textbf{\score{6.120}{0.29}} & \cellcolor{green!20}\textbf{\score{6.444}{0.23}} & \cellcolor{green!20}\textbf{\score{6.319}{0.41}} & \cellcolor{green!20}\textbf{\score{5.650}{0.74}} & \cellcolor{green!20}\textbf{\score{6.153}{0.31}} & \cellcolor{green!20}\textbf{\score{6.236}{0.40}} \\
\bottomrule
\end{tabular}
\end{table*}

\begin{figure}[t!]
\centering
\includegraphics[width=1.0\linewidth]{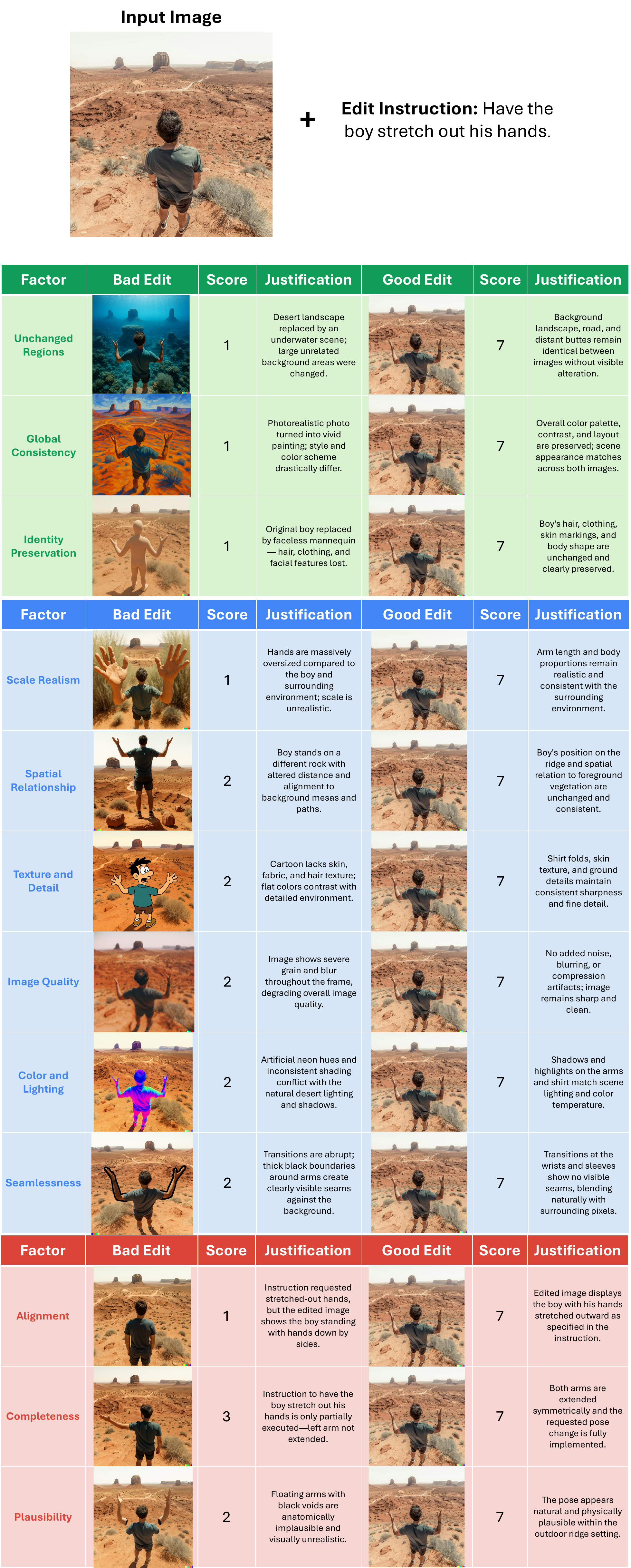}
\caption{Overview of the proposed factors used in our MLLM-as-a-Judge for image editing.
Results are shown for each factor using both poorly edited images and those that were edited well
(implementation in Fig.~\ref{fig:prompt-mllm-as-a-judge-image-editing, main}).
}
\label{fig:overview-factor}
\end{figure}

\begin{table*}[t!]
    \centering
    \scriptsize
    \setlength{\tabcolsep}{0.3em}
    \renewcommand{\arraystretch}{1.0}
    \caption{
    Our proposed taxonomy for image editing judges including higher-order categories of 
    \textcolor{googlegreen}{image preservation}, 
    \textcolor{googleblue}{edit quality}, 
    and \textcolor{googlered}{instruction fidelity}.
    We decompose these higher-order categories into finer-grained factors, including 3 factors for \textcolor{googlegreen}{image preservation}, 
    6 factors for \textcolor{googleblue}{edit quality}, 3 factors for \textcolor{googlered}{instruction fidelity}, and lastly, an \textbf{Overall} factor,
    \emph{totaling 13 factors}.
    For human evaluators, we include overall factor as a question, and for MLLM-as-a-Judge evaluators, the overall factor score is calculated from 12 other factors.
    }
    \begin{tabularx}{\textwidth}{p{30mm} p{30mm} l }
    \toprule
   \textbf{Category} 
   & \textbf{Factor}      & \textbf{Question} \\ 
    \midrule
   
    & \textcolor{googlegreen}{Unchanged Regions}  & Did the parts of the image that were not supposed to be edited remain unchanged? \\
   \textcolor{googlegreen}{\textsc{\bfseries \scshape Image Preservation}} 
    & \textcolor{googlegreen}{Global Consistency}   & Has the overall appearance (style, layout, and color) been preserved? \\
    & \textcolor{googlegreen}{Identity Preservation}  & Do people, animals, or objects maintain their original identity and features after the edit? \\
   \midrule
   
    & \textcolor{googleblue}{Scale Realism}  &  Is the scale of the edited object realistic compared to other objects in the image? \\
    \textcolor{googleblue}{{\bfseries \scshape Edit Quality}} 
    & \textcolor{googleblue}{Spatial Relationship}  &  Has the spatial relationship between objects been maintained?\\
    & \textcolor{googleblue}{Texture and Detail}  & Is the texture and detail in the edited region consistent with the surrounding areas? \\
    & \textcolor{googleblue}{Image Quality}  &  Does the edited image avoid noise, blurring, or unnatural distortions?\\
    & \textcolor{googleblue}{Color and Lighting}  & Do the colors, shadows, and lighting of the edited region match the rest of the image? \\
    & \textcolor{googleblue}{Seamlessness}  & Does the transition between edited and non-edited regions look natural? \\
    \midrule

    & \textcolor{googlered}{Alignment}  & Does the edited image align with the specific edits provided in the instructions? \\
   \textcolor{googlered}{{\bfseries \scshape Instruction Fidelity}} & \textcolor{googlered}{Completeness}   & Were all aspects of the instruction carried out fully? \\
    & \textcolor{googlered}{Plausibility}  & Does the result make sense in a real-world context? \\
       \midrule

    \textbf{\textcolor{black}{Overall}} 
    & \textcolor{black}{Overall Edit Quality} 
    & Considering all factors, how good is the edit overall? \\
    \bottomrule
    \end{tabularx}
    \label{table:Factors}
\end{table*}

\section{Related Work} \label{sec:related-work}

\subsection{Image Editing Methods}
Recent advances in text-guided image generation have enabled powerful multimodal models to synthesize images directly from natural language prompts\citep{rombach2022stablediffusion, ramesh2022dalle2, saharia2022imagen, balaji2023ediffi, ho2021cascadeddiffusion}, and many of these models have subsequently been adapted to support a wide range of image editing tasks~\citep{hertz2022prompttoprompt, kawar2022imagic, mokady2022nulltext, zhang2022sine, ruiz2022dreambooth, shi2023instantbooth, couairon2022diffedit, brooks2023instructpix2pix}. As image editing increasingly relies on language-conditioned generative models, evaluating edit quality has become a fundamentally semantic and intent-driven problem. In parallel, prior work has explored LLMs/MLLMs as judges for subjective evaluation in domains such as UI quality~\cite{mlm_uijudge2026}, chart comprehension~\cite{chart2experience2025}, text assessment~\citep{prometheus2024, fu2023judge}, and even MLLM performance assessment~\cite{li2023judging}, demonstrating their ability to capture human preferences beyond static similarity metrics. More recent efforts have also applied MLLM-based judges to image editing evaluation~\citep{fu2023judge, li2023judging, hsu2023gpt, prometheus2024}. 
However, existing MLLM-based judges are not tailored to the unique failure modes of image editing. They typically provide coarse, holistic scores, instead of explicitly decomposing image editing quality into well-defined, separable dimensions. These limitations motivate a fine-grained, human-aligned MLLM-based evaluation framework specialized for image editing.

\subsection{Traditional Metrics for Image Editing}
Following prior work~\citep{sun2023magicbrush, brooks2023instructpix2pix, humanedit2025}, we adopt a set of traditional metrics grouped into three categories, as summarized in Table~\ref{trad_met_def}. Pixel-level fidelity metrics, including L1/L2 distance, PSNR~\cite{jain1989digitalimage}, SSIM~\cite{wang2004ssim}, and LPIPS~\cite{zhang2018lpips}, measure reconstruction accuracy by comparing edited images to ground truth. Content preservation metrics, such as Mask-SSIM and Mask-LPIPS, evaluate similarity without masked regions, while background consistency~\cite{brooks2023instructpix2pix} assesses whether unedited areas remain unchanged. Semantic alignment metrics capture higher-level relevance, with CLIP Score~\cite{hessel2021clipscore} measuring text–image alignment in a reference-free manner and DINO similarity~\cite{caron2021emerging} comparing deep visual features against ground truth. 
Additional discussion of image editing paradigms and MLLM-based judges is provided in Appendix~\ref{app:related-work}.

\section{Fine-Grained MLLM Judges} \label{sec:our-approaches}
In this section, we introduce our novel framework of fine-grained MLLM judges for evaluating instruction-guided image editing. This framework formalizes human-aligned evaluation by decomposing edit quality into interpretable semantic dimensions. First, we formulate the problem and motivate the need for a fine-grained approach (Section~\ref{sec:our-approach-problem-formulation}), and then we detail our proposed set of MLLM judge factors (Section~\ref{sec:our-approach-judge-factors}).

\subsection{Problem Formulation} \label{sec:our-approach-problem-formulation}

We formalize the image-editing evaluation problem as deriving a scalar score $S(I_o, I_e, T)$ consisting of an \textit{original image} $I_{o}$, a corresponding \textit{edited image} $I_{e}$, and a natural language \textit{instruction} $T$. The score $S(I_o, I_e, T)$ must reflect a balance between two competing objectives: {adherence to the instruction and preservation of the original content.

While this objective can be captured by a single, general evaluation function, we argue that such formulations are insufficient for diagnosing image-editing behavior. Instead, our work introduces a \textbf{fine-grained MLLM judge}, defined as an evaluation function that decomposes $S$ into interpretable sub-scores, where each sub-score $s_i$ quantifies a distinct, separable aspect of edit quality. This decomposition is meaningful because image-editing failures are often nuanced: a high-fidelity edit can fail on subtle issues like lighting consistency, and traditional single-score metrics cannot distinguish these failure modes. By leveraging an MLLM's reasoning capabilities, we can automatically produce these sub-scores, providing both an overall quality assessment and a breakdown of \emph{when} and \emph{how} an editing approach succeeds or fails. Our formulation supports both the \textbf{online setting} (evaluating $I_e$ against $I_o$ and $T$) and the \textbf{offline setting} (where a ground truth $I_g$ can be used for error analysis).

\subsection{Our MLLM Judge Factors}
\label{sec:our-approach-judge-factors}
To achieve fine-grained, interpretable assessment, we introduce a set of twelve key \textbf{judge factors} for image editing tasks, which are organized into three fundamental, high-level categories that cover the full spectrum of image editing success: \textcolor{googlegreen}{\textit{Image Preservation}}, \textcolor{googleblue}{\textit{Edit Quality}}, and \textcolor{googlered}{\textit{Instruction Fidelity}} as provided in Table \ref{table:Factors}. 
Each factor is scored using a 7-point Likert scale.
An overview of these factors with examples is provided in Figure \ref{fig:overview-factor}}.

\subsubsection{\textcolor{googlegreen}{\bfseries \scshape Image Preservation}}
\label{sec:mllm-judges-image-preservation}
This category evaluates the crucial requirement that a model must preserve regions of the image that are not targeted by the instruction. 
Failures in this category typically indicate over-editing, where unedited content is altered in ways that undermine the user’s intent.
\begin{description}[
    topsep=4pt,            
    leftmargin=0pt,
    itemsep=3pt,   
    parsep=0pt,         
    font=\bfseries\color{googlegreen} 
]
\item[Unchanged Regions:] 
Evaluates whether image regions not specified or implied by the instruction remain unchanged after editing.
    
\item[Global Consistency:] 
Evaluates if the scene's background, style, composition, and color palette remain consistent with the original image outside of the edited area.
    
\item[Identity Preservation:] 
Ensures primary subjects not involved in the edit retain their original identity and recognizable features.
\end{description}

\subsubsection{\textcolor{googleblue}{\bfseries \scshape Edit Quality}}
\label{sec:mllm-judges-edit-quality}
This category evaluates the visual realism and technical correctness of the edited content itself, independent of instruction compliance or preservation of unedited regions.
This set is inspired by classic artistic and image analysis criteria concerning composition and visual coherence.

\begin{description}[
    topsep=4pt,             
    leftmargin=0pt,        
    itemsep=3pt,           
    parsep=0pt,             
    font=\bfseries\color{googleblue}
]

\item[Scale Realism:]
Assesses whether the edited object or region has a realistic size and proportion relative to the scene context and depth.

\item[Spatial Relationship:] 
Evaluates whether edited elements maintain correct spatial relationships and perspective with surrounding objects.

\item[Texture and Detail:]
Checks whether textures and fine details in the edited region are realistic and consistent with the surrounding image.

\item[Image Quality:] 
Determines whether the edited image avoids visible artifacts such as noise, blur, or unnatural distortions.

\item[Color and Lighting:]
Assesses whether the colors, shadows, and lighting of the edited region are consistent with the scene’s illumination.

\item[Seamlessness:]
Evaluates whether transitions between edited and non-edited regions are smooth and visually natural.
\end{description}

\subsubsection{\textcolor{googlered}{\bfseries \scshape Instruction Fidelity}}
\label{sec:mllm-judges-instruction-fidelity}
This category evaluates whether the edited image correctly follows the given textual instruction and reflects the user’s intended semantic content, moving beyond simple pixel comparisons to capture linguistic and contextual understanding.

\begin{description}[
    topsep=4pt,          
    leftmargin=0pt,         
    itemsep=3pt,             
    parsep=0pt,             
    font=\bfseries\color{googlered} 
]

\item[Alignment:] 
Assesses whether the specific edit type and target described in the instruction are correctly realized in the edited image.

\item[Completeness:] 
Evaluates whether all components and constraints of the instruction are fully executed, rather than partially fulfilled.

\item[Plausibility:]
Assesses whether the edited result is visually and physically plausible assuming a generally reasonable instruction, rather than judging the realism of the instruction itself.

\end{description}

\subsection{Base Models and Implementation Details}
Our fine-grained MLLM judges can be built upon any base MLLM. 
In this work, unless otherwise specified, we use GPT-5-mini as the base model, and all results leverage the general implementation provided in Appendix \ref{prompts-sec} (Figure~\ref{fig:prompt-mllm-as-a-judge-image-editing, main}).
This is one of the simplest implementations of our approach, designed to be applicable across different evaluation settings and constitutes the default throughout the paper, unless otherwise mentioned.
Nevertheless, we investigated other implementations of our fine-grained MLLM judges, and report results; please see Appendix \ref{prompts-sec} (Fig.~\ref{fig:prompt-all-factors}-\ref{fig:prompt-mllm-as-a-judge-image-editing, category}), and for other base model results, see Appendix~\ref{sec:ablation_study}.

\section{Methodology}
This section describes our methodology for collecting our benchmark data and evaluation methodology of our fine-grained MLLM judges.

\subsection{Benchmark Collection}
To curate our benchmark, we selected 100 image editing tasks from the HumanEdit data.
More specifically, we sampled uniformly at random 100 
(original image, instruction) pairs, spanning 6 distinct edit types to ensure comprehensive coverage: Add, Remove, Replace, Action, Counting, and Relation (Table \ref{tab:edit-type-nums}). 
We evaluate our MLLM judge in an online setting, where edited images are generated rather than using pre-existing ground truth edits. 
For each pair, we used gpt-image-1 to generate an edited image based on the instruction, resulting in 100 (original, instruction, edited) triplets for evaluation. 
For more insight, refer to Appendix~\ref{sec:appendix-benchmark}.
By proposing a set of fine-grained factors shown in Table~\ref{table:Factors} and grounding them in human annotations, we created a gold standard that can be used to evaluate if an MLLM judge actually thinks like a human at the finest granularity.

\subsection{Participants and Procedure} \label{sec:participants}

\paragraph{Recruitment.}
We recruited 25 annotators representing diverse demographics, including undergraduate and graduate students, early-career professionals, and experienced practitioners. This diversity ensures our evaluation captures varied perspectives on image quality and editing success across different user backgrounds and expertise levels.

\paragraph{Sample Size and Coverage.}
Our study evaluates 100 image editing tasks, each comprising an original image, an editing instruction, and an edited image. For reliability analysis, each task is rated by five annotators. We recruited 25 participants, with each evaluating 20 randomly sampled images spanning six edit types.

\paragraph{Evaluation Structure.}
Every image editing task was judged using the 12 factors described in Section 2 and summarized in Table~\ref{table:Factors}. Participants also provided an overall quality rating, yielding 13 scores per image (12 factors + 1 overall). Each participant provided 13 scores for each of their 20 assigned images, generating 260 scores per participant (13 scores × 20 images).

\subsection{Human Study Interface}
\label{sec:procedure}
\begin{figure}[h!]
    \centering
    \includegraphics[width=\linewidth]{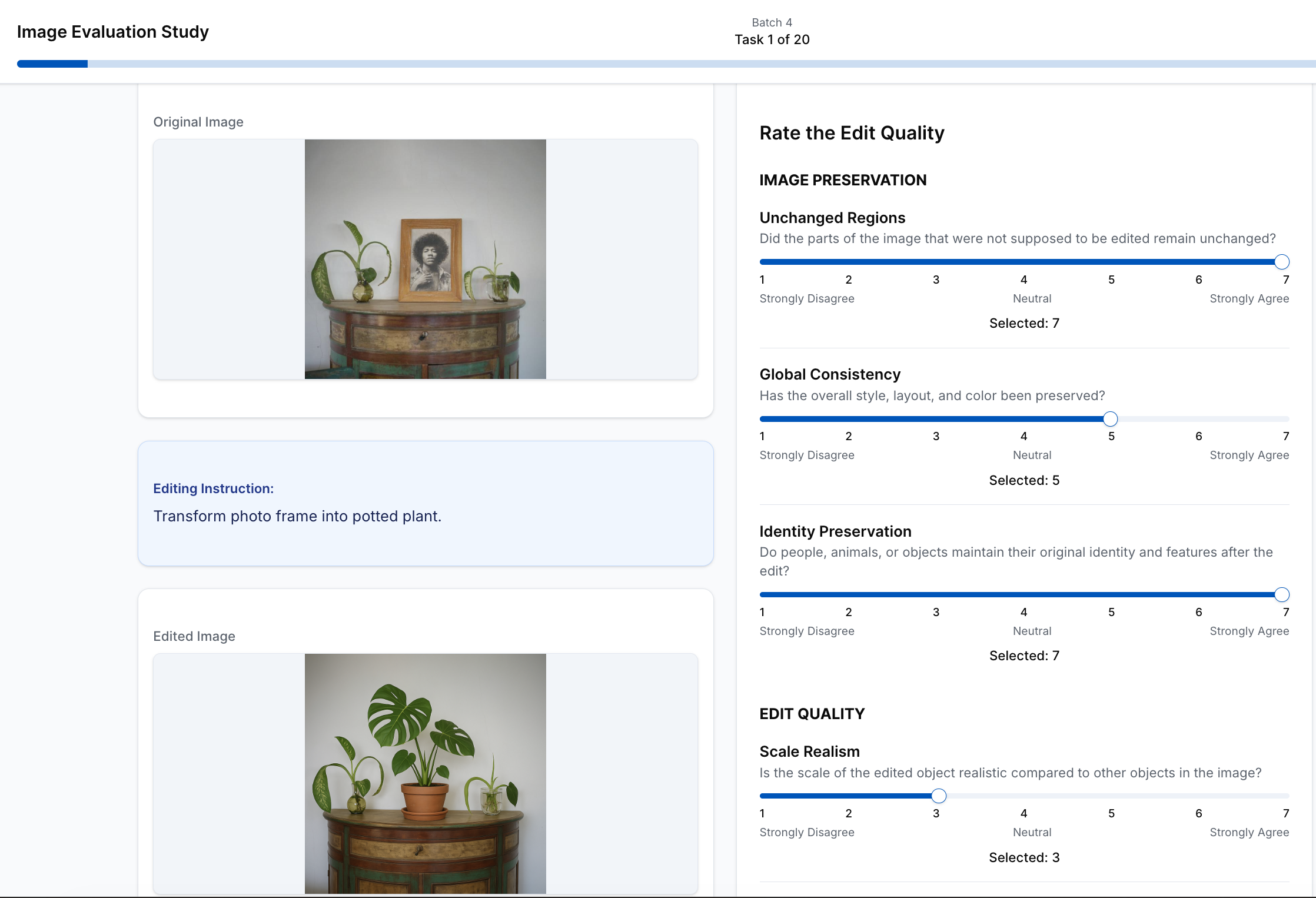}
    \caption{Human evaluation study interface illustrating an example image-editing task. Evaluators are shown the original image, the editing instruction, and the edited image (the API generated image), and are asked to rate multiple dimensions using Likert-scale judgments. These annotations form the benchmark human evaluation dataset used in our analysis. 
    }
    \label{fig:human-eval-ui-Q!}
\end{figure}

Figure~\ref{fig:human-eval-ui-Q!} shows our evaluation interface. This top-to-bottom arrangement follows natural reading flow and allows participants to easily compare before and after states while keeping the instruction visible. A progress indicator at the top displays "Task X of 20" to track evaluation progress.

\subsection{Metrics}
\label{sec:metrics}
We compute several metrics to assess both human judgment characteristics and human-MLLM alignment. For human evaluation, we analyze mean and standard deviation of scores per factor. For human-MLLM alignment, we compute: (1) score prediction accuracy using Mean Squared Error (MSE) and Mean Absolute Error (MAE), (2) ranking correlation using Pearson, Spearman's $\rho$, and Kendall's $\tau$, (3) agreement rates measuring percentage of cases where MLLM and human scores differ by $\le$1 point.
Further details for all the metrics 
are provided in Appendix~\ref{metrics}.

\renewcommand{\arraystretch}{1.0}
\begin{table}[t]
\centering
\tiny
\setlength{\tabcolsep}{3pt} 
\caption{
Our benchmark and the distribution of image edits across the various editing tasks.
}
\label{tab:edit-type-nums}
\begin{tabular}{@{}lccccccc@{}}
\toprule
 & Add & Remove & Replace & Action & Counting & Relation & Total \\
\midrule
\textbf{our benchmark} & 9 & 34 & 18 & 23 & 10 & 6 & 100 \\
\bottomrule
\end{tabular}
\end{table}

\section{Experiments}\label{sec:exp}
We evaluate the effectiveness of MLLMs and our fine-grained judges for assessing image editing tasks by comparing their judgements to human judgements collected through our study  (Section~\ref{sec: human-study}).
We compare the effectiveness of our approaches in capturing all the image editing factors proposed in Table~\ref{table:Factors}.

\subsection{Experimental Setup}
For this work, we construct our benchmark based on the HumanEdit dataset~\cite{humanedit2025}, which provides high-quality human-annotated image editing data. The benchmark is used throughout all subsequent evaluations, with additional details on its construction, tasks, and usage provided in Appendix~\ref{app:benchmark_construction}.
Our benchmark covers six edit types, namely add, replace, remove, counting, action, and relation. Each edit type targets a distinct class of image editing behaviors, and their detailed definitions are provided in Appendix~\ref{app:benchmark_tasks}.

\subsection{Results}
\label{sec:results}
We evaluate the effectiveness of our fine-grained MLLM judge by comparing its scores against human judgments across the 12 fine-grained factors in Table~\ref{table:Factors}.
Scores were averaged across annotators for each edited image.

In Table~\ref{tab:human_judge_v1}, we provide a detailed comparison between human judges and our fine-grained MLLM judges across the 12 proposed fine-grained image editing factors related to \textcolor{googlegreen}{image preservation}, \textcolor{googleblue}{edit quality}, and \textcolor{googlered}{instruction fidelity}, stratified by \emph{edit type}. 
The results in Table~\ref{tab:human_judge_v1} demonstrate a high level of alignment between the proposed MLLM judges and human evaluators across diverse types of image edits.
In particular, across nearly all factors and image edit types, our approach is shown to align strongly with human evaluators while providing more consistent scoring behavior, especially for complex edit types such as remove, replace, and relation.
Furthermore, our fine-grained judges align closely with human evaluations across all three factor groups, with especially strong agreement on \textcolor{googlegreen}{image preservation} and \textcolor{googleblue}{edit quality}. 
The judge maintains stable performance for instruction fidelity under challenging edit types, giving rise to comparable or higher aggregate averages despite increased variance at the per edit level. 

As shown in Table~\ref{tab:results-summary}, the proposed MLLM judge achieves a high degree of alignment with human evaluators across image editing factor categories (\textcolor{googlegreen}{image preservation}, \textcolor{googleblue}{edit quality}, and \textcolor{googlered}{instruction fidelity}); see Table~\ref{table:Factors} for the specific fine-grained factors under each category.
Furthermore, our judge closely tracks human evaluations across all three factor groups, with especially strong agreement on \textcolor{googlegreen}{image preservation} and \textcolor{googleblue}{edit quality}.

We now discuss several other findings we observed.
More specifically, Table~\ref{tab:human_judge_v1} reveals that the judge has strong performance on \textcolor{googleblue}{edit quality} with almost all the blocks shaded green. This means that the judge captured many of the same visual cues and criteria that humans relied on when assessing edited images in this category. Moreover, the relatively small gray standard deviations accompanying the judge scores in these aligned regions indicated stable and confident judgments, often comparable to or even lower than those of human annotators. This was particularly evident in task types Add, Replace, and Counting, where agreement between human and judge evaluations was frequent and the judge exhibited low variance across all factors. The judge also perform well on \textcolor{googlegreen}{image preservation}, with global consistency task nearly all shaded green. Even some of the block have large difference for unchanged regions, the all edits column still show strong alignment. Even for more challenging \textcolor{googlered}{instruction fidelity} aspects, several edit types showed close agreement with human scores, suggesting that the judge reliably assessed not only perceptual quality but also semantic adherence to instructions.

\subsection{Ablation Study}
\label{sec:ablation_study}

\subsubsection{Varying LLM}
When designing our MLLM judge, we evaluate two mainstream MLLMs, GPT-5-mini and Gemini-2.5-pro. Results for both models are reported in Table~\ref{tab:human_gpt_gemini}. Overall, GPT-5-mini shows consistently closer alignment with human evaluations than Gemini-2.5-pro across most factors, edit types, and the aggregated score. Nevertheless, Gemini-2.5-pro outperforms GPT-5-mini on specific factors and tasks, such as the Unchanged Regions factor in the Add task. More comprehensive results for all implementations are provided in Appendix~\ref{app:ablation_study}.

\subsubsection{Varying Judge Instructions} 
We also investigated a variety of different implementations of our fine-grained MLLM judges. 
\begin{itemize}
\setlength{\itemsep}{2pt}
\setlength{\parskip}{0pt}
\setlength{\parsep}{0pt}
    \item \textbf{Main (Figure~\ref{fig:prompt-all-factors}):} This implementation only includes the most basic instructions.
    \item \textbf{Factor-level Rubrics (Figure~\ref{fig:prompt-mllm-as-a-judge-image-editing, main}):} This implementation includes fine-grained scoring rubrics for each judge factor.
    \item \textbf{Category wise, Example guided (Figure~\ref{fig:prompt-mllm-as-a-judge-image-editing, category}):} 
    This implementation groups factors into three categories and uses category-specific prompts with detailed rubrics and examples.
\end{itemize}

As shown in Table~\ref{tab:human_judge_v1}, Table~\ref{tab:human_judge_v3}, and Table~\ref{tab:human_judge_combined_category}, the prompt illustrated in Figure~\ref{fig:prompt-all-factors} consistently achieves the best result. 
Additional result analysis are provided in Appendix~\ref{app:ablation_study}. Among all traditional metrics, CLIP and DINO Image performs the best bust still only present limited alignment with human evaluation result, showing the need for a more suitable evaluator. Full quantitative results and analysis are provided in Appendix \ref{appendix-trad}.

\section{Conclusion} \label{sec:conc}
This work advances the evaluation of image editing approaches by moving beyond coarse and opaque metrics toward a fine grained, human aligned judging paradigm. By decomposing image editing quality into twelve interpretable factors spanning image preservation, edit quality, and instruction fidelity, we provide a principled lens for understanding when and why image edits succeed or fail. The accompanying human validated benchmark, which unifies human judgments, MLLM based evaluations, model outputs, and traditional metrics, enables systematic and reproducible analysis across a diverse set of image editing tasks. Our empirical findings demonstrate that MLLM judges align closely with human evaluations at a fine granularity, while commonly used metrics are often ineffective proxies for the aspects that matter most to users. Together, these contributions establish a practical foundation for diagnosing, comparing, and improving image editing approaches, and suggest a clear path toward more interpretable, reliable, and human aligned evaluation in both offline benchmarking and online development settings.

\section{Potential Risks}
Over-reliance on human evaluator benchmarks introduces several potential risks. It may encode annotator subjectivity and cultural bias, encourage models to overfit benchmark-specific preferences, and mask inter-annotator disagreement through aggregated scores. Moreover, systems optimized for human benchmarks may avoid unconventional yet valid edits that deviate from annotator expectations, thereby discouraging creativity and limiting generalization to novel editing behaviors.

\section{Limitations} 
While this work introduces a fine-grained, human-aligned MLLM-as-a-judge framework for image editing, several limitations remain. 
First, our benchmark and evaluations focus on a fixed set of image editing tasks and factor definitions, which, although representative, may not exhaustively capture all real-world editing scenarios or emerging edit types. 
Second, our fine-grained judges leverage state-of-the-art MLLMs, and their behavior naturally reflects the capabilities of these models at the time of evaluation. As MLLMs continue to improve, the proposed framework can directly benefit from improved reasoning and perception.
Finally, while we observe strong alignment with human judgments, our study does not eliminate the need for humans in high-stakes or subjective applications, and further investigation is required to understand failure modes and edge cases.

\section{Ethical Considerations}
The human evaluations were conducted with informed consent and appropriate quality controls. While MLLM-based judges may reflect biases in their underlying models or training data, we position our framework as a diagnostic and evaluative tool, and encourage careful use and continued auditing when applied in sensitive or high-stakes image editing settings.

\bibliographystyle{acl_natbib}
\bibliography{main}

\begin{thebibliography}{30}
\expandafter\ifx\csname natexlab\endcsname\relax\def\natexlab#1{#1}\fi

\bibitem[{Anonymous(2026)}]{mlm_uijudge2026}
Anonymous. 2026.
\newblock \href {https://doi.org/10.1145/nnnnnnn.nnnnnnn} {Mllm as a ui judge: Benchmarking multimodal llms for predicting human perception of user interfaces}.
\newblock \emph{Proceedings of the ACM}.
\newblock Under review.

\bibitem[{Balaji et~al.(2023)Balaji, Nah, Huang, Vahdat, Song, Zhang, Kreis, Aittala, Aila, Laine, Catanzaro, Karras, and Liu}]{balaji2023ediffi}
Yogesh Balaji, Seungjun Nah, Xun Huang, Arash Vahdat, Jiaming Song, Qinsheng Zhang, Karsten Kreis, Miika Aittala, Timo Aila, Samuli Laine, Bryan Catanzaro, Tero Karras, and Ming-Yu Liu. 2023.
\newblock \href {https://arxiv.org/abs/2211.01324} {ediff-i: Text-to-image diffusion models with an ensemble of expert denoisers}.
\newblock \emph{arXiv preprint arXiv:2211.01324}.

\bibitem[{Basu et~al.(2023)Basu, Saberi, Bhardwaj, Chegini, Massiceti, Sanjabi, Hu, and Feizi}]{basu2023editval}
Samyadeep Basu, Mehrdad Saberi, Shweta Bhardwaj, Atoosa~Malemir Chegini, Daniela Massiceti, Maziar Sanjabi, Shell~Xu Hu, and Soheil Feizi. 2023.
\newblock \href {https://arxiv.org/abs/2310.02426} {Editval: Benchmarking diffusion based text-guided image editing methods}.
\newblock \emph{arXiv preprint arXiv:2310.02426}.

\bibitem[{Brooks et~al.(2023)Brooks, Holynski, and Efros}]{brooks2023instructpix2pix}
Tim Brooks, Aleksander Holynski, and Alexei~A. Efros. 2023.
\newblock \href {https://arxiv.org/abs/2211.09800} {Instructpix2pix: Learning to follow image editing instructions}.
\newblock \emph{arXiv preprint arXiv:2211.09800}.

\bibitem[{Caron et~al.(2021)Caron, Touvron, Misra, J{\'e}gou, Mairal, Bojanowski, and Joulin}]{caron2021emerging}
Mathilde Caron, Hugo Touvron, Ishan Misra, Herv{\'e} J{\'e}gou, Julien Mairal, Piotr Bojanowski, and Armand Joulin. 2021.
\newblock Emerging properties in self-supervised vision transformers.
\newblock \emph{arXiv preprint arXiv:2104.14294}.

\bibitem[{Couairon et~al.(2022)Couairon, Verbeek, Schwenk, and Cord}]{couairon2022diffedit}
Guillaume Couairon, Jakob Verbeek, Holger Schwenk, and Matthieu Cord. 2022.
\newblock \href {https://arxiv.org/abs/2210.11427} {Diffedit: Diffusion-based semantic image editing with mask guidance}.
\newblock \emph{arXiv preprint arXiv:2210.11427}.

\bibitem[{Fu et~al.(2023)Fu, Peng, Xu, Yan, Sun, Liu, and Zhou}]{fu2023judge}
Jinlan Fu, Hao Peng, Zhenhailong Xu, Chuanqi Yan, Mingzhe Sun, Weizhu Liu, and Jie Zhou. 2023.
\newblock Gptscore: Evaluate as you desire.
\newblock \emph{arXiv preprint arXiv:2302.04166}.

\bibitem[{Goel et~al.(2022)Goel, Bansal, Bhatia, Rossi, Vinay, and Grover}]{goel2022cyclip}
Shashank Goel, Hritik Bansal, Sumit Bhatia, Ryan~A. Rossi, Vishwa Vinay, and Aditya Grover. 2022.
\newblock \href {https://arxiv.org/abs/2205.14459} {Cyclip: Cyclic contrastive language--image pretraining}.
\newblock \emph{arXiv preprint arXiv:2205.14459}.

\bibitem[{Hertz et~al.(2022)Hertz, Mokady, Tenenbaum, Aberman, Pritch, and Cohen-Or}]{hertz2022prompttoprompt}
Amir Hertz, Ron Mokady, Jay Tenenbaum, Kfir Aberman, Yael Pritch, and Daniel Cohen-Or. 2022.
\newblock \href {https://arxiv.org/abs/2208.01626} {Prompt-to-prompt image editing with cross attention control}.
\newblock \emph{arXiv preprint arXiv:2208.01626}.

\bibitem[{Hessel et~al.(2021)Hessel, Holtzman, Forbes, Le~Bras, and Choi}]{hessel2021clipscore}
Jack Hessel, Ari Holtzman, Maxwell Forbes, Ronan Le~Bras, and Yejin Choi. 2021.
\newblock \href {https://arxiv.org/abs/2104.08718} {Clipscore: A reference-free evaluation metric for image captioning}.
\newblock \emph{arXiv preprint arXiv:2104.08718}.

\bibitem[{Ho et~al.(2021)Ho, Saharia, Chan, Fleet, Norouzi, and Salimans}]{ho2021cascadeddiffusion}
Jonathan Ho, Chitwan Saharia, William Chan, David~J. Fleet, Mohammad Norouzi, and Tim Salimans. 2021.
\newblock \href {https://arxiv.org/abs/2106.15282} {Cascaded diffusion models for high fidelity image generation}.
\newblock \emph{arXiv preprint arXiv:2106.15282}.

\bibitem[{Hsu et~al.(2023)Hsu, Huang, Rossi, Kim, Giles, and Huang}]{hsu2023gpt}
Ting-Yao Hsu, Chieh-Yang Huang, Ryan Rossi, Sungchul Kim, C~Lee Giles, and Ting-Hao~K Huang. 2023.
\newblock Gpt-4 as an effective zero-shot evaluator for scientific figure captions.
\newblock \emph{arXiv preprint arXiv:2310.15405}.

\bibitem[{Jain(1989)}]{jain1989digitalimage}
Anil~K. Jain. 1989.
\newblock \emph{Fundamentals of Digital Image Processing}.
\newblock Prentice-Hall.

\bibitem[{Kawar et~al.(2022)Kawar, Zada, Lang, Tov, Chang, Dekel, Mosseri, and Irani}]{kawar2022imagic}
Bahjat Kawar, Shiran Zada, Oran Lang, Omer Tov, Huiwen Chang, Tali Dekel, Inbar Mosseri, and Michal Irani. 2022.
\newblock \href {https://arxiv.org/abs/2210.09276} {Imagic: Text-based real image editing with diffusion models}.
\newblock \emph{arXiv preprint arXiv:2210.09276}.

\bibitem[{Kim et~al.(2025)Kim, Choi, Rossi, Koh, and Lee}]{chart2experience2025}
Seon~Gyeom Kim, Jae~Young Choi, Ryan Rossi, Eunyee Koh, and Tak~Yeon Lee. 2025.
\newblock \href {https://doi.org/10.1109/PacificVis60772.2025.00045} {Chart-to-experience: Benchmarking multimodal llms for predicting experiential impact of charts}.
\newblock In \emph{2025 IEEE 18th Pacific Visualization Conference (PacificVis)}, pages 340--345. IEEE.

\bibitem[{Kim et~al.(2024)Kim, Suk, Longpre, Lin, Shin, Welleck, Neubig, Lee, Lee, and Seo}]{prometheus2024}
Seungone Kim, Juyoung Suk, Shayne Longpre, Bill~Yuchen Lin, Jamin Shin, Sean Welleck, Graham Neubig, Moontae Lee, Kyungjae Lee, and Minjoon Seo. 2024.
\newblock \href {https://arxiv.org/abs/2405.01535} {Prometheus 2: An open source language model specialized in evaluating other language models}.
\newblock \emph{arXiv preprint arXiv:2405.01535}.

\bibitem[{Mokady et~al.(2022)Mokady, Hertz, Aberman, Pritch, and Cohen-Or}]{mokady2022nulltext}
Ron Mokady, Amir Hertz, Kfir Aberman, Yael Pritch, and Daniel Cohen-Or. 2022.
\newblock \href {https://arxiv.org/abs/2211.09794} {Null-text inversion for editing real images using guided diffusion models}.
\newblock \emph{arXiv preprint arXiv:2211.09794}.

\bibitem[{Pan et~al.(2025)Pan, He, Mao, Han, Jiang, Zhang, and Liu}]{pan2025icebench}
Yulin Pan, Xiangteng He, Chaojie Mao, Zhen Han, Zeyinzi Jiang, Jingfeng Zhang, and Yu~Liu. 2025.
\newblock \href {https://arxiv.org/abs/2503.14482} {Ice-bench: A unified and comprehensive benchmark for image creating and editing}.
\newblock \emph{arXiv preprint arXiv:2503.14482}.

\bibitem[{Ramesh et~al.(2022)Ramesh, Dhariwal, Nichol, Chu, and Chen}]{ramesh2022dalle2}
Aditya Ramesh, Prafulla Dhariwal, Alex Nichol, Casey Chu, and Mark Chen. 2022.
\newblock \href {https://arxiv.org/abs/2204.06125} {Hierarchical text-conditional image generation with clip latents}.
\newblock \emph{arXiv preprint arXiv:2204.06125}.

\bibitem[{Rombach et~al.(2022)Rombach, Blattmann, Lorenz, Esser, and Ommer}]{rombach2022stablediffusion}
Robin Rombach, Andreas Blattmann, Dominik Lorenz, Patrick Esser, and Bj{\"o}rn Ommer. 2022.
\newblock \href {https://doi.org/10.1109/CVPR52688.2022.01042} {High-resolution image synthesis with latent diffusion models}.
\newblock In \emph{Proceedings of the IEEE/CVF Conference on Computer Vision and Pattern Recognition (CVPR)}, pages 10684--10695.

\bibitem[{Ruiz et~al.(2022)Ruiz, Li, Jampani, Pritch, Rubinstein, and Aberman}]{ruiz2022dreambooth}
Nataniel Ruiz, Yuanzhen Li, Varun Jampani, Yael Pritch, Michael Rubinstein, and Kfir Aberman. 2022.
\newblock \href {https://arxiv.org/abs/2208.12242} {Dreambooth: Fine tuning text-to-image diffusion models for subject-driven generation}.
\newblock \emph{arXiv preprint arXiv:2208.12242}.

\bibitem[{Saharia et~al.(2022)Saharia, Chan, Saxena, Li, Whang, Denton, Ghasemipour, Karagol~Ayan, Mahdavi, Lopes, Salimans, Ho, Fleet, and Norouzi}]{saharia2022imagen}
Chitwan Saharia, William Chan, Saurabh Saxena, Lala Li, Jay Whang, Emily Denton, Seyed Kamyar~Seyed Ghasemipour, Burcu Karagol~Ayan, S.~Sara Mahdavi, Rapha~Gontijo Lopes, Tim Salimans, Jonathan Ho, David~J. Fleet, and Mohammad Norouzi. 2022.
\newblock \href {https://arxiv.org/abs/2205.11487} {Photorealistic text-to-image diffusion models with deep language understanding}.
\newblock \emph{arXiv preprint arXiv:2205.11487}.

\bibitem[{Shi et~al.(2023)Shi, Xiong, Lin, and Jung}]{shi2023instantbooth}
Jing Shi, Wei Xiong, Zhe Lin, and Hyun~Joon Jung. 2023.
\newblock \href {https://arxiv.org/abs/2304.03411} {Instantbooth: Personalized text-to-image generation without test-time finetuning}.
\newblock \emph{arXiv preprint arXiv:2304.03411}.

\bibitem[{Sun et~al.(2023)Sun, Zhang, Chen et~al.}]{sun2023magicbrush}
Hongyu Sun, Wenliang Zhang, Yuwei Chen, et~al. 2023.
\newblock Magicbrush: A large-scale dataset for text-guided image editing.
\newblock In \emph{Advances in Neural Information Processing Systems (NeurIPS)}.

\bibitem[{Wang et~al.(2025)}]{humanedit2025}
Bryan Wang et~al. 2025.
\newblock Humanedit: A benchmark for instruction-based human image editing.
\newblock \emph{Proceedings of the IEEE/CVF Conference on Computer Vision and Pattern Recognition (CVPR)}.

\bibitem[{Wang et~al.(2004)Wang, Bovik, Sheikh, and Simoncelli}]{wang2004ssim}
Zhou Wang, Alan~C. Bovik, Hamid~R. Sheikh, and Eero~P. Simoncelli. 2004.
\newblock \href {https://doi.org/10.1109/TIP.2003.819861} {Image quality assessment: From error visibility to structural similarity}.
\newblock \emph{IEEE Transactions on Image Processing}, 13(4):600--612.

\bibitem[{Yosef et~al.(2025)Yosef, Yanuka, Bitton, and Lischinski}]{yosef2025editinspector}
Ron Yosef, Moran Yanuka, Yonatan Bitton, and Dani Lischinski. 2025.
\newblock \href {https://arxiv.org/abs/2506.09988} {Editinspector: A benchmark for evaluation of text-guided image edits}.
\newblock \emph{arXiv preprint arXiv:2506.09988}.

\bibitem[{Zhang et~al.(2018)Zhang, Isola, Efros, Shechtman, and Wang}]{zhang2018lpips}
Richard Zhang, Phillip Isola, Alexei~A. Efros, Eli Shechtman, and Oliver Wang. 2018.
\newblock \href {https://arxiv.org/abs/1801.03924} {The unreasonable effectiveness of deep features as a perceptual metric}.
\newblock In \emph{Proceedings of the IEEE Conference on Computer Vision and Pattern Recognition (CVPR)}.

\bibitem[{Zhang et~al.(2022)Zhang, Han, Ghosh, Metaxas, and Ren}]{zhang2022sine}
Zhixing Zhang, Ligong Han, Arnab Ghosh, Dimitris Metaxas, and Jian Ren. 2022.
\newblock \href {https://arxiv.org/abs/2212.04489} {Sine: Single image editing with text-to-image diffusion models}.
\newblock \emph{arXiv preprint arXiv:2212.04489}.

\bibitem[{Zheng et~al.(2023)Zheng, Chiang, Sheng, Zhuang, Wu, Zhuang, Lin, Li, Li, Xing, Zhang, Gonzalez, and Stoica}]{li2023judging}
Lianmin Zheng, Wei-Lin Chiang, Ying Sheng, Siyuan Zhuang, Zhanghao Wu, Yonghao Zhuang, Zi~Lin, Zhuohan Li, Dacheng Li, Eric~P. Xing, Hao Zhang, Joseph~E. Gonzalez, and Ion Stoica. 2023.
\newblock Judging llm-as-a-judge with mt-bench and chatbot arena.
\newblock \emph{arXiv preprint arXiv:2306.05685}.

\end{thebibliography}

\appendix
\section*{Appendix}
\section{Problem Motivation}
\label{app: prob motivation}

The purpose of this section is to further explain why traditional metrics are structurally misaligned with goals of instruction-driven image editing. Refer to table \ref{trad_met_def} for a detailed list of the traditional metrics used in our study.

A central limitation of the traditional metrics is the fundamental mismatch between what they are designed to measure and the goal of image editing. Pixel-level metrics such as PSNR and SSIM explicitly reward similarity to the original image, whereas successful editing is defined by the quality of the intentional differences. As a result, high-quality edits that introduce desirable, instruction-aligned changes are penalized as if they were distortions, causing these metrics to incorrectly classify valid edits as errors. 

Traditional metrics also fail because they conflate pixel similarity with semantic correctness. In tasks requiring substantial visual changes—such as replacing an object (“replace the dog with a cat”)—metrics like PSNR and SSIM heavily penalize the large pixel differences that naturally arise from a successful transformation. 

\begin{figure}[H]
\centering
\includegraphics[width=\linewidth]{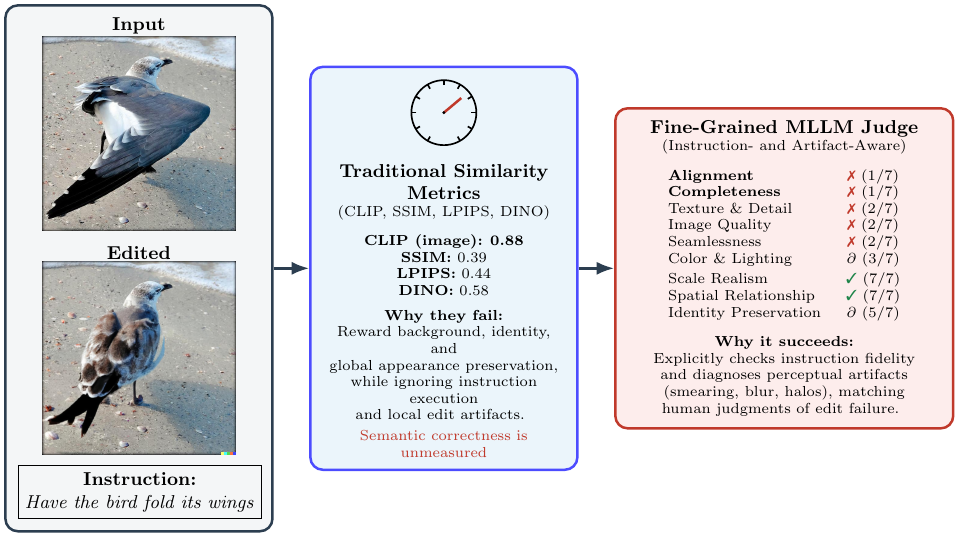} 
\vspace{-0.5cm}
\caption{
Similarity-based metrics reward global appearance and identity preservation, while fine-grained MLLM judges evaluate instruction fidelity and local edit quality, correctly diagnosing semantic edit failures.
}
\label{fig:motivation}
\end{figure}

Finally, metrics like CLIPScore, while capturing coarse semantic alignment, provide no mechanism for evaluating instructional fidelity. For a prompt like “make the sky brighter,” a high CLIPScore may simply reflect the co-occurrence of “sky” and “bright,” regardless of whether the model brightened the correct region or overexposed unrelated areas. 

\section{Additional Related Work}
\label{app:related-work}

This appendix provides a more detailed discussion of prior image editing paradigms and the use of large multimodal models as evaluators.

\subsection{Agentic and Tool-Based Image Editing}
Agentic image editing systems decompose a complex user instruction into a sequence of specialized, interpretable operations, rather than producing the edited image in a single generative step. Recent advances in multimodal large language models (MLLMs) have enabled this paradigm by allowing models to reason over an instruction and orchestrate external tools such as segmentation, inpainting, object detection, color adjustment, and style transfer. Representative works such as InstructPix2Pix~\cite{brooks2023instructpix2pix} and MagicBrush~\cite{sun2023magicbrush} demonstrate that step-wise editing pipelines can yield localized, controllable edits and improve interpretability by making intermediate decisions explicit.

This agentic paradigm is particularly well-suited for precision-critical scenarios, where edits must be tightly constrained to specific regions of the image. By explicitly separating localization, modification, and blending, agentic systems can reduce unintended global changes and better preserve non-target regions. However, these benefits come at the cost of increased computational overhead, longer inference times, and added system complexity due to tool orchestration and error propagation across steps.

\subsection{Direct Generative Image Editing}
In contrast, direct image editing methods leverage a single forward pass of a generative model conditioned on the input image and a textual instruction to produce the edited output. Approaches such as Stable Diffusion \texttt{img2img}~\cite{rombach2022stablediffusion}, DALL·E~2 inpainting~\cite{ramesh2022dalle2}, and Imagen Editor~\cite{saharia2022imagen} have demonstrated strong performance for stylistic and scene-wide transformations. These models benefit from efficiency, visual coherence, and ease of deployment, making them attractive for creative and consumer-facing applications.

However, direct methods often struggle with controllability. Localized edits can inadvertently trigger global re-rendering, altering background appearance, identity features, or scene composition beyond the user’s intent. For example, modifying the color of a single object may unintentionally affect lighting, texture, or nearby objects. These failure modes directly motivate evaluation criteria related to image preservation and edit localization, which are difficult to capture using traditional similarity-based metrics.

\subsection{MLLMs as Judges for Image Editing}
Human evaluation remains the gold standard for assessing image editing quality, but it is costly, slow, and difficult to scale. This has motivated a growing body of work on \textit{LLM-as-a-Judge} paradigms, where large language or multimodal models are used to approximate human preferences. Prior studies have applied LLMs and MLLMs as judges across a range of subjective evaluation tasks, including interface analysis, chart understanding, and text generation~\cite{mlm_uijudge2026, chart2experience2025, prometheus2024, fu2023judge, li2023judging}.

More recent efforts have explored MLLM-based judges specifically for image editing~\cite{fu2023judge, hsu2023gpt, prometheus2024}, demonstrating that MLLMs can reason about high-level semantics and user intent beyond pixel-level similarity. However, these approaches are not designed around the specific requirements of instruction-guided image editing as formalized in this work. In particular, they do not provide factorized evaluation aligned with the competing objectives of preserving non-edited content, producing technically realistic edits, and faithfully executing the instruction.

\definecolor{googleblue}{RGB}{66,133,244}
\definecolor{googlegreen}{RGB}{52,168,83}
\definecolor{googlered}{RGB}{234,67,53}
\definecolor{googleyellow}{RGB}{0,150,136}
\definecolor{googlepurple}{RGB}{142,36,170}

\begin{table*}[t]
\centering
\small
\scriptsize
\caption{
We propose a comprehensive taxonomy that summarizes the traditional image editing metrics into pixel-level fidelity, content preservation, and semantic alignment.
Each metric is categorized by purpose, assumed input(s), and reference dependence. 
Note that while we list assumed inputs, many of these also make sense for others, including those that the online setting assumes, where ground-truth is unknown.
}
\label{trad_met_def}
\begin{tabularx}{\textwidth}{@{} l p{40mm} l l}
\toprule
 & \textbf{Metric} & \textbf{Assumed Input(s)} & \textbf{Reference Type}  \\
\midrule

\multirow{5}{*}{\textcolor{googleblue}{\sc \bfseries Pixel-Level Fidelity}} 
& L1 Distance ($\downarrow$) & Edited image, Ground-truth image & Reference-based   \\
& L2 Distance / MSE ($\downarrow$) & Edited image, Ground-truth image & Reference-based   \\
& PSNR ($\uparrow$) & Edited image, Ground-truth image & Reference-based   \\
& SSIM ($\uparrow$) & Edited image, Ground-truth image & Reference-based   \\
& LPIPS ($\downarrow$) & Edited image, Ground-truth image & Reference-based  \\

\midrule

\multirow{3}{*}{\textcolor{googlegreen}{\sc \bfseries Content Preservation}} 
& Mask-SSIM ($\uparrow$) & Edited image, Ground-truth image, Edit mask & Reference-based  \\
& Mask-LPIPS ($\downarrow$) & Edited image, Ground-truth image, Edit mask & Reference-based  \\
& Background Consistency ($\uparrow$) & Edited image, Original image & Reference-based  \\

\midrule

\multirow{2}{*}{\textcolor{googlered}{\sc \bfseries Semantic Alignment}} 
& CLIPScore ($\uparrow$) & Edited image, Instruction & Reference-free  \\
& DINO Similarity ($\uparrow$) & Edited image, Ground-truth image & Reference-based  \\

\bottomrule
\end{tabularx}
\end{table*}

\section{Traditional Metrics}
\label{appendix-trad}
Refer to Tables \ref{tab:L1_L2_gpt_gemini_image_preservation_offline} through for detailed correlation results between our Judge and the traditional metrics.
\subsection{Pixel-Level Traditional Metrics Analysis Overview}

To contextualize the performance of our MLLM-based judge, we evaluate a comprehensive set of traditional automated metrics, including pixel-level, perceptual, mask-based, and semantic similarity measures, as summarized in Table \ref{trad_met_def}. Pixel-level and perceptual metrics are computed under both offline and online evaluation settings. To better isolate localized edits, we further consider mask-based variants of these metrics. In addition, we assess semantic similarity using CLIP text embeddings, CLIP image embeddings, and DINO image representations. Detailed comparison results for all metrics are provided in Table~\ref{tab:trad_judge_evals}.

Traditional pixel-level and perceptual metrics primarily measure visual similarity and fail to capture higher-level properties such as semantic correctness, instruction fidelity, and perceptual coherence. Consistent with this limitation, we observe extremely weak correlations between all traditional metrics and our fine-grained judge factors, indicating that these metrics do not reflect the nuanced aspects of editing quality that are most relevant to human judgment. Notably, correlations in the online setting are generally stronger (i.e., more negative), suggesting that the online judge applies stricter criteria when evaluating edited images.

\subsubsection{L1: Absolute Pixel Difference}
L1 measures the pixel-wise absolute difference between two images, where lower values indicate greater similarity. L1 scores are typically lower in the offline setting because the edited image is compared to a known ground-truth target. Among all pixel-level metrics, L1 exhibits the strongest correlation with the \textit{Unchanged Regions} factor in both settings, as it is sensitive to unintended pixel changes in areas that should remain untouched. L1 also correlates most strongly with \textit{Scale Realism} in the online setting. Unrealistic scaling of the edited object alters surrounding pixels, resulting in larger absolute differences that L1 detects. Despite these being the strongest relationships among pixel-level metrics, all correlations remain very weak.

\subsubsection{L2: Squared Pixel Difference}
L2 measures the squared pixel-wise difference between images. It is most correlated with the \textit{Spatial Relationship} factor in both online and offline settings. Spatial relationship errors—such as misalignment, incorrect placement, or overlapping elements—produce large pixel discrepancies that L2 captures. However, the magnitude of these correlations is still small, showing that L2 does not effectively capture higher-level semantic correctness in image editing.

\subsubsection{PSNR: Peak Signal-to-Noise Ratio}
PSNR compares the image signal to the noise, with higher scores indicating closer similarity. PSNR shows almost no correlation with any judge factor. Its highest correlation, still close to zero, is with \textit{Identity Preservation} in the offline setting. Because PSNR treats all deviations as noise and lacks any semantic understanding, it fails to provide meaningful insight into editing quality.

\subsubsection{SSIM: Structural Similarity Index}
SSIM measures how similar two images are based on local structural information, evaluating luminance, contrast, and structural coherence using patches rather than raw pixels. SSIM demonstrates somewhat stronger correlations with judge factors tied to overall coherence, including \textit{Global Consistency}, \textit{Unchanged Regions}, and \textit{Color and Lighting}. Although these correlations remain weak, SSIM outperforms basic pixel-level metrics because its patch-based structure better captures global image consistency.

\subsubsection{LPIPS: Learned Perceptual Image Patch Similarity}
LPIPS compares deep visual features rather than individual pixels, making it more aligned with human perceptual judgments. LPIPS shows the strongest correlations across all pixel-level metrics, particularly with the \textit{Texture and Detail} judge factor. This occurs because LPIPS embeddings capture fine-grained visual inconsistencies, texture mismatches, and perceptual artifacts that humans naturally identify. LPIPS also responds strongly to semantic or textual editing errors that produce noticeable perceptual differences.

\subsubsection{Summary}
Across all pixel-level metrics, we find that none correlate strongly with judge factors, highlighting the mismatch between pixel similarity and human evaluations of editing correctness. The online judge generally exhibits stronger correlations, suggesting higher sensitivity to editing flaws. Among all metrics, LPIPS aligns most closely with human-relevant perceptual qualities, whereas simple pixel-based metrics fail to reflect the semantic, structural, and perceptual dimensions central to instruction-based image editing.

\subsection{Metrics}
For comparison of traditional metrics and our MLLM-as-a-Judge metrics, we use the following:

Spearman’s ρ, and Kendall’s τ, Pearson’s r, Precision, Recall, F1, MSE, MAE, Pairwise Accuracy

\subsubsection{Ranked-based Correlation Metrics}
Spearman’s ρ, and Kendall’s τ, Pearson’s r, 
Precision, Recall, F1

\subsubsection{Error-based Metrics}
MSE, MAE, Accuracy

\subsubsection{Pairwise Accuracy}
Let there be a set of $n$ edited images (for specific edit type), and two scoring functions $f_A$ and $f_B$ (e.g., MLLM-as-a-Judge and a traditional metric like LPIPS). For every unordered pair of distinct images $(i, j)$, we determine whether both scoring functions agree on their relative ordering. The \textit{pairwise accuracy} between $A$ and $B$ is defined as the proportion of all item pairs for which the two systems produce consistent pairwise rankings. Formally,
\begin{equation}
\small
\frac{1}{N} \sum_{i<j} 
\mathbf{1}\!\left[
\operatorname{sign}\!\big(f_A(i) - f_A(j)\big)
=
\operatorname{sign}\!\big(f_B(i) - f_B(j)\big)
\right]
\end{equation}
where $N = \frac{n(n-1)}{2}$
is the total number of unique image pairs, and $\mathbf{1}[\cdot]$ denotes the indicator function, which equals 1 when the condition is true and 0 otherwise. Pairs where either $f(i) = f(j)$ are typically excluded to avoid ambiguity due to ties.

The resulting value lies in the range $[0, 1]$:
\begin{itemize}
    \item $1.0$ indicates perfect agreement in all pairwise comparisons,
    \item $0.5$ indicates random agreement (no correlation),
    \item $0.0$ represents complete disagreement (inverse ranking).
\end{itemize}

\subsection{Semantic Metrics}
\label{app: semantic metrics}
Across the online evaluation regime, semantic similarity metrics based on CLIP and DINO exhibit only limited explanatory power for our factorized judge scores. While \textbf{DINO Image} consistently emerges as the strongest among the traditional baselines—achieving the lowest predictive error and the highest correlation values (Table \ref{tab:trad_judge_evals}). \textbf{CLIP Image} also offers positive alignment but with noticeably weaker associations, and \textbf{CLIP Text} has a marginal positive alignment only with task counting, underscoring the inadequacy of text-conditioned embeddings for capturing structural integrity, identity coherence, or viewpoint realism in edited imagery. 

\subsection{Traditional Metrics Analysis}
\label{app:traditional metris analysis}
In the online setting, correlations between our 12 judge factors and traditional evaluation metrics remain uniformly weak, underscoring the inability of similarity-based measures to capture instruction-driven editing quality. Pixel-level metrics (L1, L2, PSNR, SSIM, LPIPS) show minimal alignment overall: L1 and L2 exhibit limited sensitivity to unintended pixel changes and spatial relationship errors, PSNR provides almost no signal, SSIM modestly reflects global coherence, and LPIPS aligns most closely with human-relevant texture and detail judgments due to its feature-level perceptual modeling (Table \ref{tab: detailed-online-pixel}). Semantic similarity metrics based on CLIP and DINO also demonstrate limited explanatory power. DINO Image performs best for content-preservation factors such as identity and global consistency, while CLIP Image is inconsistent and CLIP Text often correlates negatively with visual quality factors. These effects are amplified in the online regime, where correlations further degrade, particularly for instruction-fidelity dimensions (Table \ref{tab:clip_dino_gpt_gemini_instruction_fidelity_online}), highlighting a fundamental mismatch between generic similarity metrics and human-centered editing criteria.

\section{Human Study}
\label{sec: human-study}
\subsection{User Interface}
\begin{figure}[H]
    \centering
    \includegraphics[width=\linewidth]{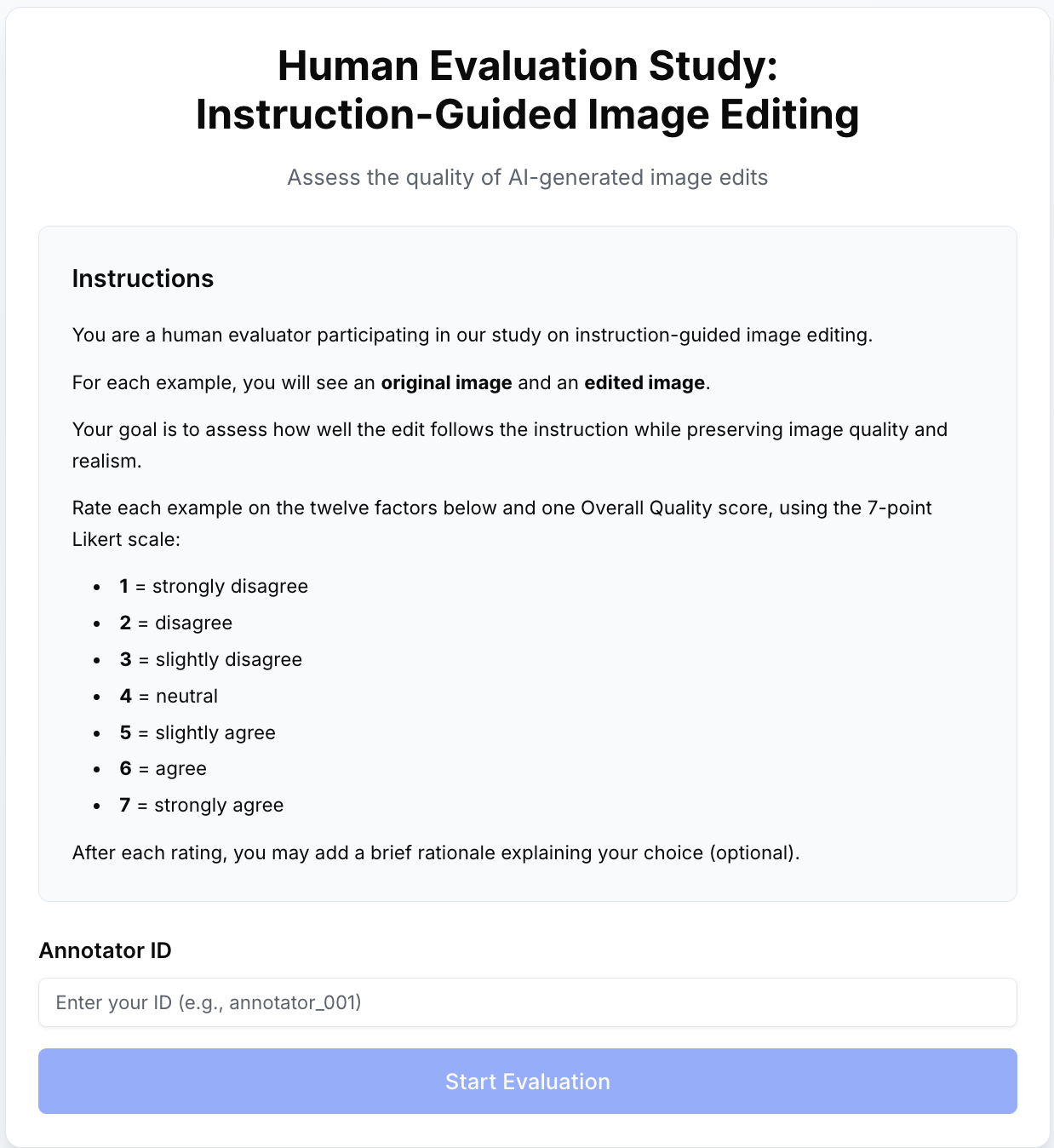}
    \caption{Human Evaluation Study UI for Instruction-Guided Image Editing}
    \label{fig:human-eval-ui-initial-screen}
\end{figure}

Refer to Figure\ref{fig:human-eval-ui-initial-screen}.

\subsection{Human Judge Evaluation Metrics}
Refer to Table~\ref{tab:human_judge_v1} and Table~\ref{tab:human_judge_combined_category}

\subsection{Evaluation Metrics}
\label{metrics}

\paragraph{Mean and Standard Deviation.}
For each factor $f$ and edit type $e$, we compute the mean score across all human evaluators:
\begin{equation}
\small
\mu_{f,e} = \frac{1}{N} \sum_{i=1}^{N} s_{i,f,e}
\end{equation}
where $s_{i,f,e}$ is the score given by evaluator $i$ for factor $f$ on edit type $e$, and $N$ is the number of evaluators. The standard deviation is:
\begin{equation}
\small
\sigma_{f,e} = \sqrt{\frac{1}{N-1} \sum_{i=1}^{N} (s_{i,f,e} - \mu_{f,e})^2}
\end{equation}

\paragraph{Intraclass Correlation Coefficient (ICC).}
We use ICC(2,k) (two-way random effects, average measures) to measure inter-rater reliability:
\begin{equation}
\small
\text{ICC}(2,k) = \frac{\text{MS}_R - \text{MS}_E}{\text{MS}_R + (k-1)\text{MS}_E + \frac{k}{n}(\text{MS}_C - \text{MS}_E)}
\end{equation}
where $\text{MS}_R$ is mean square for rows (images), $\text{MS}_E$ is mean square error, $\text{MS}_C$ is mean square for columns (raters), $k$ is the number of raters per image, and $n$ is the number of images.

\subsection{Human-MLLM Alignment Metrics}

\paragraph{Mean Squared Error (MSE).}
\begin{equation}
\small
\text{MSE} = \frac{1}{K} \sum_{k=1}^{K} (s_k^{\text{human}} - s_k^{\text{MLLM}})^2
\end{equation}
where $K$ is the number of evaluations, $s_k^{\text{human}}$ is the average human score for evaluation $k$, and $s_k^{\text{MLLM}}$ is the MLLM judge score.

\paragraph{Mean Absolute Error (MAE).}
\begin{equation}
\small
\text{MAE} = \frac{1}{K} \sum_{k=1}^{K} |s_k^{\text{human}} - s_k^{\text{MLLM}}|
\end{equation}

\paragraph{Pearson Correlation Coefficient.}
\begin{equation}
\small
\begin{aligned}
r =
\frac{\sum_{k=1}^{K}(h_k-\bar{h})(m_k-\bar{m})}
{\sqrt{\sum_{k=1}^{K}(h_k-\bar{h})^2}
\sqrt{\sum_{k=1}^{K}(m_k-\bar{m})^2}}
\end{aligned}
\end{equation}

where $h_k$ and $m_k$ denote human and MLLM scores, respectively.

\paragraph{Spearman's Rank Correlation ($\rho$).}
Spearman's $\rho$ is computed as the Pearson correlation coefficient applied to the rank-transformed scores.

\paragraph{Kendall's Tau ($\tau$).}
\begin{equation}
\small
\tau = \frac{n_c - n_d}{\frac{1}{2}K(K-1)}
\end{equation}
where $n_c$ is the number of concordant pairs and $n_d$ is the number of discordant pairs among all $\frac{1}{2}K(K-1)$ possible pairs of evaluations.

\subsection{Data Storage Format}
For reproducibility, we provide the exact data structure of our released benchmark dataset. Each evaluation record contains:

\begin{itemize}
    \item \texttt{participant\_id}: Anonymized evaluator identifier (string)
    \item \texttt{image\_id}: Unique image identifier (string)
    \item \texttt{edit\_type}: One of \{Add, Remove, Replace, Action, Counting, Relation\}
    \item \texttt{factor\_scores}: Dictionary mapping each of the 12 factors to scores (1-7)
    \item \texttt{overall\_score}: Overall quality rating (1-7)
    \item \texttt{timestamp\_start}: Evaluation start time (ISO 8601 format)
    \item \texttt{timestamp\_end}: Evaluation end time (ISO 8601 format)
    \item \texttt{annotator\_id}: Participant-provided identifier (string)
\end{itemize}

Data are provided in both CSV (flat format with one row per factor per evaluation) and JSONL (nested format with one JSON object per evaluation) for convenience.

\section{Human vs Judge vs Traditional Agreement Metrics}
\subsection{Pairwise and Pointwise results }
\label{sec:pair-point}

\begin{table*}[t]
\centering
\caption{
We compare three implementations of our fine-grained MLLM judges against human annotations across our 12 evaluation factors. 
Agreement is measured using error-based metrics (MSE, MAE; lower is better) and correlation-based metrics (Pearson, Spearman, Kendall's~$\tau$). For Pearson, Spearman and Kendall's~$\tau$, values larger or equal to 0.25 are bolded.
}
\label{tab:human_judge_pointwise}
\footnotesize
\resizebox{\textwidth}{!}{
\begin{tabular}{@{}l l c c c c c c c@{}}
\toprule
\textbf{Factor} & \textbf{Evaluator} & \textbf{MSE $\downarrow$} & \textbf{MAE $\downarrow$} & \textbf{ACC $\uparrow$} & \textbf{ACC$\pm$1 $\uparrow$} & \textbf{Pearson $\uparrow$} & \textbf{Spearman $\uparrow$} & \textbf{Kendall $\uparrow$} \\
\midrule
\multirow{3}{*}{\textcolor{GoogleGreen}{\textbf{Unchanged Regions}}} & \textbf{Main} (Fig.~\ref{fig:prompt-all-factors}) & 2.787 & 1.216 & 0.189 & 0.695 & 0.238 \textcolor{gray}{\tiny (0.020)} & 0.130 \textcolor{gray}{\tiny (0.210)} & 0.103 \textcolor{gray}{\tiny (0.226)} \\
 & \textbf{Factor-level Rubrics} (Fig.~\ref{fig:prompt-mllm-as-a-judge-image-editing, main}) & 3.050 & 1.258 & 0.168 & 0.695 & \textbf{0.357} \textcolor{gray}{\tiny (<0.001)} & 0.214 \textcolor{gray}{\tiny (0.037)} & 0.178 \textcolor{gray}{\tiny (0.036)} \\
 & \textbf{Category wise, Example guided} (Fig.~\ref{fig:prompt-mllm-as-a-judge-image-editing, category}) & 2.261 & 1.111 & 0.200 & 0.716 & \textbf{0.423} \textcolor{gray}{\tiny (<0.001)} & \textbf{0.384} \textcolor{gray}{\tiny (<0.001)} & \textbf{0.310} \textcolor{gray}{\tiny (<0.001)} \\
\cmidrule(l){1-9}
\multirow{3}{*}{\textcolor{GoogleGreen}{\textbf{Global Consistency}}} & \textbf{Main} (Fig.~\ref{fig:prompt-all-factors}) & 1.813 & 0.963 & 0.242 & 0.779 & 0.133 \textcolor{gray}{\tiny (0.200)} & 0.021 \textcolor{gray}{\tiny (0.843)} & 0.019 \textcolor{gray}{\tiny (0.832)} \\
 & \textbf{Factor-level Rubrics} (Fig.~\ref{fig:prompt-mllm-as-a-judge-image-editing, main}) & 2.171 & 1.100 & 0.168 & 0.737 & 0.188 \textcolor{gray}{\tiny (0.069)} & 0.134 \textcolor{gray}{\tiny (0.196)} & 0.118 \textcolor{gray}{\tiny (0.182)} \\
 & \textbf{Category wise, Example guided} (Fig.~\ref{fig:prompt-mllm-as-a-judge-image-editing, category}) & 2.003 & 0.995 & 0.221 & 0.811 & 0.162 \textcolor{gray}{\tiny (0.116)} & 0.191 \textcolor{gray}{\tiny (0.064)} & 0.165 \textcolor{gray}{\tiny (0.055)} \\
\cmidrule(l){1-9}
\multirow{3}{*}{\textcolor{GoogleGreen}{\textbf{Identity Preservation}}} & \textbf{Main} (Fig.~\ref{fig:prompt-all-factors}) & 2.987 & 1.153 & 0.326 & 0.674 & -0.067 \textcolor{gray}{\tiny (0.517)} & 0.006 \textcolor{gray}{\tiny (0.956)} & 0.001 \textcolor{gray}{\tiny (0.988)} \\
 & \textbf{Factor-level Rubrics} (Fig.~\ref{fig:prompt-mllm-as-a-judge-image-editing, main}) & 3.345 & 1.289 & 0.242 & 0.653 & -0.012 \textcolor{gray}{\tiny (0.911)} & -0.049 \textcolor{gray}{\tiny (0.639)} & -0.040 \textcolor{gray}{\tiny (0.649)} \\
 & \textbf{Category wise, Example guided} (Fig.~\ref{fig:prompt-mllm-as-a-judge-image-editing, category}) & 2.713 & 1.163 & 0.242 & 0.705 & 0.145 \textcolor{gray}{\tiny (0.160)} & 0.240 \textcolor{gray}{\tiny (0.019)} & 0.201 \textcolor{gray}{\tiny (0.022)} \\
\cmidrule(l){1-9}
\multirow{3}{*}{\textcolor{GoogleBlue}{\textbf{Scale Realism}}} & \textbf{Main} (Fig.~\ref{fig:prompt-all-factors}) & 1.155 & 0.679 & 0.421 & 0.874 & 0.225 \textcolor{gray}{\tiny (0.028)} & \textbf{0.264} \textcolor{gray}{\tiny (0.010)} & 0.237 \textcolor{gray}{\tiny (0.010)} \\
 & \textbf{Factor-level Rubrics} (Fig.~\ref{fig:prompt-mllm-as-a-judge-image-editing, main}) & 1.334 & 0.679 & 0.495 & 0.811 & 0.047 \textcolor{gray}{\tiny (0.648)} & 0.030 \textcolor{gray}{\tiny (0.769)} & 0.028 \textcolor{gray}{\tiny (0.763)} \\
 & \textbf{Category wise, Example guided} (Fig.~\ref{fig:prompt-mllm-as-a-judge-image-editing, category}) & 0.924 & 0.626 & 0.421 & 0.905 & \textbf{0.337} \textcolor{gray}{\tiny (<0.001)} & \textbf{0.300} \textcolor{gray}{\tiny (0.003)} & \textbf{0.269} \textcolor{gray}{\tiny (0.004)} \\
\cmidrule(l){1-9}
\multirow{3}{*}{\textcolor{GoogleBlue}{\textbf{Spatial Relationship}}} & \textbf{Main} (Fig.~\ref{fig:prompt-all-factors}) & 1.047 & 0.684 & 0.389 & 0.874 & \textbf{0.355} \textcolor{gray}{\tiny (<0.001)} & 0.236 \textcolor{gray}{\tiny (0.021)} & 0.211 \textcolor{gray}{\tiny (0.019)} \\
 & \textbf{Factor-level Rubrics} (Fig.~\ref{fig:prompt-mllm-as-a-judge-image-editing, main}) & 1.226 & 0.695 & 0.421 & 0.853 & \textbf{0.335} \textcolor{gray}{\tiny (<0.001)} & \textbf{0.261} \textcolor{gray}{\tiny (0.011)} & 0.235 \textcolor{gray}{\tiny (0.010)} \\
 & \textbf{Category wise, Example guided} (Fig.~\ref{fig:prompt-mllm-as-a-judge-image-editing, category}) & 1.847 & 0.905 & 0.316 & 0.832 & 0.073 \textcolor{gray}{\tiny (0.484)} & 0.105 \textcolor{gray}{\tiny (0.312)} & 0.094 \textcolor{gray}{\tiny (0.295)} \\
\cmidrule(l){1-9}
\multirow{3}{*}{\textcolor{GoogleBlue}{\textbf{Texture and Detail}}} & \textbf{Main} (Fig.~\ref{fig:prompt-all-factors}) & 1.292 & 0.816 & 0.274 & 0.832 & -0.044 \textcolor{gray}{\tiny (0.674)} & -0.036 \textcolor{gray}{\tiny (0.726)} & -0.030 \textcolor{gray}{\tiny (0.732)} \\
 & \textbf{Factor-level Rubrics} (Fig.~\ref{fig:prompt-mllm-as-a-judge-image-editing, main}) & 1.345 & 0.868 & 0.221 & 0.832 & 0.081 \textcolor{gray}{\tiny (0.435)} & -0.012 \textcolor{gray}{\tiny (0.907)} & -0.007 \textcolor{gray}{\tiny (0.935)} \\
 & \textbf{Category wise, Example guided} (Fig.~\ref{fig:prompt-mllm-as-a-judge-image-editing, category}) & 1.450 & 0.953 & 0.200 & 0.747 & 0.094 \textcolor{gray}{\tiny (0.363)} & 0.029 \textcolor{gray}{\tiny (0.779)} & 0.026 \textcolor{gray}{\tiny (0.764)} \\
\cmidrule(l){1-9}
\multirow{3}{*}{\textcolor{GoogleBlue}{\textbf{Image Quality}}} & \textbf{Main} (Fig.~\ref{fig:prompt-all-factors}) & 0.871 & 0.668 & 0.326 & 0.895 & 0.129 \textcolor{gray}{\tiny (0.211)} & 0.123 \textcolor{gray}{\tiny (0.235)} & 0.111 \textcolor{gray}{\tiny (0.223)} \\
 & \textbf{Factor-level Rubrics} (Fig.~\ref{fig:prompt-mllm-as-a-judge-image-editing, main}) & 1.018 & 0.700 & 0.326 & 0.884 & 0.201 \textcolor{gray}{\tiny (0.051)} & 0.090 \textcolor{gray}{\tiny (0.384)} & 0.082 \textcolor{gray}{\tiny (0.371)} \\
 & \textbf{Category wise, Example guided} (Fig.~\ref{fig:prompt-mllm-as-a-judge-image-editing, category}) & 0.934 & 0.732 & 0.284 & 0.863 & 0.164 \textcolor{gray}{\tiny (0.113)} & 0.093 \textcolor{gray}{\tiny (0.369)} & 0.082 \textcolor{gray}{\tiny (0.363)} \\
\cmidrule(l){1-9}
\multirow{3}{*}{\textcolor{GoogleBlue}{\textbf{Color and Lighting}}} & \textbf{Main} (Fig.~\ref{fig:prompt-all-factors}) & 1.655 & 0.911 & 0.295 & 0.779 & 0.058 \textcolor{gray}{\tiny (0.577)} & 0.079 \textcolor{gray}{\tiny (0.446)} & 0.069 \textcolor{gray}{\tiny (0.430)} \\
 & \textbf{Factor-level Rubrics} (Fig.~\ref{fig:prompt-mllm-as-a-judge-image-editing, main}) & 1.571 & 0.932 & 0.242 & 0.789 & \textbf{0.291} \textcolor{gray}{\tiny (0.004)} & 0.224 \textcolor{gray}{\tiny (0.029)} & 0.193 \textcolor{gray}{\tiny (0.031)} \\
 & \textbf{Category wise, Example guided} (Fig.~\ref{fig:prompt-mllm-as-a-judge-image-editing, category}) & 1.897 & 0.995 & 0.284 & 0.726 & 0.131 \textcolor{gray}{\tiny (0.205)} & 0.137 \textcolor{gray}{\tiny (0.187)} & 0.115 \textcolor{gray}{\tiny (0.174)} \\
\cmidrule(l){1-9}
\multirow{3}{*}{\textcolor{GoogleBlue}{\textbf{Seamlessness}}} & \textbf{Main} (Fig.~\ref{fig:prompt-all-factors}) & 1.239 & 0.774 & 0.295 & 0.832 & 0.202 \textcolor{gray}{\tiny (0.049)} & 0.222 \textcolor{gray}{\tiny (0.030)} & 0.191 \textcolor{gray}{\tiny (0.029)} \\
 & \textbf{Factor-level Rubrics} (Fig.~\ref{fig:prompt-mllm-as-a-judge-image-editing, main}) & 1.566 & 0.879 & 0.253 & 0.789 & 0.167 \textcolor{gray}{\tiny (0.106)} & 0.083 \textcolor{gray}{\tiny (0.424)} & 0.070 \textcolor{gray}{\tiny (0.420)} \\
 & \textbf{Category wise, Example guided} (Fig.~\ref{fig:prompt-mllm-as-a-judge-image-editing, category}) & 2.218 & 1.100 & 0.200 & 0.695 & 0.177 \textcolor{gray}{\tiny (0.086)} & 0.205 \textcolor{gray}{\tiny (0.046)} & 0.168 \textcolor{gray}{\tiny (0.048)} \\
\cmidrule(l){1-9}
\multirow{3}{*}{\textcolor{GoogleRed}{\textbf{Alignment}}} & \textbf{Main} (Fig.~\ref{fig:prompt-all-factors}) & 2.237 & 0.979 & 0.379 & 0.737 & \textbf{0.459} \textcolor{gray}{\tiny (<0.001)} & \textbf{0.312} \textcolor{gray}{\tiny (0.002)} & \textbf{0.272} \textcolor{gray}{\tiny (0.002)} \\
 & \textbf{Factor-level Rubrics} (Fig.~\ref{fig:prompt-mllm-as-a-judge-image-editing, main}) & 2.532 & 1.032 & 0.379 & 0.716 & \textbf{0.377} \textcolor{gray}{\tiny (<0.001)} & \textbf{0.275} \textcolor{gray}{\tiny (0.007)} & 0.243 \textcolor{gray}{\tiny (0.007)} \\
 & \textbf{Category wise, Example guided} (Fig.~\ref{fig:prompt-mllm-as-a-judge-image-editing, category}) & 1.858 & 0.884 & 0.400 & 0.758 & \textbf{0.593} \textcolor{gray}{\tiny (<0.001)} & \textbf{0.483} \textcolor{gray}{\tiny (<0.001)} & \textbf{0.422} \textcolor{gray}{\tiny (<0.001)} \\
\cmidrule(l){1-9}
\multirow{3}{*}{\textcolor{GoogleRed}{\textbf{Completeness}}} & \textbf{Main} (Fig.~\ref{fig:prompt-all-factors}) & 2.147 & 0.947 & 0.389 & 0.758 & \textbf{0.464} \textcolor{gray}{\tiny (<0.001)} & \textbf{0.297} \textcolor{gray}{\tiny (0.003)} & \textbf{0.258} \textcolor{gray}{\tiny (0.004)} \\
 & \textbf{Factor-level Rubrics} (Fig.~\ref{fig:prompt-mllm-as-a-judge-image-editing, main}) & 2.221 & 0.905 & 0.421 & 0.789 & \textbf{0.434} \textcolor{gray}{\tiny (<0.001)} & \textbf{0.304} \textcolor{gray}{\tiny (0.003)} & \textbf{0.273} \textcolor{gray}{\tiny (0.003)} \\
 & \textbf{Category wise, Example guided} (Fig.~\ref{fig:prompt-mllm-as-a-judge-image-editing, category}) & 1.895 & 0.832 & 0.442 & 0.800 & \textbf{0.574} \textcolor{gray}{\tiny (<0.001)} & \textbf{0.392} \textcolor{gray}{\tiny (<0.001)} & \textbf{0.348} \textcolor{gray}{\tiny (<0.001)} \\
\cmidrule(l){1-9}
\multirow{3}{*}{\textcolor{GoogleRed}{\textbf{Plausibility}}} & \textbf{Main} (Fig.~\ref{fig:prompt-all-factors}) & 1.774 & 0.800 & 0.432 & 0.779 & \textbf{0.278} \textcolor{gray}{\tiny (0.006)} & 0.165 \textcolor{gray}{\tiny (0.111)} & 0.147 \textcolor{gray}{\tiny (0.107)} \\
 & \textbf{Factor-level Rubrics} (Fig.~\ref{fig:prompt-mllm-as-a-judge-image-editing, main}) & 1.932 & 0.842 & 0.432 & 0.747 & \textbf{0.256} \textcolor{gray}{\tiny (0.012)} & 0.168 \textcolor{gray}{\tiny (0.104)} & 0.149 \textcolor{gray}{\tiny (0.101)} \\
 & \textbf{Category wise, Example guided} (Fig.~\ref{fig:prompt-mllm-as-a-judge-image-editing, category}) & 1.953 & 0.853 & 0.421 & 0.758 & \textbf{0.288} \textcolor{gray}{\tiny (0.005)} & 0.134 \textcolor{gray}{\tiny (0.194)} & 0.120 \textcolor{gray}{\tiny (0.190)} \\
\midrule
\multirow{3}{*}{\textbf{All}} & \textbf{Main} (Fig.~\ref{fig:prompt-all-factors}) & 1.750 & 0.882 & 0.330 & 0.792 & 0.249 \textcolor{gray}{\tiny (<0.001)} & 0.206 \textcolor{gray}{\tiny (<0.001)} & 0.177 \textcolor{gray}{\tiny (<0.001)} \\
 & \textbf{Factor-level Rubrics} (Fig.~\ref{fig:prompt-mllm-as-a-judge-image-editing, main}) & 1.943 & 0.932 & 0.314 & 0.775 & \textbf{0.275} \textcolor{gray}{\tiny (<0.001)} & 0.195 \textcolor{gray}{\tiny (<0.001)} & 0.170 \textcolor{gray}{\tiny (<0.001)} \\
 & \textbf{Category wise, Example guided} (Fig.~\ref{fig:prompt-mllm-as-a-judge-image-editing, category}) & 1.829 & 0.929 & 0.303 & 0.776 & \textbf{0.315} \textcolor{gray}{\tiny (<0.001)} & \textbf{0.267} \textcolor{gray}{\tiny (<0.001)} & 0.224 \textcolor{gray}{\tiny (<0.001)} \\
\bottomrule
\end{tabular}
}
\end{table*}

\begin{table*}[t]
\centering
\caption{Accuracy in predicting pairwise human preferences over different prompts. The MLLM-as-a-Judge Model selected here is GPT-5-mini. for all factors and prompts. The last column shows
the overall accuracy, weighted average over all factors. The highest accuracy, for both individual factors and overall, is reported in bold. For the name mapping, the order of the factors are the same with the order in Table~\ref{tab:human_judge_pointwise}. For this table, we only select image edit pairs with human evaluation score difference larger than 2 to calculate pairwise acc.}
\label{tab:human_judge_pairwise}
\footnotesize
\resizebox{\textwidth}{!}{
\begin{tabular}{@{}l ccccccccccccc@{}}
\toprule
\textbf{Evaluator} & \textcolor{GoogleGreen}{\textbf{UR}} & \textcolor{GoogleGreen}{\textbf{GC}} & \textcolor{GoogleGreen}{\textbf{IP}} & \textcolor{GoogleBlue}{\textbf{SR}} & \textcolor{GoogleBlue}{\textbf{SP}} & \textcolor{GoogleBlue}{\textbf{TD}} & \textcolor{GoogleBlue}{\textbf{IQ}} & \textcolor{GoogleBlue}{\textbf{CL}} & \textcolor{GoogleBlue}{\textbf{SM}} & \textcolor{GoogleRed}{\textbf{AL}} & \textcolor{GoogleRed}{\textbf{CP}} & \textcolor{GoogleRed}{\textbf{PL}} & \textbf{All} \\
\midrule
\textbf{Main} (Fig.~\ref{fig:prompt-all-factors}) & 0.45 & 0.47 & 0.26 & 0.21 & \textbf{0.66} & 0.39 & \textbf{0.44} & 0.45 & \textbf{0.48} & 0.47 & 0.58 & \textbf{0.35} & 0.44 \\
\textbf{Factor-level Rubrics} (Fig.~\ref{fig:prompt-mllm-as-a-judge-image-editing, main}) & 0.52 & \textbf{0.55} & 0.16 & 0.00 & 0.66 & \textbf{0.50} & 0.35 & \textbf{0.56} & 0.48 & 0.36 & 0.41 & 0.27 & 0.40 \\
\textbf{Category wise, Example guided} (Fig.~\ref{fig:prompt-mllm-as-a-judge-image-editing, category}) & \textbf{0.64} & 0.43 & \textbf{0.50} & \textbf{0.70} & 0.31 & 0.40 & 0.40 & 0.37 & 0.39 & \textbf{0.67} & \textbf{0.67} & 0.28 & \textbf{0.52} \\
\bottomrule
\end{tabular}
}
\end{table*}

Table~\ref{tab:human_judge_pointwise} provides detailed pointwise and pairwise agreement metrics between human evaluators and the MLLM judge across image edit types. We report error-based metrics (MSE, MAE), ACC, ACC$\pm$1, and rank correlation measures (Pearson, Spearman, Kendall’s $\tau$), and derived pairwise preference agreement statistics. For ACC, it is defined that the score for our judge and human evaluation must match each other. For ACC$\pm$1 it means the absolute difference between human evaluation and our judge is within 1.

Table~\ref{tab:human_judge_pairwise} derived pairwise preference agreement. For each factor. Pairwise accuracy measures how often the judge recovers human preferences between image pairs.

\begin{table*}[t]
\centering
\caption{
\textbf{Traditional metrics fail to match human evaluation, while the MLLM judge aligns closely across edit types.}
We report normalized mean $\pm$ standard deviation scores for human judgments, the MLLM judge, and common traditional image evaluation metrics across six instruction-guided image edit types (Add, Remove, Replace, Action, Counting, Relation) and their aggregate (All Edits). The prompt used for MLLM Judge is Main (Figure~\ref{fig:prompt-all-factors}).
Pixel-level error metrics (L1, L2) and structural similarity measures (SSIM, Mask SSIM, PSNR) show consistently weak alignment with human judgments, while feature-based semantic metrics (CLIP Image, CLIP Text, DINO) capture only partial agreement.
In contrast, the MLLM judge closely tracks human scores across all edit types, highlighting its ability to assess instruction fidelity and semantic correctness beyond visual similarity.
Higher values indicate stronger agreement for similarity-based metrics, while lower values indicate better performance for error-based metrics (L1, L2).
Shading denotes closeness to human judgments, with darker green indicating stronger alignment.
}

\label{tab:trad_judge_evals}
\small
\footnotesize
\begin{tabular}{lccccccc}
\toprule
Metric & Add & Remove & Replace & Action & Counting & Relation & All Edits \\
\midrule
\textbf{Human Avg} & \score{0.786}{0.02} & \score{0.856}{0.02} & \score{0.801}{0.03} & \score{0.783}{0.07} & \score{0.668}{0.11} & \score{0.794}{0.04} & \score{0.781}{0.08} \\
\textbf{Judge Avg} & \cellcolor{green!30}\score{0.799}{0.07} & \cellcolor{green!10}\score{0.798}{0.07} & \cellcolor{green!30}\score{0.779}{0.09} & \cellcolor{green!10}\score{0.801}{0.09} & \cellcolor{green!30}\score{0.731}{0.08} & \cellcolor{green!30}\score{0.804}{0.04} & \cellcolor{green!30}\score{0.785}{0.08} \\
\midrule
Background Consistency & \score{0.499}{0.27} & \score{0.500}{0.22} & \score{0.494}{0.22} & \score{0.473}{0.28} & \score{0.554}{0.28} & \score{0.504}{0.27} & \score{0.499}{0.24} \\
Clip Image Norm & \cellcolor{green!10} \score{0.893}{0.04} & \cellcolor{green!10} \score{0.915}{0.04} & \score{0.906}{0.04} & \score{0.914}{0.03} & \score{0.925}{0.03} & \score{0.922}{0.03} & \score{0.912}{0.04} \\
Clip Text Norm & \score{0.612}{0.02} & \score{0.571}{0.03} & \score{0.605}{0.03} & \score{0.619}{0.02} & \cellcolor{green!10} \score{0.619}{0.02} & \score{0.630}{0.02} & \score{0.593}{0.03} \\
Dino Image Norm & \score{0.671}{0.25} & \score{0.794}{0.20} & \cellcolor{green!10}\score{0.755}{0.21} & \cellcolor{green!30} \score{0.808}{0.30} & \score{0.809}{0.07} & \cellcolor{green!30} \score{0.805}{0.15} & \cellcolor{green!30} \score{0.778}{0.19} \\
L1 Error & \score{0.390}{0.24} & \score{0.410}{0.20} & \score{0.345}{0.18} & \score{0.545}{0.28} & \score{0.382}{0.27} & \score{0.380}{0.21} & \score{0.402}{0.22} \\
L2 Error & \score{0.323}{0.24} & \score{0.294}{0.22} & \score{0.234}{0.18} & \score{0.412}{0.23} & \score{0.294}{0.26} & \score{0.332}{0.23} & \score{0.296}{0.22} \\
Lpips & \score{0.550}{0.26} & \score{0.498}{0.19} & \score{0.525}{0.20} & \score{0.404}{0.23} & \score{0.526}{0.28} & \score{0.475}{0.20} & \score{0.500}{0.21} \\
Mask Lpips & \score{0.550}{0.26} & \score{0.498}{0.19} & \score{0.525}{0.20} & \score{0.404}{0.23} & \score{0.526}{0.28} & \score{0.475}{0.20} & \score{0.500}{0.21} \\
Mask Ssim & \score{0.499}{0.27} & \score{0.500}{0.22} & \score{0.494}{0.22} & \score{0.473}{0.28} & \score{0.554}{0.28} & \score{0.504}{0.27} & \score{0.499}{0.24} \\
Psnr & \score{0.485}{0.27} & \score{0.378}{0.21} & \score{0.499}{0.20} & \score{0.411}{0.23} & \score{0.515}{0.28} & \score{0.432}{0.24} & \score{0.429}{0.23} \\
Ssim & \score{0.499}{0.27} & \score{0.500}{0.22} & \score{0.494}{0.22} & \score{0.473}{0.28} & \score{0.554}{0.28} & \score{0.504}{0.27} & \score{0.499}{0.24} \\
\bottomrule
\end{tabular}
\end{table*}

\subsection{Key Findings}
For this analysis, we refer to Table~\ref{tab:trad_judge_evals}.

\paragraph{MLLM judge exhibits strongest alignment with human evaluations.}
Across all edit types, the MLLM judge achieves an average score of
$0.785 \pm 0.08$, closely matching and slightly exceeding the human average
($0.781 \pm 0.08$). This level of agreement is consistently higher than that of
any traditional evaluation metric, indicating that the judge captures
human-aligned notions of instruction fidelity and semantic correctness that are
not accessible to low-level similarity measures.

\paragraph{Pixel-level similarity metrics fail to reflect human judgment.}
Error-based and structural metrics such as L1 ($0.402 \pm 0.22$), L2
($0.296 \pm 0.22$), SSIM ($0.499 \pm 0.24$), Mask SSIM ($0.499 \pm 0.24$), and
PSNR ($0.429 \pm 0.23$) exhibit weak alignment with human scores across all edit
types. These metrics primarily assess reconstruction fidelity and are largely
insensitive to whether the semantic intent of the instruction has been satisfied,
leading to poor correspondence with human evaluation.

\paragraph{Semantic similarity metrics capture partial alignment.}
Feature-based metrics show improved performance relative to pixel-level measures.
CLIP Image Norm ($0.912 \pm 0.04$), CLIP Text Norm ($0.593 \pm 0.03$), and DINO
Image Norm ($0.778 \pm 0.19$) demonstrate moderate alignment with human judgments,
suggesting that semantic representations encode some aspects of instruction
compliance. However, these metrics remain insufficient for reliably assessing
complex edits, as they do not explicitly model edit intent, instruction grounding,
or fine-grained relational correctness.

\paragraph{Consistency across edit types.}
Traditional metrics exhibit substantial variability across edit categories,
particularly degrading on Action, Counting, and Relation edits. In contrast, the
MLLM judge maintains comparatively stable performance across edit types, with
performance drops primarily observed for inherently ambiguous cases such as
Counting, where even human agreement is lower.

\paragraph{Implications.}
Overall, these results demonstrate that MLLM-based judges provide a more
human-aligned evaluation signal than traditional metrics. While pixel-level and
semantic similarity measures capture aspects of visual fidelity and content
overlap, they fail to assess instruction satisfaction. By explicitly reasoning
over intent, semantics, and contextual consistency, the MLLM judge consistently
outperforms all traditional metrics in similarity to human evaluation.

\subsection{Error Metrics for All Prompts}
\label{sec:error-metrics-prompts}

Table~\ref{tab:human_judge_pointwise} reports a meta-evaluation of three MLLM-as-a-judge prompting strategies by comparing their agreement with human annotations across 12 fine-grained evaluation factors spanning Image Preservation, Edit Quality, and Instruction Fidelity. We report both error-based metrics (MSE, MAE; lower is better) and correlation-based metrics (Pearson, Spearman, Kendall’s $\tau$; higher is better), enabling a comprehensive assessment of how different prompting designs align with human judgment.

In contrast, Main consistently outperforms Factor-level Rubrics across most factors in Table~\ref{tab:human_judge_pointwise}, exhibiting lower error (MSE and MAE) and stronger correlations with human judgments for a broad range of evaluation dimensions. In particular, Main achieves higher Pearson, Spearman, and Kendall correlations for all Instruction Fidelity factors, including Alignment, Completeness, and Plausibility, indicating a closer match to human relative preferences. Meanwhile, although Factor-level Rubrics introduces explicit scoring rules for each factor, its more constrained evaluation scheme appears to limit alignment with human annotations, resulting in weaker correlations and higher variance in many cases. 

In contrast, Main generally exhibits lower correlation and higher error, indicating that jointly evaluating all factors within a single prompt can obscure nuanced distinctions between evaluation dimensions. Factor-level Rubrics shows intermediate performance, often improving upon Main while remaining slightly less correlated with human judgments than Category wise, Example guided prompt. This trend suggests that while a unified prompt offers practical advantages in consistency and simplicity, it may trade off some fine-grained alignment relative to factor-specific prompting.

Performance also varies systematically across evaluation categories. All judges demonstrate stronger agreement with humans on semantic related factors such as Alignment, Completeness, and Plausibility, while exhibiting weaker alignment on Image Preservation and Edit Quality factors.

Overall, these results indicate that prompt design plays a critical role in MLLM-based evaluation. While Category wise, Example guided prompting yields the highest alignment with human judgments, the Main evaluator used in this work provides a competitive and scalable alternative, balancing evaluation quality with practical deployment considerations.

\begin{figure}[h!]
    \centering
    \includegraphics[width=0.9\linewidth]{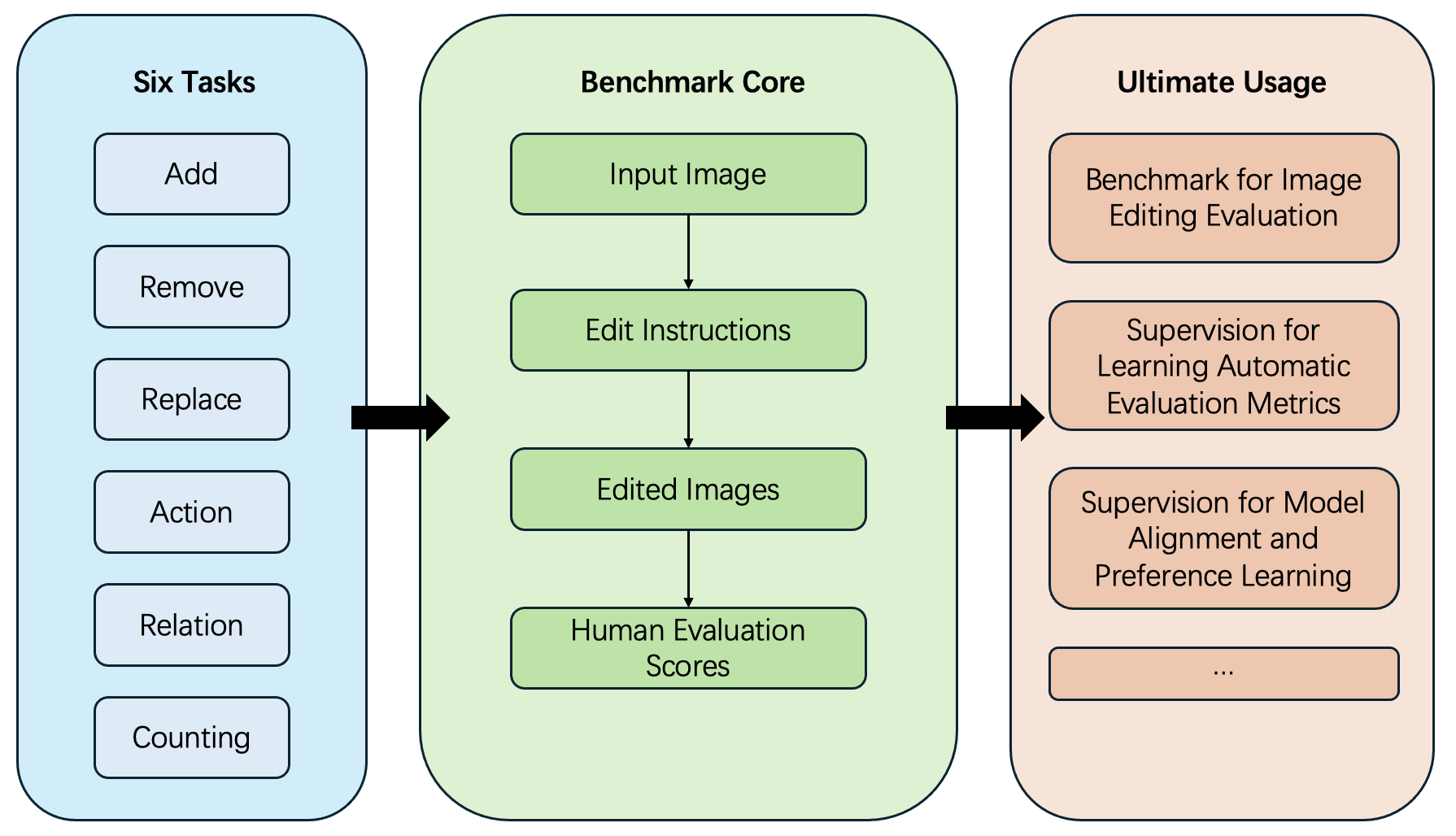}
    \caption{Structure of our benchmark and usage examples.}
    \label{fig:benchmark_flowchart}
\end{figure}

\section{Additional Benchmark Details}
\label{sec:appendix-benchmark}
In this section, we provide detailed description and discussion of our proposed benchmark.

\subsection{Benchmark Construction}
\label{app:benchmark_construction}
We construct our benchmark based on the HumanEdit dataset. Specifically, for each input image and its corresponding edit instruction in HumanEdit, we use GPT-image to generate a new edited image. The original edited image provided by HumanEdit is treated as the ground-truth image. In the online setting, we evaluate the tuple $(\text{input image}, \text{edit instruction}, \text{edited image})$ by collecting human ratings across our designed 12 judging factors, and we additionally obtain scores from both MLLM-as-a-Judge and traditional image quality metrics. In the offline setting, we evaluate the tuple 
\\
$(\text{edited image}, \text{edit instruction}, \text{ground truth image})$ using MLLM-as-a-Judge and traditional metrics, without human involvement.

\subsection{Image Preparation}
All images were standardized to consistent resolution and format to ensure uniform presentation quality across evaluations. Original and edited images were presented at identical dimensions to facilitate direct visual comparison.

\subsection{Benchmark Tasks}
\label{app:benchmark_tasks}
Constructed on HumanEdit, our benchmark covers six image editing tasks, including Add, Remove, Replace, Action, Relation, and Counting. The detailed definition for each edit task is listed below:
\begin{itemize}
\setlength{\itemsep}{2pt}
\setlength{\parskip}{0pt}
\setlength{\parsep}{0pt}
    \item \textbf{Add}: Add an object to the original image.
    \item \textbf{Remove}: Removing certain objects from an image, typically those that are more prominent or easily distinguishable.
    \item \textbf{Replace}: Modify the type of an object, change a part of an object, or alter its shape.
    \item \textbf{Counting}: Alter the number of objects in the image.
    \item \textbf{Action}: Alter the subject's action if the subject is a specific organism.
    \item \textbf{Relation}: Modifying the relationships between objects.
\end{itemize}

\subsection{Benchmark Schema}
\label{app:benchmark_schema}
 For each image editing task, our benchmark contains multiple image editing instances, each consisting of an input image, an edit instruction, and the corresponding edited image generated by GPT-image. For every edited result, the benchmark provides a comprehensive set of evaluations, including traditional metric scores, five human evaluation scores, and scores produced by our MLLM-based judge, along with concise textual justifications explaining each judgment. This rich and fine-grained annotation enables systematic analysis of image editing quality across diverse tasks and evaluation perspectives.

\subsection{Use Cases}
\label{app:benchmark_use_cases}
The dataset serves as a benchmark for evaluating image editing models using human judgments as ground truth. It enables direct assessment of instruction following, semantic correctness, and contextual consistency, particularly in semantic editing scenarios where traditional metrics such as PSNR, SSIM, and LPIPS fail to reflect true edit quality.

The human ratings further provide supervision for learning and validating automatic evaluation metrics, including MLLM-based judges. This allows systematic analysis of metric–human alignment and supports the identification of key evaluation factors that drive accurate assessment of image editing quality.

Beyond evaluation, the dataset can be leveraged for model alignment with human preferences. Human scores or preference signals can be used to train reward models or support preference-based learning, encouraging image editing models to produce outputs that better match human expectations and editing intent.

The inclusion of complete editing context—original image, instruction, edited result, and human score—also enables fine-grained error analysis. This facilitates the study of failure modes related to instruction understanding, semantic consistency, and unintended visual artifacts.

Finally, the dataset supports systematic analysis of the gap between automatic metrics and human judgment. By comparing human ratings with metric scores, it reveals cases where existing evaluation methods overestimate or underestimate edit quality, motivating the need for more human-aligned, reasoning-based evaluation frameworks.

\subsection{How to Use}
Access to the full set of prompt templates, dataset loaders and evaluation code is available at our GitHub repository:  
\href{https://github.com/mllmasajudge-anonymous/MLLM-as-a-Judge}
{\nolinkurl{https://github.com/mllmasajudge-anonymous/MLLM-as-a-Judge}}.

\paragraph{}

\section{Prompts}
\label{prompts-sec}
For all 12 factors, the prompt requires the Judge to assign a discrete score between 1 and 7 for each factor corresponding to the input image. In addition, for every judge factor, we supply textual examples illustrating scores of 1, 4, and 7 to help the Judge make more accurate quality assessments.

To examine whether different prompt formats influence the Judge’s outputs, we designed the following three formats:
\begin{itemize}
    \setlength{\itemsep}{2pt}
    \setlength{\parskip}{0pt}
    \setlength{\parsep}{0pt}
    
    \item \textbf{{Generate all factor results at once}}: The prompt includes a single sentence definition for each factor. Each judging run produces scores for all factors at once. Refer to Figure \ref{fig:prompt-all-factors} for more detail on the prompt's structure. 
    \item \textbf{Produce results separately for each major category of factors}: For the three major categories Image Preservation, Edit Quality, and Instruction Fidelity, we generate one prompt for each category, covering all the factors within that category. Each prompt includes examples of when to score high or low for a given factor (Figure \ref{fig:prompt-mllm-as-a-judge-image-editing, category})
    \item \textbf{Produce results for all factors using a scoring rubric}: This prompt is a combination of the first two version. It generates all factors at once but also includes a scoring rubric, as illustrated in Table \ref{tab:judge_rubric}. Refer to Figure \ref{fig:prompt-mllm-as-a-judge-image-editing, main} for a more detailed illustration of the prompt's structure.
\end{itemize}

After having human evaluators try three prompt formats, we found that the first format did not provide enough information, as no scoring rubric was provided, and that the second format was too verbose - it required reading excessive amounts of text. Therefore, we ultimately chose to use the third prompt format, as it is a combination of the first two.

\subsection{Generic Framework for all factors + offline setting}
We adopt a general prompt as the input to the MLLM-as-a-Judge, which is designed to be applicable to both online and offline evaluation settings. The core structure and instructions of the prompt remain identical across settings, with only minor adjustments to the image input format to accommodate the corresponding inference pipelines. We present below the three prompt templates explored under the offline setting.

\begin{figure}[H]
\vspace{-9pt}
  \centering
  \begin{promptbox}{\small Main Prompt}
    
    \begin{tiny}
\textbf{ROLE}: You are an expert image editing evaluator. Your evaluations must be objective, consistent, and grounded entirely in visual comparison and task intent. \\

\textbf{CONTEXT:} You are provided with three inputs:
\begin{compactenum}
    \item Input Image – the unedited image.
 
    {\quad\quad\quad\centering{{\bfseries\ttfamily [input image]}}}

   \item Edited Image – the image produced after editing.
   
   {\quad\quad\quad\centering{{\bfseries\ttfamily [edited image]}}}
   \item Edit Instruction – a natural language description of the intended modification.
   
   {\quad\quad\quad\centering{{\bfseries\ttfamily [text instruction]}}}
\end{compactenum}

Your task is to evaluate how well the Edited Image aligns with the Input Image according to the Edit Instruction. \\

\textbf{FACTORS:} 
\begin{compactenum}
    \item Unchanged Regions: Did the parts of the image that were not supposed to be edited remain unchanged?
    \item Global Consistency: Has the overall appearance (style, layout, and color) been preserved?  
    \item Identity Preservation: Do people, animals, or objects maintain their original identity and features after the edit?
    \item Scale Realism: Is the scale of the edited object realistic compared to other objects in the image?
    \item Spatial Relationship: Has the spatial relationship between objects been maintained?
    \item Texture and Detail: Is the texture and detail in the edited region consistent with the surrounding areas?
    \item Image Quality: Does the edited image avoid noise, blurring, or unnatural distortions?
    \item Color and Lighting: Do the colors, shadows, and lighting of the edited region match the rest of the image?
    \item Seamlessness: Does the transition between edited and non-edited regions look natural? 
    \item Alignment: Does the edited image align with the specific edits provided in the instructions?
    \item Completeness: Were all aspects of the instruction carried out fully?
    \item Plausibility: Does the result make sense in a real-world context?
\end{compactenum}

\textbf{EVALUATION STEPS:}
\begin{compactenum}
    \item Compare the Edited Image to the Ground Truth Image in the context of the Edit Instruction.
    \item Assess how well the edited image satisfies each factor definition.
    \item Assign a score between 1 and 7 (integers only) using the rubric above.
    \item Provide a concise justification (10–25 words) describing what evidence supports your score.
\end{compactenum}

\textbf{SCORING (7-point Likert Scale)}:

{\quad\quad\quad\bfseries \textcolor{thedarkblue}{ [Refer to Table \ref{tab:scoring-rubric} for a detailed scoring rubric]}}
\\

Decimal values are not allowed. Use the rubric to guide your scoring. \\

\textbf{OUTPUT FORMAT (strict JSON)}:

{\quad\quad\quad\bfseries \textcolor{thedarkblue}{ [Refer to section \ref{json-schema} for a detailed JSON output schema]}}
\\

\textbf{CONSTRAINTS}:
\begin{compactenum}
    \item Respond with only one JSON block.
    \item The score must be an integer between 1 and 7.
    \item The justifications must reference specific visible evidence (not speculation).
    \item Do not restate the definition or include reasoning chains.
    \item Keep the tone factual, concise, and visually grounded.
\end{compactenum}

    \end{tiny}
  \end{promptbox}
\captionof{figure}{MLLM-as-a-Judge Image Editing Main prompt with no explicit rubric provided for each fine-grained factor, online setting.}
\label{fig:prompt-all-factors}
\end{figure}

\begin{figure}[H]
  \centering
  \begin{promptbox}{\small Factor-level Rubrics Prompt}
    
    \begin{tiny}
\textbf{ROLE}: You are an expert image editing evaluator. Your evaluations must be objective, consistent, and grounded entirely in visual comparison and task intent. \\

\textbf{CONTEXT:} You are provided with three inputs:
\begin{compactenum}
    \item Input Image – the unedited image.

    {\quad\quad\quad\centering{{\bfseries\ttfamily [input image]}}}

   \item Edited Image – the image produced after editing.
   
   {\quad\quad\quad\centering{{\bfseries\ttfamily [edited image]}}}
   \item Edit Instruction – a natural language description of the intended modification.
   
   {\quad\quad\quad\centering{{\bfseries\ttfamily [text instruction]}}}
\end{compactenum}

Your task is to evaluate how well the Edited Image aligns with the Input Image according to the Edit Instruction. \\

\textbf{FACTORS:} 

{\quad\quad\bfseries \textcolor{thedarkblue}{ [Refer to Table \ref{tab:judge_rubric} for a detailed list of factors and a rubric for each factor]}}
\\

\textbf{EVALUATION STEPS:}
\begin{compactenum}
    \item Compare the Edited Image to the Ground Truth Image in the context of the Edit Instruction.
    \item Assess how well the edited image satisfies each factor definition.
    \item Assign a score between 1 and 7 (integers only) using the rubric above.
    \item Provide a concise justification (10–25 words) describing what evidence supports your score.
\end{compactenum}

\textbf{SCORING (7-point Likert Scale)}:

{\quad\quad\quad\bfseries \textcolor{thedarkblue}{ [Refer to Table \ref{tab:scoring-rubric} for a detailed scoring rubric]}}
\\

Decimal values are not allowed. Use the rubric to guide your scoring. \\

\textbf{OUTPUT FORMAT (strict JSON)}:

{\quad\quad\quad\bfseries \textcolor{thedarkblue}{ [Refer to section \ref{json-schema} for a detailed JSON output schema]}}
\\

\textbf{CONSTRAINTS}:
\begin{compactenum}
    \item Respond with only one JSON block.
    \item The score must be an integer between 1 and 7.
    \item The justifications must reference specific visible evidence (not speculation).
    \item Do not restate the definition or include reasoning chains.
    \item Keep the tone factual, concise, and visually grounded.
\end{compactenum}

    \end{tiny}
  \end{promptbox}
\captionof{figure}{MLLM-as-a-Judge Image Editing Factor-level Rubrics prompt used for MLLM-as-a-Judge, online setting.}
\label{fig:prompt-mllm-as-a-judge-image-editing, main}
\end{figure}

\begin{figure}[H]
  \centering
  \begin{promptbox}{\small Category wise, Example guided Prompt}
    
    \begin{tiny}
\textbf{ROLE:}
You are an expert image editing evaluator specializing in instruction fidelity analysis. Your evaluations must be objective, consistent, and grounded entirely in assessing how well the edit follows the given instruction. \\

\textbf{CONTEXT}:
You are provided with three inputs:
\begin{compactenum}
\item \textbf{Input Image} – the original image before any editing [input image]
\item \textbf{Edited Image} – the image produced after applying the edit instruction [edited image]  
\item \textbf{Edit Instruction} – a natural language description of the intended modification [text instruction]
\end{compactenum}

Your task is to evaluate how faithfully the Edited Image executes the Edit Instruction. You will assess three specific factors related to instruction fidelity.

\textbf{FACTORS UNDER REVIEW:}

=== FACTOR 1: ALIGNMENT ===

\textbf{Definition}: Evaluates whether the edited image aligns with the specific edits provided in the instructions—whether what was requested was actually done. \\

What to examine:
\begin{compactenum}
    \item Parse the Edit Instruction carefully to identify all requested changes (e.g., "change the car to red" requests a color change to red)
    \item Check whether each requested change is present in the Edited Image
    \item Verify accuracy of the changes: If the instruction asks for "red," is it red (not orange or pink)? If it asks for "a dog," is there a dog (not a cat)?
    \item Assess specificity matching: If the instruction specifies "vintage wooden chair," does the result show a vintage wooden chair (not a modern plastic chair)?
    \item Check for correct targets: If the instruction says "change the woman's hat," was the woman's hat changed (not someone else's or a different item)?
    \item Evaluate whether the edit follows the instruction's intent and specific requirements
\end{compactenum}

\textbf{Important}: This factor focuses on WHETHER the requested changes match what was asked for. Do not evaluate if ALL parts were done (that's Completeness) or if the result is realistic (that's Plausibility). \\

\textbf{Score high (6-7) when}:
\begin{compactenum}
    \item All requested changes accurately match the instruction's specifications
    \item Target objects/attributes are correctly identified and modified
    \item Specific details (colors, object types, attributes) align precisely with what was requested
    \item The edit clearly follows the instruction's intent
\end{compactenum}

\textbf{Score low (1-3) when}:
\begin{compactenum}
    \item Requested changes are incorrect or inaccurate (wrong color, wrong object type, etc.)
    \item Wrong elements were modified instead of the specified targets
    \item The edit contradicts or misinterprets the instruction
    \item Specific requirements are ignored or incorrectly executed
\end{compactenum}

\textbf{EVALUATION STEPS:}
\begin{compactenum}
    \item Read the Edit Instruction carefully and identify all requested changes, targets, and specifications
    \item For each factor, systematically examine the Edited Image in relation to the instruction
    \item Compare the Edited Image to the Input Image to understand what changed
    \item Look for specific evidence relevant to each factor definition
    \item Assign a score between 1 and 7 (integers only) for each factor using the Likert scale below
    \item Provide a concise justification (15-30 words) for each factor, citing specific observable evidence
\end{compactenum}

\textbf{SCORING (7-point Likert Scale)}:

{\quad\quad\quad\bfseries \textcolor{thedarkblue}{ [Refer to Table \ref{tab:scoring-rubric} for a detailed scoring rubric]}}
\\

Decimal values are not allowed. Use the rubric to guide your scoring. \\

\textbf{OUTPUT FORMAT (strict JSON)}:

{\quad\quad\quad\bfseries \textcolor{thedarkblue}{ [Refer to section \ref{json-schema} for a detailed JSON output schema]}}
\\

\textbf{CONSTRAINTS}:
\begin{compactenum}
    \item Respond with only one JSON block containing all three factors
    \item Each score must be an integer between 1 and 7
    \item Each justification must reference specific, observable visual evidence (e.g., "instruction requested red car but result shows blue car" not "color doesn't match")
    \item Do not restate definitions or include reasoning chains in justifications
    \item Be precise: identify WHAT aspects of the instruction were or weren't followed
    \item Remain objective: evaluate only what is visible and what the instruction requested
    \item Keep tone factual, concise, and visually grounded
    \item Evaluate each factor independently—do not let one factor's assessment influence another
    \item For Alignment: focus on accuracy of what was done
    \item For Completeness: focus on thoroughness—whether everything was done
    \item For Plausibility: focus on real-world possibility of the result
\end{compactenum}

    \end{tiny}
  \end{promptbox}
\captionof{figure}{MLLM-as-a-Judge Image Editing Category wise, Example guided prompt for the alignment factor within the instruction fidelity category, separated by categories, online setting. The other factors follow the same detailed format.}
\label{fig:prompt-mllm-as-a-judge-image-editing, category}
\end{figure}

\begin{table}[t]
\centering
\caption{7-point Likert Scale Score used by the MLLM Judge and human evaluators.
}
\label{tab:scoring-rubric}
\small
\begin{tabular}{c l}
\toprule
\textbf{Score} & \textbf{Description} \\
\midrule
1 & Strongly Disagree \\
2 & Disagree \\
3 & Somewhat Disagree \\
4 & Neither Agree nor Disagree \\
5 & Somewhat Agree \\
6 & Agree \\
7 & Strongly Agree \\
\bottomrule
\end{tabular}
\end{table}

\begin{table*}[t]
\centering
\scriptsize
\setlength{\tabcolsep}{0.4em}
\renewcommand{\arraystretch}{1.35}

\caption{Rubric for Image Editing Factors (7-point Likert scale). This rubric is used in our fine-grained MLLM judge for image editing implementation shown in.}
\label{tab:judge_rubric}
\begin{tabularx}{\textwidth}{lXXX}
\toprule
\textbf{Factor} & \textbf{Score 1} & \textbf{Score 4} & \textbf{Score 7} \\
\midrule

\textcolor{GoogleGreen}{Unchanged Regions} 
& The model changed large areas unrelated to the instruction 
& Small artifacts exist but most regions are intact 
& No unintended change is visible \\
\midrule

\textcolor{GoogleGreen}{Global Consistency}
& The overall style, layout, or color scheme is drastically different 
& Minor inconsistencies in style or layout are present 
& The overall appearance is fully consistent \\
\midrule

\textcolor{GoogleGreen}{Identity Preservation}
& Core identifying features have been significantly altered or lost 
& Some features have changed but entities remain generally recognizable 
& All entities retain their distinguishing characteristics perfectly \\
\midrule

\textcolor{GoogleBlue}{Scale Realism}
& The edited object's scale is highly unrealistic or implausible 
& The scale is somewhat off but not jarringly unrealistic 
& The scale is completely realistic and proportionate \\
\midrule

\textcolor{GoogleBlue}{Spatial Relationship}
& Objects are misplaced or spatial relationships are severely disrupted 
& Minor spatial inconsistencies exist but overall relationships hold 
& All spatial relationships are perfectly maintained \\
\midrule

\textcolor{GoogleBlue}{Texture and Detail}
& Texture is notably different or detail is significantly degraded 
& Texture matches reasonably well with minor detail inconsistencies 
& Texture and detail are seamlessly consistent throughout \\
\midrule

\textcolor{GoogleBlue}{Image Quality}
& Severe noise, blurring, or distortions are present 
& Minor quality issues are noticeable but not severe 
& Image quality is excellent with no artifacts \\
\midrule

\textcolor{GoogleBlue}{Color and Lighting}
& Colors or lighting are severely mismatched with obvious inconsistencies 
& Colors and lighting mostly match with minor discrepancies 
& Colors, shadows, and lighting are perfectly harmonious \\
\midrule

\textcolor{GoogleBlue}{Seamlessness}
& Transitions are obvious with clear visible boundaries or seams 
& Transitions are mostly smooth with minor detectable edges 
& Transitions are completely seamless and undetectable \\
\midrule

\textcolor{GoogleRed}{Alignment}
& The edit does not match the instruction or contradicts it 
& The edit partially matches but misses some key aspects 
& The edit perfectly matches all aspects of the instruction \\
\midrule

\textcolor{GoogleRed}{Completeness} 
& Major parts of the instruction were not executed 
& Most aspects were completed but some elements are missing 
& Every aspect of the instruction was fully executed \\
\midrule

\textcolor{GoogleRed}{Plausibility}
& The result is highly implausible or violates real-world logic 
& The result is somewhat plausible but has noticeable oddities 
& The result is completely plausible and realistic \\
\bottomrule

\end{tabularx}
\end{table*}

\subsection{JSON output}

\begin{promptbox}{\small JSON scheme}
\label{json-schema}

\begin{tiny}
\begin{verbatim}
{
  "image_id": "<image_identifier>",
  "offline_factor_results": {
    "unchanged_regions": {
      "score": <integer_1_to_7>,
      "justification": "<brief_justification>"
    },
    "global_consistency": {
      "score": <integer_1_to_7>,
      "justification": "<brief_justification>"
    },
    "identity_preservation": {
      "score": <integer_1_to_7>,
      "justification": "<brief_justification>"
    },
    "scale_realism": {
      "score": <integer_1_to_7>,
      "justification": "<brief_justification>"
    },
    "spatial_relationship": {
      "score": <integer_1_to_7>,
      "justification": "<brief_justification>"
    },
    "texture_and_detail": {
      "score": <integer_1_to_7>,
      "justification": "<brief_justification>"
    },
    "image_quality": {
      "score": <integer_1_to_7>,
      "justification": "<brief_justification>"
    },
    "color_and_lighting": {
      "score": <integer_1_to_7>,
      "justification": "<brief_justification>"
    },
    "seamlessness": {
      "score": <integer_1_to_7>,
      "justification": "<brief_justification>"
    },
    "alignment": {
      "score": <integer_1_to_7>,
      "justification": "<brief_justification>"
    },
    "completeness": {
      "score": <integer_1_to_7>,
      "justification": "<brief_justification>"
    },
    "plausibility": {
      "score": <integer_1_to_7>,
      "justification": "<brief_justification>"
    }
  }
}
\end{verbatim}
\end{tiny}

\end{promptbox}

\section{Ablation Study}
\label{app:ablation_study}
In this section, we will provide detailed results on how different MLLM base model and different prompt formats may influence the MLLM-as-a-Judge's performance.

\subsection{Human Evaluation Results Comparison}

\begin{table*}[t]
\centering
\caption{\textbf{Human vs. GPT vs. Gemini judge scores across factors and edit types.} Mean $\pm$ std 1--7 Likert scores for human annotations and two MLLM judges (GPT and Gemini). When the difference between our judge score and human evaluation is closer than $0.5$, its background is dark green. When the difference is closer than $1.0$, its background is light green. The prompt used in both models is Factor-level Rubrics (Figure~\ref{fig:prompt-mllm-as-a-judge-image-editing, main})}
\label{tab:human_gpt_gemini}
\tiny
\setlength{\tabcolsep}{3pt}
\renewcommand{\arraystretch}{1.1}
\begin{tabular}{@{}c c l *{6}{c} c@{}}
\toprule
& & & \multicolumn{6}{c}{\sc Image Edit Types} \\
\cmidrule{4-9}
& \textbf{Factor} & & \textbf{Add} & \textbf{Remove} & \textbf{Replace} & \textbf{Action} & \textbf{Counting} & \textbf{Relation} & \textbf{All Edits} \\
\midrule
\multirow{9}{*}{\rotatebox[origin=c]{90}{\textcolor{GoogleGreen}{\textsc{\sc \bfseries Image Preserv.}}}} & \multirow{3}{*}{\centering\arraybackslash \textcolor{GoogleGreen}{\textbf{Unchanged Regions}}} & Human & \score{5.172}{0.82} & \score{5.731}{0.75} & \score{4.972}{1.02} & \score{4.352}{1.06} & \score{3.393}{0.68} & \score{4.992}{0.43} & \score{4.769}{0.74} \\
 &  & GPT Judge & \score{5.889}{1.20} & \cellcolor{green!20}\textbf{\score{5.235}{1.68}} & \score{5.778}{1.69} & \score{5.522}{1.44} & \cellcolor{green!30}\textbf{\score{3.600}{1.96}} & \cellcolor{green!30}\textbf{\score{5.167}{1.67}} & \cellcolor{green!20}\textbf{\score{5.198}{0.76}} \\
 &  & Gemini Judge & \cellcolor{green!30}\textbf{\score{5.111}{2.69}} & \score{4.206}{2.67} & \score{3.444}{2.39} & \score{2.957}{2.69} & \score{2.200}{2.40} & \score{4.000}{3.00} & \score{3.653}{0.93} \\
\cmidrule(l){2-10}
 & \multirow{3}{*}{\centering\arraybackslash \textcolor{GoogleGreen}{\textbf{Global Consistency}}} & Human & \score{5.602}{0.70} & \score{5.982}{0.61} & \score{5.551}{0.90} & \score{4.769}{0.87} & \score{5.243}{1.13} & \score{5.444}{0.39} & \score{5.432}{0.37} \\
 &  & GPT Judge & \cellcolor{green!30}\textbf{\score{5.556}{1.07}} & \cellcolor{green!20}\textbf{\score{5.618}{1.51}} & \cellcolor{green!30}\textbf{\score{5.778}{1.62}} & \score{5.783}{1.06} & \score{4.200}{1.94} & \cellcolor{green!30}\textbf{\score{5.500}{1.50}} & \cellcolor{green!30}\textbf{\score{5.406}{0.55}} \\
 &  & Gemini Judge & \cellcolor{green!20}\textbf{\score{5.111}{2.69}} & \score{4.824}{2.53} & \score{4.333}{2.40} & \score{3.870}{2.52} & \score{2.400}{2.33} & \score{4.333}{2.75} & \score{4.145}{0.87} \\
\cmidrule(l){2-10}
 & \multirow{3}{*}{\centering\arraybackslash \textcolor{GoogleGreen}{\textbf{Identity Preservation}}} & Human & \score{5.613}{0.62} & \score{5.913}{0.82} & \score{5.625}{0.84} & \score{4.871}{1.12} & \score{4.227}{1.12} & \score{5.714}{0.20} & \score{5.327}{0.59} \\
 &  & GPT Judge & \cellcolor{green!30}\textbf{\score{5.444}{1.57}} & \cellcolor{green!20}\textbf{\score{6.324}{1.51}} & \cellcolor{green!20}\textbf{\score{5.889}{1.91}} & \score{6.696}{0.86} & \score{5.400}{1.36} & \score{5.167}{2.11} & \cellcolor{green!20}\textbf{\score{5.820}{0.54}} \\
 &  & Gemini Judge & \cellcolor{green!30}\textbf{\score{5.556}{2.27}} & \score{5.324}{2.37} & \score{4.222}{2.64} & \score{3.478}{2.72} & \score{1.400}{0.80} & \score{3.500}{2.57} & \score{3.913}{1.38} \\
\midrule
\multirow{18}{*}{\rotatebox[origin=c]{90}{\textcolor{GoogleBlue}{\textsc{\sc \bfseries Edit Quality}}}} & \multirow{3}{*}{\centering\arraybackslash \textcolor{GoogleBlue}{\textbf{Scale Realism}}} & Human & \score{5.276}{0.95} & \score{6.286}{0.54} & \score{5.865}{0.80} & \score{5.984}{0.61} & \score{5.510}{0.68} & \score{6.033}{0.57} & \score{5.826}{0.34} \\
 &  & GPT Judge & \cellcolor{green!20}\textbf{\score{5.000}{1.94}} & \cellcolor{green!30}\textbf{\score{6.471}{1.40}} & \score{6.667}{1.00} & \score{7.000}{0.00} & \cellcolor{green!20}\textbf{\score{5.800}{1.47}} & \score{5.333}{1.37} & \cellcolor{green!30}\textbf{\score{6.045}{0.72}} \\
 &  & Gemini Judge & \score{5.889}{1.97} & \score{6.824}{1.01} & \score{6.667}{1.37} & \cellcolor{green!30}\textbf{\score{6.130}{1.96}} & \score{6.800}{0.40} & \cellcolor{green!20}\textbf{\score{6.500}{0.76}} & \score{6.468}{0.35} \\
\cmidrule(l){2-10}
 & \multirow{3}{*}{\centering\arraybackslash \textcolor{GoogleBlue}{\textbf{Spatial Relationship}}} & Human & \score{5.561}{0.79} & \score{6.225}{0.50} & \score{5.890}{0.63} & \score{5.948}{0.74} & \score{4.650}{1.19} & \score{5.728}{0.63} & \score{5.667}{0.50} \\
 &  & GPT Judge & \score{4.778}{1.55} & \cellcolor{green!30}\textbf{\score{6.471}{1.33}} & \score{6.833}{0.50} & \cellcolor{green!20}\textbf{\score{6.217}{1.02}} & \cellcolor{green!20}\textbf{\score{4.200}{1.60}} & \score{4.000}{1.91} & \cellcolor{green!20}\textbf{\score{5.417}{1.13}} \\
 &  & Gemini Judge & \score{4.222}{2.90} & \cellcolor{green!30}\textbf{\score{6.235}{1.90}} & \cellcolor{green!30}\textbf{\score{5.889}{2.16}} & \score{4.696}{2.56} & \score{3.600}{2.33} & \score{4.333}{2.75} & \score{4.829}{0.93} \\
\cmidrule(l){2-10}
 & \multirow{3}{*}{\centering\arraybackslash \textcolor{GoogleBlue}{\textbf{Texture and Detail}}} & Human & \score{5.504}{0.56} & \score{5.844}{0.61} & \score{5.483}{0.80} & \score{5.643}{0.84} & \score{5.157}{0.90} & \score{5.639}{0.59} & \score{5.545}{0.21} \\
 &  & GPT Judge & \cellcolor{green!20}\textbf{\score{5.222}{0.79}} & \score{5.176}{1.48} & \score{4.611}{1.83} & \cellcolor{green!30}\textbf{\score{5.826}{0.87}} & \cellcolor{green!30}\textbf{\score{5.000}{1.79}} & \cellcolor{green!20}\textbf{\score{5.167}{1.21}} & \cellcolor{green!20}\textbf{\score{5.167}{0.36}} \\
 &  & Gemini Judge & \score{4.000}{2.40} & \score{3.735}{2.59} & \score{4.389}{2.38} & \score{3.609}{2.62} & \score{4.400}{2.33} & \score{3.667}{2.56} & \score{3.967}{0.33} \\
\cmidrule(l){2-10}
 & \multirow{3}{*}{\centering\arraybackslash \textcolor{GoogleBlue}{\textbf{Image Quality}}} & Human & \score{5.569}{0.54} & \score{6.048}{0.66} & \score{5.513}{0.71} & \score{5.947}{0.71} & \score{5.247}{0.53} & \score{5.683}{0.75} & \score{5.668}{0.27} \\
 &  & GPT Judge & \score{6.667}{0.47} & \cellcolor{green!30}\textbf{\score{6.235}{0.94}} & \score{6.667}{0.75} & \cellcolor{green!20}\textbf{\score{6.348}{0.91}} & \score{5.800}{1.60} & \cellcolor{green!20}\textbf{\score{6.000}{1.00}} & \score{6.286}{0.32} \\
 &  & Gemini Judge & \cellcolor{green!30}\textbf{\score{5.556}{2.11}} & \score{5.000}{2.35} & \cellcolor{green!30}\textbf{\score{5.611}{2.16}} & \score{4.652}{2.46} & \score{7.000}{0.00} & \cellcolor{green!30}\textbf{\score{5.500}{2.29}} & \cellcolor{green!30}\textbf{\score{5.553}{0.73}} \\
\cmidrule(l){2-10}
 & \multirow{3}{*}{\centering\arraybackslash \textcolor{GoogleBlue}{\textbf{Color and Lighting}}} & Human & \score{5.515}{0.75} & \score{5.855}{0.71} & \score{5.442}{0.86} & \score{5.549}{0.72} & \score{5.403}{0.83} & \score{5.553}{0.64} & \score{5.553}{0.15} \\
 &  & GPT Judge & \cellcolor{green!20}\textbf{\score{6.000}{0.94}} & \cellcolor{green!30}\textbf{\score{5.794}{1.21}} & \cellcolor{green!30}\textbf{\score{5.611}{1.57}} & \cellcolor{green!20}\textbf{\score{5.913}{1.06}} & \cellcolor{green!30}\textbf{\score{5.200}{2.23}} & \cellcolor{green!30}\textbf{\score{5.500}{1.12}} & \cellcolor{green!30}\textbf{\score{5.670}{0.27}} \\
 &  & Gemini Judge & \score{4.222}{2.39} & \score{3.441}{2.52} & \score{4.611}{2.52} & \score{3.826}{2.51} & \score{3.600}{2.80} & \score{4.667}{2.21} & \score{4.061}{0.47} \\
\cmidrule(l){2-10}
 & \multirow{3}{*}{\centering\arraybackslash \textcolor{GoogleBlue}{\textbf{Seamlessness}}} & Human & \score{5.722}{0.64} & \score{6.101}{0.58} & \score{5.767}{0.74} & \score{5.598}{0.83} & \score{5.357}{1.00} & \score{5.578}{0.73} & \score{5.687}{0.23} \\
 &  & GPT Judge & \cellcolor{green!30}\textbf{\score{5.667}{0.82}} & \score{5.118}{1.79} & \cellcolor{green!30}\textbf{\score{5.778}{1.44}} & \cellcolor{green!20}\textbf{\score{5.913}{1.06}} & \score{4.200}{2.04} & \score{4.667}{1.37} & \cellcolor{green!20}\textbf{\score{5.224}{0.63}} \\
 &  & Gemini Judge & \cellcolor{green!20}\textbf{\score{5.333}{2.36}} & \score{4.059}{2.63} & \cellcolor{green!30}\textbf{\score{6.000}{2.00}} & \score{4.348}{2.53} & \cellcolor{green!30}\textbf{\score{5.200}{2.40}} & \score{4.167}{2.85} & \score{4.851}{0.71} \\
\midrule
\multirow{9}{*}{\rotatebox[origin=c]{90}{\textcolor{GoogleRed}{\textsc{\sc \bfseries Instruct. Fidel.}}}} & \multirow{3}{*}{\centering\arraybackslash \textcolor{GoogleRed}{\textbf{Alignment}}} & Human & \score{5.556}{0.49} & \score{5.927}{0.97} & \score{5.681}{0.65} & \score{5.666}{1.13} & \score{3.437}{1.40} & \score{5.178}{0.99} & \score{5.241}{0.84} \\
 &  & GPT Judge & \cellcolor{green!30}\textbf{\score{5.444}{1.71}} & \score{4.706}{2.52} & \score{3.833}{2.63} & \cellcolor{green!30}\textbf{\score{5.522}{2.38}} & \score{2.800}{1.60} & \score{3.667}{2.56} & \score{4.329}{0.99} \\
 &  & Gemini Judge & \score{3.111}{1.29} & \score{3.559}{2.80} & \score{2.278}{1.69} & \score{3.217}{2.48} & \score{4.000}{2.68} & \score{2.667}{2.05} & \score{3.139}{0.56} \\
\cmidrule(l){2-10}
 & \multirow{3}{*}{\centering\arraybackslash \textcolor{GoogleRed}{\textbf{Completeness}}} & Human & \score{5.693}{0.72} & \score{5.966}{1.17} & \score{5.789}{0.71} & \score{5.719}{1.14} & \score{3.537}{1.74} & \score{5.556}{0.66} & \score{5.376}{0.83} \\
 &  & GPT Judge & \cellcolor{green!20}\textbf{\score{6.111}{1.66}} & \score{4.618}{2.49} & \score{3.833}{2.61} & \cellcolor{green!20}\textbf{\score{5.435}{2.37}} & \score{2.600}{1.50} & \score{3.500}{2.36} & \score{4.349}{1.18} \\
 &  & Gemini Judge & \score{5.000}{2.83} & \score{4.294}{2.92} & \score{2.778}{2.46} & \score{3.957}{2.79} & \score{4.600}{2.94} & \score{3.167}{2.73} & \score{3.966}{0.78} \\
\cmidrule(l){2-10}
 & \multirow{3}{*}{\centering\arraybackslash \textcolor{GoogleRed}{\textbf{Plausibility}}} & Human & \score{5.209}{1.03} & \score{6.023}{0.78} & \score{5.743}{0.80} & \score{5.692}{1.17} & \score{4.917}{1.08} & \score{5.586}{0.89} & \score{5.528}{0.36} \\
 &  & GPT Judge & \score{6.000}{1.15} & \cellcolor{green!20}\textbf{\score{6.353}{1.26}} & \score{6.444}{1.01} & \score{6.826}{0.64} & \cellcolor{green!20}\textbf{\score{5.200}{1.17}} & \cellcolor{green!30}\textbf{\score{5.500}{0.76}} & \score{6.054}{0.56} \\
 &  & Gemini Judge & \cellcolor{green!20}\textbf{\score{5.556}{2.17}} & \score{5.353}{2.58} & \cellcolor{green!20}\textbf{\score{6.056}{2.12}} & \score{4.391}{2.81} & \score{7.000}{0.00} & \score{4.000}{3.00} & \cellcolor{green!30}\textbf{\score{5.393}{1.00}} \\
\bottomrule
\end{tabular}
\end{table*}

\subsubsection{Varying MLLMs}
Table~\ref{tab:human_gpt_gemini} presents a comparative analysis of human evaluations alongside two MLLM judges, GPT and Gemini, across all factors and image edit types. Overall, both judges captured meaningful aspects of human judgment, but the GPT judge exhibited consistently stronger alignment with human evaluations. This was reflected by the larger number of green-highlighted cells for GPT, indicating smaller absolute differences from human scores across a wide range of factors.

In particular, GPT showed robust agreement with human ratings for image preservation and edit quality factors, including Unchanged Regions, Global Consistency, Spatial Relationship, Texture and Detail, Color and Lighting, and Seamlessness, where its scores closely tracked human averages across most edit types. In contrast, even though the Gemini judge perform better than GPT judge in several cases, like Plausibility\&Add, Spatial Relationship\&Replace, and Scale Realism\&Action, the Gemini judge displayed larger deviations and higher variance in most factors, and got a weaker alignment than GPT judge in nearly all the factors results.

Taken together, these results suggested that the GPT judge provided a more reliable and human-aligned evaluation signal than Gemini, particularly when assessing fine-grained perceptual quality and overall edit correctness across diverse editing scenarios.

\begin{table*}[t!]
\setlength{\tabcolsep}{2mm} 
\centering
\caption{
\textbf{Human and our MLLM-as-a-Judge scores for all factors and across all edit types.}
We report the average score over all image edit types in the last column and over all factors in the last row.  When the difference between our judge score and human evaluation is closer than $0.5$, its background is dark green. When the difference is closer than $1.0$, its background is light green. The gray text is the standard deviation of the scores from which the average is computed. Judge scores are generated using the
Factor-level Rubrics prompt shown in Fig.~\ref{fig:prompt-mllm-as-a-judge-image-editing, main}.
}
\label{tab:human_judge_v3}
\tiny
\setlength{\tabcolsep}{3pt}
\renewcommand{\arraystretch}{1.1}
\begin{tabular}{@{}c c l *{6}{c} c@{}}
\toprule
& & &
\multicolumn{6}{c}{\sc Image Edit Types} \\
\cmidrule{4-9}
&
\textbf{Factor} &
&
\makecell{\textbf{Add}} &
\makecell{\textbf{Remove}} &
\makecell{\textbf{Replace}} &
\makecell{\textbf{Action}} &
\makecell{\textbf{Counting}} &
\makecell{\textbf{Relation}} &
\textbf{All Edits} \\
\midrule
\multirow{6}{*}{\rotatebox[origin=c]{90}{\textcolor{GoogleGreen}{\textsc{\sc \bfseries Image Preserv.}}}} & \multirow{2}{*}{\centering\arraybackslash \textcolor{GoogleGreen}{\textbf{Unchanged Regions}}} & Human & \score{5.172}{0.82} & \score{5.731}{0.75} & \score{4.972}{1.02} & \score{4.352}{1.06} & \score{3.393}{0.68} & \score{4.992}{0.43} & \score{4.769}{0.74} \\
 &  & Our Judge & \score{6.778}{0.42} & \cellcolor{green!30}\textbf{\score{6.000}{1.14}} & \score{6.667}{0.47} & \score{5.826}{1.01} & \cellcolor{green!30}\textbf{\score{3.600}{2.06}} & \cellcolor{green!30}\textbf{\score{5.167}{1.67}} & \cellcolor{green!20}\textbf{\score{5.673}{1.07}} \\
\cmidrule(l){2-10}
 & \multirow{2}{*}{\centering\arraybackslash \textcolor{GoogleGreen}{\textbf{Global Consistency}}} & Human & \score{5.602}{0.70} & \score{5.982}{0.61} & \score{5.551}{0.90} & \score{4.769}{0.87} & \score{5.243}{1.13} & \score{5.444}{0.39} & \score{5.432}{0.37} \\
 &  & Our Judge & \score{6.778}{0.42} & \cellcolor{green!30}\textbf{\score{6.471}{0.88}} & \score{6.667}{0.67} & \score{6.348}{0.70} & \cellcolor{green!30}\textbf{\score{5.600}{1.36}} & \cellcolor{green!20}\textbf{\score{6.167}{1.21}} & \cellcolor{green!20}\textbf{\score{6.338}{0.39}} \\
\cmidrule(l){2-10}
 & \multirow{2}{*}{\centering\arraybackslash \textcolor{GoogleGreen}{\textbf{Identity Preservation}}} & Human & \score{5.613}{0.62} & \score{5.913}{0.82} & \score{5.625}{0.84} & \score{4.871}{1.12} & \score{4.227}{1.12} & \score{5.714}{0.20} & \score{5.327}{0.59} \\
 &  & Our Judge & \score{7.000}{0.00} & \cellcolor{green!20}\textbf{\score{6.529}{0.95}} & \cellcolor{green!20}\textbf{\score{6.556}{1.38}} & \score{6.957}{0.20} & \score{6.400}{0.80} & \cellcolor{green!20}\textbf{\score{6.667}{0.75}} & \score{6.685}{0.22} \\
\midrule
\multirow{12}{*}{\rotatebox[origin=c]{90}{\textcolor{GoogleBlue}{\textsc{\sc \bfseries Edit Quality}}}} & \multirow{2}{*}{\centering\arraybackslash \textcolor{GoogleBlue}{\textbf{Scale Realism}}} & Human & \score{5.276}{0.95} & \score{6.286}{0.54} & \score{5.865}{0.80} & \score{5.984}{0.61} & \score{5.510}{0.68} & \score{6.033}{0.57} & \score{5.826}{0.34} \\
 &  & Our Judge & \score{6.667}{0.94} & \cellcolor{green!20}\textbf{\score{6.853}{0.60}} & \score{6.889}{0.31} & \score{7.000}{0.00} & \score{7.000}{0.00} & \cellcolor{green!30}\textbf{\score{6.333}{1.11}} & \cellcolor{green!20}\textbf{\score{6.790}{0.23}} \\
\cmidrule(l){2-10}
 & \multirow{2}{*}{\centering\arraybackslash \textcolor{GoogleBlue}{\textbf{Spatial Relationship}}} & Human & \score{5.561}{0.79} & \score{6.225}{0.50} & \score{5.890}{0.63} & \score{5.948}{0.74} & \score{4.650}{1.19} & \score{5.728}{0.63} & \score{5.667}{0.50} \\
 &  & Our Judge & \score{6.667}{0.94} & \cellcolor{green!30}\textbf{\score{6.529}{1.06}} & \score{6.944}{0.23} & \cellcolor{green!20}\textbf{\score{6.609}{0.71}} & \cellcolor{green!20}\textbf{\score{5.200}{1.17}} & \cellcolor{green!20}\textbf{\score{6.500}{1.12}} & \cellcolor{green!20}\textbf{\score{6.408}{0.56}} \\
\cmidrule(l){2-10}
 & \multirow{2}{*}{\centering\arraybackslash \textcolor{GoogleBlue}{\textbf{Texture and Detail}}} & Human & \score{5.504}{0.56} & \score{5.844}{0.61} & \score{5.483}{0.80} & \score{5.643}{0.84} & \score{5.157}{0.90} & \score{5.639}{0.59} & \score{5.545}{0.21} \\
 &  & Our Judge & \cellcolor{green!20}\textbf{\score{6.222}{1.03}} & \cellcolor{green!30}\textbf{\score{5.794}{0.80}} & \cellcolor{green!20}\textbf{\score{6.167}{0.37}} & \cellcolor{green!20}\textbf{\score{6.217}{0.66}} & \score{6.200}{1.17} & \cellcolor{green!30}\textbf{\score{5.833}{1.07}} & \cellcolor{green!20}\textbf{\score{6.072}{0.18}} \\
\cmidrule(l){2-10}
 & \multirow{2}{*}{\centering\arraybackslash \textcolor{GoogleBlue}{\textbf{Image Quality}}} & Human & \score{5.569}{0.54} & \score{6.048}{0.66} & \score{5.513}{0.71} & \score{5.947}{0.71} & \score{5.247}{0.53} & \score{5.683}{0.75} & \score{5.668}{0.27} \\
 &  & Our Judge & \score{6.778}{0.42} & \cellcolor{green!30}\textbf{\score{6.412}{0.60}} & \score{6.889}{0.31} & \cellcolor{green!20}\textbf{\score{6.652}{0.56}} & \score{7.000}{0.00} & \cellcolor{green!20}\textbf{\score{6.500}{1.12}} & \score{6.705}{0.21} \\
\cmidrule(l){2-10}
 & \multirow{2}{*}{\centering\arraybackslash \textcolor{GoogleBlue}{\textbf{Color and Lighting}}} & Human & \score{5.515}{0.75} & \score{5.855}{0.71} & \score{5.442}{0.86} & \score{5.549}{0.72} & \score{5.403}{0.83} & \score{5.553}{0.64} & \score{5.553}{0.15} \\
 &  & Our Judge & \cellcolor{green!20}\textbf{\score{6.444}{0.96}} & \cellcolor{green!20}\textbf{\score{6.500}{0.78}} & \score{6.556}{0.50} & \cellcolor{green!20}\textbf{\score{6.391}{0.49}} & \score{6.600}{0.80} & \cellcolor{green!20}\textbf{\score{6.167}{0.90}} & \cellcolor{green!20}\textbf{\score{6.443}{0.14}} \\
\cmidrule(l){2-10}
 & \multirow{2}{*}{\centering\arraybackslash \textcolor{GoogleBlue}{\textbf{Seamlessness}}} & Human & \score{5.722}{0.64} & \score{6.101}{0.58} & \score{5.767}{0.74} & \score{5.598}{0.83} & \score{5.357}{1.00} & \score{5.578}{0.73} & \score{5.687}{0.23} \\
 &  & Our Judge & \cellcolor{green!20}\textbf{\score{6.444}{1.07}} & \cellcolor{green!30}\textbf{\score{5.941}{1.06}} & \cellcolor{green!20}\textbf{\score{6.556}{0.50}} & \cellcolor{green!20}\textbf{\score{6.391}{0.82}} & \cellcolor{green!20}\textbf{\score{6.000}{1.10}} & \cellcolor{green!30}\textbf{\score{5.667}{1.70}} & \cellcolor{green!30}\textbf{\score{6.167}{0.32}} \\
\midrule
\multirow{6}{*}{\rotatebox[origin=c]{90}{\textcolor{GoogleRed}{\textsc{\sc \bfseries Instruct. Fidel.}}}} & \multirow{2}{*}{\centering\arraybackslash \textcolor{GoogleRed}{\textbf{Alignment}}} & Human & \score{5.556}{0.49} & \score{5.927}{0.97} & \score{5.681}{0.65} & \score{5.666}{1.13} & \score{3.437}{1.40} & \score{5.178}{0.99} & \score{5.241}{0.84} \\
 &  & Our Judge & \score{6.889}{0.31} & \cellcolor{green!20}\textbf{\score{6.471}{1.09}} & \cellcolor{green!20}\textbf{\score{6.500}{1.01}} & \score{6.739}{0.85} & \score{5.800}{2.40} & \score{7.000}{0.00} & \score{6.566}{0.39} \\
\cmidrule(l){2-10}
 & \multirow{2}{*}{\centering\arraybackslash \textcolor{GoogleRed}{\textbf{Completeness}}} & Human & \score{5.693}{0.72} & \score{5.966}{1.17} & \score{5.789}{0.71} & \score{5.719}{1.14} & \score{3.537}{1.74} & \score{5.556}{0.66} & \score{5.376}{0.83} \\
 &  & Our Judge & \score{6.889}{0.31} & \cellcolor{green!20}\textbf{\score{6.500}{1.14}} & \cellcolor{green!20}\textbf{\score{6.389}{1.11}} & \score{6.826}{0.64} & \score{6.000}{2.00} & \score{6.667}{0.75} & \score{6.545}{0.30} \\
\cmidrule(l){2-10}
 & \multirow{2}{*}{\centering\arraybackslash \textcolor{GoogleRed}{\textbf{Plausibility}}} & Human & \score{5.209}{1.03} & \score{6.023}{0.78} & \score{5.743}{0.80} & \score{5.692}{1.17} & \score{4.917}{1.08} & \score{5.586}{0.89} & \score{5.528}{0.36} \\
 &  & Our Judge & \score{6.667}{0.67} & \cellcolor{green!20}\textbf{\score{6.618}{0.87}} & \score{6.944}{0.23} & \score{6.870}{0.61} & \score{7.000}{0.00} & \cellcolor{green!20}\textbf{\score{6.167}{1.21}} & \score{6.711}{0.28} \\
\midrule
 & \textbf{Average} & Human & \score{5.499}{0.17} & \score{5.992}{0.15} & \score{5.610}{0.24} & \score{5.478}{0.50} & \score{4.673}{0.78} & \score{5.557}{0.26} & \score{5.652}{0.47} \\
 &  & Our Judge & \score{6.685}{0.21} & \cellcolor{green!30}\textbf{\score{6.385}{0.30}} & \score{6.644}{0.23} & \score{6.569}{0.33} & \score{6.033}{0.92} & \cellcolor{green!20}\textbf{\score{6.236}{0.48}} & \cellcolor{green!20}\textbf{\score{6.479}{0.40}} \\
\bottomrule
\end{tabular}
\end{table*}

\begin{table*}[t]
\centering
\caption{
\textbf{Human and our MLLM-as-a-Judge scores for all factors and across all edit types.}
We report the average score over all image edit types in the last column and over all factors in the last row. When the difference between our judge score and human evaluation is closer than $0.5$, its background is dark green. When the difference is closer than $1.0$, its background is light green. The gray text is the standard deviation of the scores from which the average is computed. Judge scores are generated using the
Category wise, Example guided prompt shown in Fig.~\ref{fig:prompt-mllm-as-a-judge-image-editing, category}.
}
\label{tab:human_judge_combined_category}
\tiny
\setlength{\tabcolsep}{3pt}
\renewcommand{\arraystretch}{1.1}
\begin{tabular}{@{}c c l *{6}{c} c@{}}
\toprule
& & &
\multicolumn{6}{c}{\sc Image Edit Types} \\
\cmidrule{4-9}
&
\textbf{Factor} &
&
\makecell{\textbf{Add}} &
\makecell{\textbf{Remove}} &
\makecell{\textbf{Replace}} &
\makecell{\textbf{Action}} &
\makecell{\textbf{Counting}} &
\makecell{\textbf{Relation}} &
\textbf{All Edits} \\
\midrule
\multirow{6}{*}{\rotatebox[origin=c]{90}{\textcolor{GoogleGreen}{\textsc{\sc \bfseries Image Preserv.}}}} & \multirow{2}{*}{\centering\arraybackslash \textcolor{GoogleGreen}{\textbf{Unchanged Regions}}} & Human & \score{5.172}{0.82} & \score{5.731}{0.75} & \score{4.972}{1.02} & \score{4.352}{1.06} & \score{3.393}{0.68} & \score{4.992}{0.43} & \score{4.769}{0.74} \\
 &  & Our Judge & \score{6.333}{0.47} & \cellcolor{green!30}\textbf{\score{5.382}{1.19}} & \score{6.000}{0.75} & \cellcolor{green!30}\textbf{\score{4.739}{1.07}} & \cellcolor{green!30}\textbf{\score{3.600}{1.62}} & \cellcolor{green!30}\textbf{\score{4.500}{1.38}} & \cellcolor{green!30}\textbf{\score{5.092}{0.93}} \\
\cmidrule(l){2-10}
 & \multirow{2}{*}{\centering\arraybackslash \textcolor{GoogleGreen}{\textbf{Global Consistency}}} & Human & \score{5.602}{0.70} & \score{5.982}{0.61} & \score{5.551}{0.90} & \score{4.769}{0.87} & \score{5.243}{1.13} & \score{5.444}{0.39} & \score{5.432}{0.37} \\
 &  & Our Judge & \score{6.667}{0.47} & \cellcolor{green!30}\textbf{\score{6.059}{1.11}} & \score{6.389}{0.49} & \score{5.957}{0.69} & \cellcolor{green!30}\textbf{\score{5.000}{1.41}} & \cellcolor{green!30}\textbf{\score{5.500}{0.96}} & \cellcolor{green!30}\textbf{\score{5.928}{0.55}} \\
\cmidrule(l){2-10}
 & \multirow{2}{*}{\centering\arraybackslash \textcolor{GoogleGreen}{\textbf{Identity Preservation}}} & Human & \score{5.613}{0.62} & \score{5.913}{0.82} & \score{5.625}{0.84} & \score{4.871}{1.12} & \score{4.227}{1.12} & \score{5.714}{0.20} & \score{5.327}{0.59} \\
 &  & Our Judge & \score{7.000}{0.00} & \cellcolor{green!30}\textbf{\score{6.265}{1.04}} & \cellcolor{green!20}\textbf{\score{6.278}{1.41}} & \score{6.435}{0.58} & \score{5.200}{1.33} & \cellcolor{green!20}\textbf{\score{6.333}{1.11}} & \score{6.252}{0.53} \\
\midrule
\multirow{12}{*}{\rotatebox[origin=c]{90}{\textcolor{GoogleBlue}{\textsc{\sc \bfseries Edit Quality}}}} & \multirow{2}{*}{\centering\arraybackslash \textcolor{GoogleBlue}{\textbf{Scale Realism}}} & Human & \score{5.276}{0.95} & \score{6.286}{0.54} & \score{5.865}{0.80} & \score{5.984}{0.61} & \score{5.510}{0.68} & \score{6.033}{0.57} & \score{5.826}{0.34} \\
 &  & Our Judge & \cellcolor{green!20}\textbf{\score{6.000}{0.47}} & \cellcolor{green!20}\textbf{\score{6.824}{0.38}} & \cellcolor{green!20}\textbf{\score{6.389}{0.95}} & \cellcolor{green!30}\textbf{\score{6.217}{0.51}} & \score{6.800}{0.40} & \cellcolor{green!30}\textbf{\score{5.833}{1.34}} & \cellcolor{green!20}\textbf{\score{6.344}{0.37}} \\
\cmidrule(l){2-10}
 & \multirow{2}{*}{\centering\arraybackslash \textcolor{GoogleBlue}{\textbf{Spatial Relationship}}} & Human & \score{5.561}{0.79} & \score{6.225}{0.50} & \score{5.890}{0.63} & \score{5.948}{0.74} & \score{4.650}{1.19} & \score{5.728}{0.63} & \score{5.667}{0.50} \\
 &  & Our Judge & \cellcolor{green!30}\textbf{\score{5.778}{1.13}} & \cellcolor{green!30}\textbf{\score{6.088}{1.20}} & \cellcolor{green!20}\textbf{\score{6.500}{0.60}} & \cellcolor{green!30}\textbf{\score{5.870}{0.95}} & \score{6.600}{0.49} & \cellcolor{green!30}\textbf{\score{5.667}{1.25}} & \cellcolor{green!30}\textbf{\score{6.084}{0.35}} \\
\cmidrule(l){2-10}
 & \multirow{2}{*}{\centering\arraybackslash \textcolor{GoogleBlue}{\textbf{Texture and Detail}}} & Human & \score{5.504}{0.56} & \score{5.844}{0.61} & \score{5.483}{0.80} & \score{5.643}{0.84} & \score{5.157}{0.90} & \score{5.639}{0.59} & \score{5.545}{0.21} \\
 &  & Our Judge & \cellcolor{green!30}\textbf{\score{5.333}{1.05}} & \cellcolor{green!30}\textbf{\score{5.529}{0.81}} & \cellcolor{green!30}\textbf{\score{5.889}{0.66}} & \cellcolor{green!30}\textbf{\score{5.739}{0.61}} & \score{6.000}{0.63} & \cellcolor{green!30}\textbf{\score{5.500}{1.26}} & \cellcolor{green!30}\textbf{\score{5.665}{0.23}} \\
\cmidrule(l){2-10}
 & \multirow{2}{*}{\centering\arraybackslash \textcolor{GoogleBlue}{\textbf{Image Quality}}} & Human & \score{5.569}{0.54} & \score{6.048}{0.66} & \score{5.513}{0.71} & \score{5.947}{0.71} & \score{5.247}{0.53} & \score{5.683}{0.75} & \score{5.668}{0.27} \\
 &  & Our Judge & \cellcolor{green!30}\textbf{\score{5.778}{0.42}} & \cellcolor{green!30}\textbf{\score{5.824}{0.57}} & \cellcolor{green!20}\textbf{\score{6.222}{0.53}} & \cellcolor{green!30}\textbf{\score{5.913}{0.28}} & \score{6.800}{0.40} & \cellcolor{green!30}\textbf{\score{5.667}{1.25}} & \cellcolor{green!30}\textbf{\score{6.034}{0.38}} \\
\cmidrule(l){2-10}
 & \multirow{2}{*}{\centering\arraybackslash \textcolor{GoogleBlue}{\textbf{Color and Lighting}}} & Human & \score{5.515}{0.75} & \score{5.855}{0.71} & \score{5.442}{0.86} & \score{5.549}{0.72} & \score{5.403}{0.83} & \score{5.553}{0.64} & \score{5.553}{0.15} \\
 &  & Our Judge & \cellcolor{green!30}\textbf{\score{5.111}{0.99}} & \cellcolor{green!30}\textbf{\score{5.735}{1.04}} & \cellcolor{green!30}\textbf{\score{5.556}{0.83}} & \cellcolor{green!30}\textbf{\score{5.391}{0.82}} & \score{6.800}{0.40} & \cellcolor{green!30}\textbf{\score{5.333}{1.25}} & \cellcolor{green!30}\textbf{\score{5.654}{0.55}} \\
\cmidrule(l){2-10}
 & \multirow{2}{*}{\centering\arraybackslash \textcolor{GoogleBlue}{\textbf{Seamlessness}}} & Human & \score{5.722}{0.64} & \score{6.101}{0.58} & \score{5.767}{0.74} & \score{5.598}{0.83} & \score{5.357}{1.00} & \score{5.578}{0.73} & \score{5.687}{0.23} \\
 &  & Our Judge & \cellcolor{green!30}\textbf{\score{5.333}{1.15}} & \score{5.265}{1.27} & \cellcolor{green!30}\textbf{\score{5.889}{0.87}} & \cellcolor{green!30}\textbf{\score{5.391}{0.87}} & \score{6.200}{0.40} & \cellcolor{green!20}\textbf{\score{4.833}{1.57}} & \cellcolor{green!30}\textbf{\score{5.485}{0.44}} \\
\midrule
\multirow{6}{*}{\rotatebox[origin=c]{90}{\textcolor{GoogleRed}{\textsc{\sc \bfseries Instruct. Fidel.}}}} & \multirow{2}{*}{\centering\arraybackslash \textcolor{GoogleRed}{\textbf{Alignment}}} & Human & \score{5.556}{0.49} & \score{5.927}{0.97} & \score{5.681}{0.65} & \score{5.666}{1.13} & \score{3.437}{1.40} & \score{5.178}{0.99} & \score{5.241}{0.84} \\
 &  & Our Judge & \score{7.000}{0.00} & \cellcolor{green!20}\textbf{\score{6.471}{1.24}} & \score{6.556}{0.76} & \score{6.609}{1.17} & \score{5.800}{1.94} & \score{6.000}{1.83} & \score{6.406}{0.40} \\
\cmidrule(l){2-10}
 & \multirow{2}{*}{\centering\arraybackslash \textcolor{GoogleRed}{\textbf{Completeness}}} & Human & \score{5.693}{0.72} & \score{5.966}{1.17} & \score{5.789}{0.71} & \score{5.719}{1.14} & \score{3.537}{1.74} & \score{5.556}{0.66} & \score{5.376}{0.83} \\
 &  & Our Judge & \score{7.000}{0.00} & \cellcolor{green!30}\textbf{\score{6.235}{1.54}} & \cellcolor{green!30}\textbf{\score{6.278}{1.19}} & \score{6.609}{1.28} & \score{6.000}{2.00} & \cellcolor{green!30}\textbf{\score{6.000}{1.83}} & \score{6.354}{0.35} \\
\cmidrule(l){2-10}
 & \multirow{2}{*}{\centering\arraybackslash \textcolor{GoogleRed}{\textbf{Plausibility}}} & Human & \score{5.209}{1.03} & \score{6.023}{0.78} & \score{5.743}{0.80} & \score{5.692}{1.17} & \score{4.917}{1.08} & \score{5.586}{0.89} & \score{5.528}{0.36} \\
 &  & Our Judge & \score{6.889}{0.31} & \cellcolor{green!20}\textbf{\score{6.706}{0.75}} & \score{7.000}{0.00} & \score{6.783}{1.02} & \score{7.000}{0.00} & \score{6.833}{0.37} & \score{6.868}{0.11} \\
\midrule
 & \textbf{Average} & Human & \score{5.499}{0.17} & \score{5.992}{0.15} & \score{5.610}{0.24} & \score{5.478}{0.50} & \score{4.673}{0.78} & \score{5.557}{0.26} & \score{5.652}{0.47} \\
 &  & Our Judge & \cellcolor{green!20}\textbf{\score{6.185}{0.69}} & \cellcolor{green!30}\textbf{\score{6.032}{0.48}} & \cellcolor{green!20}\textbf{\score{6.245}{0.36}} & \cellcolor{green!30}\textbf{\score{5.971}{0.58}} & \score{5.983}{0.95} & \cellcolor{green!30}\textbf{\score{5.667}{0.60}} & \cellcolor{green!30}\textbf{\score{6.046}{0.57}} \\
\bottomrule
\end{tabular}
\end{table*}

\subsubsection{Varying Prompts}

Refer to Table~\ref{tab:human_judge_v1}, Table~\ref{tab:human_judge_v3} and Table~\ref{tab:human_judge_combined_category} for a comprehensive comparison between human evaluations and our MLLM-as-a-Judge under different prompt designs. Overall, the Main evaluator(Figure~\ref{fig:prompt-all-factors}) demonstrated the strongest general alignment with human evaluations across factors and edit types, as evidenced by the noticeably larger number of green-highlighted cells in these tables. Since green regions indicate small absolute differences between judge and human scores, this pattern suggested that the Main evaluator most consistently reproduced human rating behavior at the factor level.

As reported in Table~\ref{tab:human_judge_pointwise}, our Main evaluator (Fig~\ref{fig:prompt-all-factors}) achieved consistently low MSE and MAE across a wide range of evaluation factors, by observing that MAE is less than 1 across nearly all factors indicating a close alignment with human absolute score annotations. 
However, it should be noticed that in some cases MAE is less than 1 but MSE is larger than 1 like for factor Global Consistency, Scale Realism, and Alignment. This means that although the judge performs well generally, the performance is not good as expected in some difficult cases. The Main evaluator also exhibited stable accuracy and tolerance-based accuracy (ACC and ACC$\pm$1) across most factors, with ACC close to $0.3$ in most cases and ACC$\pm$1 reaching almost $0.9$ in some factors, suggesting reliable pointwise agreement with human judgments.

Beyond error-based metrics, the Factor evaluator showed meaningful positive correlations with human annotations for several perceptual and structural factors (Table~\ref{tab:human_judge_v1}), notably Spatial Relationship, Image Quality, Color and Lighting, and Seamlessness, reflecting its ability to preserve relative ranking and preference structure in human evaluations. This behavior was further supported by the pairwise preference prediction results in Table~\ref{tab:human_judge_pairwise}, where the Factor evaluator attained strong accuracy across multiple factors and a competitive overall weighted accuracy. 

A more fine-grained analysis of the pointwise and pairwise evaluation results in Table~\ref{tab:human_judge_pointwise} and Table~\ref{tab:human_judge_pairwise} shows that in terms of pointwise error metrics (MSE and MAE) and pairwise preference prediction accuracy, the Category wise, Example guided prompt (Figure~\ref{fig:prompt-mllm-as-a-judge-image-editing, category}) often achieved the strongest numerical performance across multiple factors, particularly for instruction-fidelity dimensions such as Alignment and Completeness. However, despite its favorable quantitative results, the Category wise, Example guided introduced a substantially longer and more complex instruction format that was impractical for human evaluators, as it imposed a higher cognitive load and increased the likelihood of annotation fatigue. In contrast, the Main evaluator mirrored the human evaluation setup in both structure and level of detail, enabling a more direct and fair comparison between human and judge assessments. Consequently, when the objective was to evaluate alignment with realistic human evaluation behavior rather than optimizing raw agreement metrics in isolation, the Main evaluator emerged as the most suitable and representative configuration.

Finally, while the Factor-level Rubrics prompt (Figure~\ref{fig:prompt-mllm-as-a-judge-image-editing, main}) underperformed relative to the other two variants in many cases while adding a detailed rubric on the base of Main evaluator, the underlying causes of this degradation presented an interesting direction for future investigation and were therefore deferred to the ablation study. We think this phenomena is interesting and worse to dug deeper in, but we won't discuss about this much in this paper.

\subsection{Traditional Metrics Results Comparison}

\subsubsection{Varying MLLMs}
For this section, we will compare the performance of two different MLLMs, GPT-5-mini and Gemini-2.5-pro. The prompts we used is the Factor-level Rubrics prompt as can be seen in Appendix~\ref{prompts-sec} Figure~\ref{fig:prompt-mllm-as-a-judge-image-editing, main}.
The results for offline setting can be seen in Table~\ref{tab:L1_L2_gpt_gemini_image_preservation_offline}, Table~\ref{tab:L1_L2_gpt_gemini_edit_quality_offline}, Table~\ref{tab:L1_L2_gpt_gemini_instruction_fidelity_offline},Table~\ref{tab:Mask_SSIM_gpt_gemini_image_preservation_offline}, Table~\ref{tab:Mask_SSIM_gpt_gemini_edit_quality_offline}, Table~\ref{tab:Mask_SSIM_gpt_gemini_instruction_fidelity_offline}, Table~\ref{tab:clip_dino_gpt_gemini_image_preservation_offline}, Table~\ref{tab:clip_dino_gpt_gemini_edit_quality_offline}, and Table~\ref{tab:clip_dino_gpt_gemini_instruction_fidelity_offline}.
Alternatively, the results for online setting can be seen in Table~\ref{tab:L1_L2_gpt_gemini_image_preservation_online}, Table~\ref{tab:L1_L2_gpt_gemini_edit_quality_online}, Table~\ref{tab:L1_L2_gpt_gemini_instruction_fidelity_online}, Table~\ref{tab:Mask_SSIM_gpt_gemini_image_preservation_online}, Table~\ref{tab:Mask_SSIM_gpt_gemini_edit_quality_online}, Table~\ref{tab:Mask_SSIM_gpt_gemini_instruction_fidelity_online}, Table~\ref{tab:clip_dino_gpt_gemini_image_preservation_online}, Table~\ref{tab:clip_dino_gpt_gemini_edit_quality_online}, and Table~\ref{tab:clip_dino_gpt_gemini_instruction_fidelity_online}

\section{AI Usage}
For this work, we used AI for paraphrasing and polishing our original essay. The use of tools only assists with proofreading.

\section{Benchmark License}
Our dataset is released under a non-commercial, research-only license that permits use, redistribution, and modification for academic and internal research purposes, while prohibiting commercial use. The dataset inherits and complies with the licensing terms of HumanEdit (CC-BY 4.0), and all annotations we contribute are released under the same or a compatible license to avoid downstream ambiguity.

\clearpage

\setlength{\LTpre}{0pt}
\setlength{\LTpost}{0pt}

\newcolumntype{Y}{>{\centering\arraybackslash}X}

\scriptsize
\setlength{\tabcolsep}{4pt}

\begin{table*}[t]
\centering
\caption{
Results of Image Preservation showing five evaluation metrics for all factors using traditional metrics
(L1, L2, PSNR, SSIM, LPIPS) (OFFLINE setting, refer to Figure \ref{fig:prompt-mllm-as-a-judge-image-editing, main} for prompt).
Evaluation measures score prediction (MSE, MAE) and ranking quality
(Pearson, Spearman, Kendall $\tau$).
Scores are reported for both GPT and Gemini.
}
\tiny
\label{tab:L1_L2_gpt_gemini_image_preservation_offline}
\begin{tabular}{@{}c c l c *{5}{c}@{}}
\toprule
& \textbf{Factor} &
\makecell{\textbf{Metric}} &
\makecell{\textbf{Model}} &
\methead{MSE $\downarrow$} & \methead{MAE $\downarrow$} &
\methead{Pearson $\uparrow$} & \methead{Spearman $\uparrow$} &
\methead{Kendall $\tau$ $\uparrow$} \\
\midrule
\multirow{39}{*}{\rotatebox[origin=c]{90}{\textcolor{GoogleGreen}{\textsc{\bfseries Image Preservation}}}}
& \multirow{13}{*}{\textcolor{GoogleGreen}{\textbf{Unchanged Regions}}}
& \multirow{2}{*}{L1 ($\downarrow$)} & GPT
& 0.320 & 0.504 & -0.252 & \textbf{-0.212} & \textbf{-0.176} \\
& & & Gemini
& 0.366 & 0.547 & -0.197 & -0.225 & -0.180 \\
\cmidrule(lr){3-9}
& & \multirow{2}{*}{L2 ($\downarrow$)} & GPT
& 0.365 & 0.535 & -0.211 & -0.179 & -0.138 \\
& & & Gemini
& 0.406 & 0.568 & -0.190 & -0.158 & -0.127 \\
\cmidrule(lr){3-9}
& & \multirow{2}{*}{PSNR ($\uparrow$)} & GPT
& 0.183 & \textbf{0.333} & 0.234 & 0.179 & 0.138 \\
& & & Gemini
& 0.261 & 0.434 & 0.081 & 0.158 & 0.127 \\
\cmidrule(lr){3-9}
& & \multirow{2}{*}{SSIM ($\uparrow$)} & GPT
& \textbf{0.175} & 0.338 & \textbf{0.264} & 0.180 & 0.138 \\
\bottomrule
\end{tabular}
\end{table*}

\begin{table*}[t]
\centering
\caption{
Results of Edit Quality showing five evaluation metrics for all factors using traditional metrics
(L1, L2, PSNR, SSIM, LPIPS) (OFFLINE setting, refer to Figure \ref{fig:prompt-mllm-as-a-judge-image-editing, main} for prompt).
Evaluation measures score prediction (MSE, MAE) and ranking quality
(Pearson, Spearman, Kendall $\tau$).
Scores are reported for both GPT and Gemini.
}
\tiny
\label{tab:L1_L2_gpt_gemini_edit_quality_offline}
\begin{tabular}{@{}c c l c *{5}{c}@{}}
\toprule
& \textbf{Factor} &
\makecell{\textbf{Metric}} &
\makecell{\textbf{Model}} &
\methead{MSE $\downarrow$} & \methead{MAE $\downarrow$} &
\methead{Pearson $\uparrow$} & \methead{Spearman $\uparrow$} &
\methead{Kendall $\tau$ $\uparrow$} \\
\midrule

\multirow{78}{*}{\rotatebox[origin=c]{90}{\textcolor{GoogleBlue}{\textsc{\bfseries Edit Quality}}}}
& \multirow{13}{*}{\textcolor{GoogleBlue}{\textbf{Scale Realism}}}
& \multirow{2}{*}{L1 ($\downarrow$)} & GPT
& 0.212 & 0.394 & 0.077 & 0.007 & 0.015 \\
& & & Gemini
& 0.440 & 0.616 & -0.463 & -0.461 & -0.368 \\
\cmidrule(lr){3-9}

& & \multirow{2}{*}{L2 ($\downarrow$)} & GPT
& 0.240 & 0.413 & 0.098 & 0.049 & 0.036 \\
& & & Gemini
& 0.514 & 0.675 & -0.555 & -0.558 & -0.429 \\
\cmidrule(lr){3-9}

& & \multirow{2}{*}{PSNR ($\uparrow$)} & GPT
& 0.226 & 0.407 & \textbf{-0.123} & -0.049 & -0.036 \\
& & & Gemini
& 0.192 & 0.373 & 0.491 & 0.558 & 0.429 \\
\cmidrule(lr){3-9}

& & \multirow{2}{*}{SSIM ($\uparrow$)} & GPT
& 0.189 & 0.354 & 0.050 & \textbf{0.119} & \textbf{0.092} \\
\bottomrule
\end{tabular}
\end{table*}

\begin{table*}[t]
\centering
\caption{
Results of Instruction Fidelity showing five evaluation metrics for all factors using traditional metrics
(L1, L2, PSNR, SSIM, LPIPS) (OFFLINE setting, refer to Figure \ref{fig:prompt-mllm-as-a-judge-image-editing, main} for prompt).
Evaluation measures score prediction (MSE, MAE) and ranking quality
(Pearson, Spearman, Kendall $\tau$).
Scores are reported for both GPT and Gemini.
}
\tiny
\label{tab:L1_L2_gpt_gemini_instruction_fidelity_offline}
\begin{tabular}{@{}c c l c *{5}{c}@{}}
\toprule
& \textbf{Factor} &
\makecell{\textbf{Metric}} &
\makecell{\textbf{Model}} &
\methead{MSE $\downarrow$} & \methead{MAE $\downarrow$} &
\methead{Pearson $\uparrow$} & \methead{Spearman $\uparrow$} &
\methead{Kendall $\tau$ $\uparrow$} \\
\midrule
\multirow{39}{*}{\rotatebox[origin=c]{90}{\textcolor{GoogleRed}{\textsc{\bfseries Instruction Fidelity}}}}
& \multirow{13}{*}{\textcolor{GoogleRed}{\textbf{Alignment}}}
& \multirow{2}{*}{L1 ($\downarrow$)} & GPT
& 0.304 & 0.471 & -0.037 & -0.015 & -0.008 \\
& & & Gemini
& 0.173 & 0.329 & -0.161 & -0.107 & -0.086 \\
\cmidrule(lr){3-9}
& & \multirow{2}{*}{L2 ($\downarrow$)} & GPT
& 0.361 & 0.517 & -0.083 & \textbf{-0.145} & \textbf{-0.098} \\
& & & Gemini
& 0.152 & 0.307 & -0.173 & -0.179 & -0.135 \\
\cmidrule(lr){3-9}
& & \multirow{2}{*}{PSNR ($\uparrow$)} & GPT
& \textbf{0.220} & \textbf{0.385} & \textbf{0.097} & 0.145 & 0.098 \\
& & & Gemini
& 0.206 & 0.375 & 0.149 & 0.179 & 0.135 \\
\cmidrule(lr){3-9}
& & \multirow{2}{*}{SSIM ($\uparrow$)} & GPT
& 0.251 & 0.419 & -0.094 & -0.053 & -0.042 \\
\bottomrule
\end{tabular}
\end{table*}

\begin{table*}[t]
\centering
\caption{
Results of Image Preservation showing five evaluation metrics for all factors
using traditional metrics (L1, L2, PSNR, SSIM, LPIPS) (ONLINE setting, refer to Figure \ref{fig:prompt-mllm-as-a-judge-image-editing, main} for prompt)).
Evaluation measures score prediction (MSE, MAE)
and ranking quality (Pearson, Spearman, Kendall $\tau$).
Scores are reported for both GPT and Gemini.
}
\label{tab: detailed-online-pixel}
\tiny
\label{tab:L1_L2_gpt_gemini_image_preservation_online}
\begin{tabular}{@{}c c l c *{5}{c}@{}}
\toprule
& \textbf{Factor} &
\makecell{\textbf{Metric}} &
\makecell{\textbf{Model}} &
\methead{MSE $\downarrow$} & \methead{MAE $\downarrow$} &
\methead{Pearson $\uparrow$} & \methead{Spearman $\uparrow$} &
\methead{Kendall $\tau$ $\uparrow$} \\
\midrule
\multirow{39}{*}{\rotatebox[origin=c]{90}{\textcolor{GoogleGreen}{\textsc{\bfseries Image Preservation}}}}
& \multirow{13}{*}{\textcolor{GoogleGreen}{\textbf{Unchanged Regions}}}
& \multirow{2}{*}{L1 ($\downarrow$)} & GPT
& 0.345 & 0.526 & \textbf{0.181} & \textbf{0.164} & \textbf{0.136} \\
& & & Gemini
& 0.343 & 0.510 & 0.075 & -0.002 & 0.004 \\
\cmidrule(lr){3-9}
& & \multirow{2}{*}{L2 ($\downarrow$)} & GPT
& 0.408 & 0.575 & 0.127 & 0.106 & 0.088 \\
& & & Gemini
& 0.389 & 0.551 & 0.122 & 0.100 & 0.076 \\
\cmidrule(lr){3-9}
& & \multirow{2}{*}{PSNR ($\uparrow$)} & GPT
& 0.369 & 0.550 & -0.116 & -0.106 & -0.088 \\
& & & Gemini
& 0.346 & 0.521 & -0.142 & -0.100 & -0.076 \\
\cmidrule(lr){3-9}
& & \multirow{2}{*}{SSIM ($\uparrow$)} & GPT
& 0.342 & 0.523 & -0.072 & -0.067 & -0.056 \\
\bottomrule
\end{tabular}
\end{table*}

\begin{table*}[t]
\centering
\vspace{-5mm}
\caption{
Results of Edit Quality showing five evaluation metrics for all factors
using traditional metrics (L1, L2, PSNR, SSIM, LPIPS) (ONLINE setting, refer to Figure \ref{fig:prompt-mllm-as-a-judge-image-editing, main} for prompt)).
Evaluation measures score prediction (MSE, MAE)
and ranking quality (Pearson, Spearman, Kendall $\tau$).
Scores are reported for both GPT and Gemini.
}
\tiny
\label{tab:L1_L2_gpt_gemini_edit_quality_online}

\begin{tabular}{l l c *{5}{c} }
\toprule
\textbf{Factor} & \textbf{Metric} & \textbf{Model} &
\methead{MSE $\downarrow$} &
\methead{MAE $\downarrow$} &
\methead{Pearson $\uparrow$} &
\methead{Spearman $\uparrow$} &
\methead{Kendall $\tau$ $\uparrow$} \\
\midrule

\multirow{13}{*}{\textcolor{GoogleBlue}{\textbf{Overall}}}
& \multirow{2}{*}{L1 ($\downarrow$)} & GPT
& 0.309 & 0.498 & 0.195 & 0.105 & 0.078 \\
& & Gemini
& 0.388 & 0.571 & -0.200 & -0.232 & -0.184 \\
\cmidrule(lr){2-8}

& \multirow{2}{*}{L2 ($\downarrow$)} & GPT
& 0.366 & 0.536 & 0.221 & \textbf{0.177} & \textbf{0.130} \\
& & Gemini
& 0.443 & 0.615 & -0.194 & -0.196 & -0.169 \\
\cmidrule(lr){2-8}

& \multirow{2}{*}{PSNR ($\uparrow$)} & GPT
& 0.338 & 0.525 & \textbf{-0.322} & -0.177 & -0.130 \\
& & Gemini
& 0.296 & 0.480 & 0.132 & 0.204 & 0.177 \\
\cmidrule(lr){2-8}

& \multirow{2}{*}{SSIM ($\uparrow$)} & GPT
& 0.271 & 0.462 & 0.010 & 0.009 & 0.026 \\
\bottomrule
\end{tabular}
\end{table*}

\begin{table*}[t]
\centering
\caption{
Results of Instruction Fidelity showing five evaluation metrics for all factors
using traditional metrics (L1, L2, PSNR, SSIM, LPIPS) (ONLINE setting, refer to Figure \ref{fig:prompt-mllm-as-a-judge-image-editing, main} for prompt)).
Evaluation measures score prediction (MSE, MAE)
and ranking quality (Pearson, Spearman, Kendall $\tau$).
Scores are reported for both GPT and Gemini.
}
\tiny
\label{tab:L1_L2_gpt_gemini_instruction_fidelity_online}
\begin{tabular}{@{}c c l c *{5}{c}@{}}
\toprule
& \textbf{Factor} &
\makecell{\textbf{Metric}} &
\makecell{\textbf{Model}} &
\methead{MSE $\downarrow$} & \methead{MAE $\downarrow$} &
\methead{Pearson $\uparrow$} & \methead{Spearman $\uparrow$} &
\methead{Kendall $\tau$ $\uparrow$} \\
\midrule

\multirow{39}{*}{\rotatebox[origin=c]{90}{\textcolor{GoogleRed}{\textsc{\bfseries Instruction Fidelity}}}}

& \multirow{13}{*}{\textcolor{GoogleRed}{\textbf{Alignment}}}
& \multirow{2}{*}{L1 ($\downarrow$)} & GPT
& 0.378 & 0.547 & -0.004 & -0.061 & -0.054 \\
& & & Gemini
& 0.403 & 0.591 & -0.019 & 0.018 & 0.019 \\
\cmidrule(lr){3-9}

& & \multirow{2}{*}{L2 ($\downarrow$)} & GPT
& 0.440 & 0.593 & 0.013 & -0.104 & -0.071 \\
& & & Gemini
& 0.471 & 0.646 & 0.006 & 0.067 & 0.057 \\
\cmidrule(lr){3-9}

& & \multirow{2}{*}{PSNR ($\uparrow$)} & GPT
& 0.331 & 0.513 & 0.007 & 0.104 & 0.071 \\
& & & Gemini
& 0.353 & 0.539 & -0.030 & -0.067 & -0.057 \\
\cmidrule(lr){3-9}

& & \multirow{2}{*}{SSIM ($\uparrow$)} & GPT
& 0.320 & 0.508 & -0.083 & 0.073 & 0.071 \\
\bottomrule
\end{tabular}
\end{table*}

\begin{table*}[t]
\centering
\caption{
Results of Image Preservation showing evaluation metrics
using mask-based metrics (Mask SSIM, Mask LPIPS, Background Consistency) (OFFLINE setting, refer to Figure \ref{fig:prompt-mllm-as-a-judge-image-editing, main} for prompt)).
Evaluation measures score prediction (MSE, MAE)
and ranking quality (Pearson, Spearman, Kendall $\tau$).
Scores are reported for both GPT and Gemini.
}
\tiny
\label{tab:Mask_SSIM_gpt_gemini_image_preservation_offline}
\begin{tabular}{@{} l l c *{5}{c} @{}}
\toprule
\textbf{Factor} & \textbf{Metric} & \textbf{Model} &
\methead{MSE $\downarrow$} &
\methead{MAE $\downarrow$} &
\methead{Pearson $\uparrow$} &
\methead{Spearman $\uparrow$} &
\methead{Kendall $\tau$ $\uparrow$} \\
\midrule
\multirow{7}{*}{\textcolor{GoogleGreen}{\textbf{Unchanged Regions}}}
& \multirow{2}{*}{Mask SSIM ($\uparrow$)} & GPT
& \textbf{0.175} & \textbf{0.338} & \textbf{0.264} & \textbf{0.180} & \textbf{0.138} \\
& & Gemini
& 0.276 & 0.442 & -0.004 & -0.022 & -0.009 \\
\cmidrule(lr){2-8}
& \multirow{2}{*}{Mask LPIPS ($\downarrow$)} & GPT
& 0.199 & 0.385 & 0.036 & 0.104 & 0.073 \\
& & Gemini
& 0.253 & 0.454 & 0.042 & -0.044 & -0.027 \\
\cmidrule(lr){2-8}
& \multirow{2}{*}{BG Consistency ($\uparrow$)} & GPT
& \textbf{0.175} & \textbf{0.338} & \textbf{0.264} & \textbf{0.180} & \textbf{0.138} \\
& & Gemini
& 0.276 & 0.442 & -0.004 & -0.022 & -0.009 \\
\cmidrule(lr){2-8}
& \multirow{2}{*}{Avg} & GPT
& 0.183 & 0.354 & 0.188 & 0.155 & 0.116 \\
\bottomrule
\end{tabular}
\end{table*}

\begin{table*}[t]
\centering
\caption{
Results of Edit Quality showing evaluation metrics
using mask-based metrics (Mask SSIM, Mask LPIPS, Background Consistency) (OFFLINE setting, refer to Figure \ref{fig:prompt-mllm-as-a-judge-image-editing, main} for prompt)).
Evaluation measures score prediction (MSE, MAE)
and ranking quality (Pearson, Spearman, Kendall $\tau$).
Scores are reported for both GPT and Gemini.
}
\tiny
\label{tab:Mask_SSIM_gpt_gemini_edit_quality_offline}

\begin{tabular}{@{} l l c *{5}{c} @{}}
\toprule
\textbf{Factor} & \textbf{Metric} & \textbf{Model} &
\methead{MSE $\downarrow$} &
\methead{MAE $\downarrow$} &
\methead{Pearson $\uparrow$} &
\methead{Spearman $\uparrow$} &
\methead{Kendall $\tau$ $\uparrow$} \\
\midrule

\multirow{7}{*}{\textcolor{GoogleBlue}{\textbf{Scale Realism}}}
& \multirow{2}{*}{Mask SSIM ($\uparrow$)} & GPT
& 0.189 & 0.354 & 0.050 & \textbf{0.119} & \textbf{0.092} \\
& & Gemini
& 0.232 & 0.403 & 0.266 & 0.339 & 0.266 \\
\cmidrule(lr){2-8}

& \multirow{2}{*}{Mask LPIPS ($\downarrow$)} & GPT
& \textbf{0.173} & \textbf{0.346} & \textbf{0.059} & -0.024 & -0.010 \\
& & Gemini
& 0.292 & 0.466 & -0.153 & -0.188 & -0.143 \\
\cmidrule(lr){2-8}

& \multirow{2}{*}{BG Consistency ($\uparrow$)} & GPT
& 0.189 & 0.354 & 0.050 & \textbf{0.119} & \textbf{0.092} \\
& & Gemini
& 0.232 & 0.403 & 0.266 & 0.339 & 0.266 \\
\cmidrule(lr){2-8}

& \multirow{2}{*}{Avg} & GPT
& 0.184 & 0.351 & 0.053 & 0.071 & 0.058 \\
\bottomrule
\end{tabular}
\end{table*}

\begin{table*}[t]
\centering
\caption{
Results of Instruction Fidelity showing evaluation metrics
using mask-based metrics (Mask SSIM, Mask LPIPS, Background Consistency) (OFFLINE setting, refer to Figure \ref{fig:prompt-mllm-as-a-judge-image-editing, main} for prompt)).
Evaluation measures score prediction (MSE, MAE)
and ranking quality (Pearson, Spearman, Kendall $\tau$).
Scores are reported for both GPT and Gemini.
}
\tiny
\label{tab:Mask_SSIM_gpt_gemini_instruction_fidelity_offline}
\begin{tabular}{@{} l l c *{5}{c} @{}}
\toprule
\textbf{Factor} & \textbf{Metric} & \textbf{Model} &
\methead{MSE $\downarrow$} &
\methead{MAE $\downarrow$} &
\methead{Pearson $\uparrow$} &
\methead{Spearman $\uparrow$} &
\methead{Kendall $\tau$ $\uparrow$} \\
\midrule
\multirow{7}{*}{\textcolor{GoogleRed}{\textbf{Alignment}}}
& \multirow{2}{*}{Mask SSIM ($\uparrow$)} & GPT
& 0.251 & 0.419 & \textbf{-0.094} & -0.053 & -0.042 \\
& & Gemini
& 0.182 & 0.359 & 0.288 & 0.282 & 0.224 \\
\cmidrule(lr){2-8}
& \multirow{2}{*}{Mask LPIPS ($\downarrow$)} & GPT
& \textbf{0.223} & \textbf{0.389} & -0.010 & \textbf{0.122} & \textbf{0.093} \\
& & Gemini
& 0.238 & 0.404 & -0.277 & -0.217 & -0.158 \\
\cmidrule(lr){2-8}
& \multirow{2}{*}{BG Consistency ($\uparrow$)} & GPT
& 0.251 & 0.419 & \textbf{-0.094} & -0.053 & -0.042 \\
& & Gemini
& 0.182 & 0.359 & 0.288 & 0.282 & 0.224 \\
\cmidrule(lr){2-8}
& \multirow{2}{*}{Avg} & GPT
& 0.242 & 0.409 & -0.066 & 0.005 & 0.003 \\
\bottomrule
\end{tabular}
\end{table*}

\begin{table*}[t]
\centering
\caption{
Results of Image Preservation using mask-based metrics
(Mask SSIM, Mask LPIPS, Background Consistency) (ONLINE setting, refer to Figure \ref{fig:prompt-mllm-as-a-judge-image-editing, main} for prompt)).
Evaluation measures score prediction (MSE, MAE)
and ranking quality (Pearson, Spearman, Kendall $\tau$).
Scores are reported for both GPT and Gemini.
}
\tiny
\label{tab:Mask_SSIM_gpt_gemini_image_preservation_online}
\begin{tabular}{@{} c l l c *{5}{c} @{}}
\toprule
& \textbf{Factor} & \textbf{Metric} & \textbf{Model} &
\methead{MSE $\downarrow$} &
\methead{MAE $\downarrow$} &
\methead{Pearson $\uparrow$} &
\methead{Spearman $\uparrow$} &
\methead{Kendall $\tau$ $\uparrow$} \\
\midrule
\multirow{21}{*}{\rotatebox[origin=c]{90}{\textcolor{GoogleGreen}{\textsc{\bfseries Image Preservation}}}}
& \multirow{7}{*}{\textcolor{GoogleGreen}{\textbf{Unchanged Regions}}}
& \multirow{2}{*}{Mask SSIM ($\uparrow$)} & GPT
& 0.342 & 0.523 & \textbf{-0.072} & \textbf{-0.067} & \textbf{-0.056} \\
& & & Gemini
& 0.261 & 0.445 & 0.214 & 0.260 & 0.205 \\
\cmidrule(lr){3-9}
& & \multirow{2}{*}{Mask LPIPS ($\downarrow$)} & GPT
& \textbf{0.269} & \textbf{0.437} & 0.030 & 0.029 & 0.024 \\
& & & Gemini
& 0.290 & 0.475 & -0.219 & -0.272 & -0.220 \\
\cmidrule(lr){3-9}
& & \multirow{2}{*}{BG Consistency ($\uparrow$)} & GPT
& 0.342 & 0.523 & \textbf{-0.072} & \textbf{-0.067} & \textbf{-0.056} \\
& & & Gemini
& 0.261 & 0.445 & 0.214 & 0.260 & 0.205 \\
\cmidrule(lr){3-9}
& & \multirow{2}{*}{Avg} & GPT
& 0.318 & 0.494 & -0.038 & -0.035 & -0.029 \\
\bottomrule
\end{tabular}
\end{table*}

\begin{table*}[t]
\centering
\caption{
Results of Edit Quality using mask-based metrics
(Mask SSIM, Mask LPIPS, Background Consistency) (ONLINE setting, refer to Figure \ref{fig:prompt-mllm-as-a-judge-image-editing, main} for prompt)).
Evaluation measures score prediction (MSE, MAE)
and ranking quality (Pearson, Spearman, Kendall $\tau$).
Scores are reported for both GPT and Gemini.
}
\tiny
\label{tab:Mask_SSIM_gpt_gemini_edit_quality_online}

\begin{tabular}{@{} c l l c *{5}{c} @{}}
\toprule
& \textbf{Factor} & \textbf{Metric} & \textbf{Model} &
\methead{MSE $\downarrow$} &
\methead{MAE $\downarrow$} &
\methead{Pearson $\uparrow$} &
\methead{Spearman $\uparrow$} &
\methead{Kendall $\tau$ $\uparrow$} \\
\midrule

\multirow{42}{*}{\rotatebox[origin=c]{90}{\textcolor{GoogleBlue}{\textsc{\bfseries Edit Quality}}}}

& \multirow{7}{*}{\textcolor{GoogleBlue}{\textbf{Scale Realism}}}
& \multirow{2}{*}{Mask SSIM ($\uparrow$)} & GPT
& 0.271 & 0.462 & 0.010 & 0.009 & 0.026 \\
& & & Gemini
& 0.256 & 0.450 & 0.245 & 0.316 & 0.256 \\
\cmidrule(lr){3-9}

& & \multirow{2}{*}{Mask LPIPS ($\downarrow$)} & GPT
& \textbf{0.202} & \textbf{0.374} & \textbf{0.051} & \textbf{-0.108} & \textbf{-0.085} \\
& & & Gemini
& 0.256 & 0.433 & -0.119 & -0.238 & -0.177 \\
\cmidrule(lr){3-9}

& & \multirow{2}{*}{BG Consistency ($\uparrow$)} & GPT
& 0.271 & 0.462 & 0.010 & 0.009 & 0.026 \\
& & & Gemini
& 0.256 & 0.450 & 0.245 & 0.316 & 0.256 \\
\cmidrule(lr){3-9}

& & \multirow{2}{*}{Avg} & GPT
& 0.248 & 0.433 & 0.024 & -0.030 & -0.011 \\
\bottomrule
\end{tabular}
\end{table*}

\begin{table*}[t]
\centering
\caption{
Results of Instruction Fidelity using mask-based metrics
(Mask SSIM, Mask LPIPS, Background Consistency) (ONLINE setting, refer to Figure \ref{fig:prompt-mllm-as-a-judge-image-editing, main} for prompt)).
Evaluation measures score prediction (MSE, MAE)
and ranking quality (Pearson, Spearman, Kendall $\tau$).
Scores are reported for both GPT and Gemini.
}
\tiny
\label{tab:Mask_SSIM_gpt_gemini_instruction_fidelity_online}
\begin{tabular}{@{} c l l c *{5}{c} @{}}
\toprule
& \textbf{Factor} & \textbf{Metric} & \textbf{Model} &
\methead{MSE $\downarrow$} &
\methead{MAE $\downarrow$} &
\methead{Pearson $\uparrow$} &
\methead{Spearman $\uparrow$} &
\methead{Kendall $\tau$ $\uparrow$} \\
\midrule
\multirow{21}{*}{\rotatebox[origin=c]{90}{\textcolor{GoogleRed}{\textsc{\bfseries Instruction Fidelity}}}}
& \multirow{7}{*}{\textcolor{GoogleRed}{\textbf{Alignment}}}
& \multirow{2}{*}{Mask SSIM ($\uparrow$)} & GPT
& 0.320 & 0.508 & -0.083 & 0.073 & 0.071 \\
& & & Gemini
& 0.312 & 0.494 & 0.103 & 0.041 & 0.032 \\
\cmidrule(lr){3-9}
& & \multirow{2}{*}{Mask LPIPS ($\downarrow$)} & GPT
& \textbf{0.184} & \textbf{0.343} & \textbf{0.359} & \textbf{0.140} & \textbf{0.112} \\
& & & Gemini
& 0.229 & 0.391 & 0.099 & 0.149 & 0.121 \\
\cmidrule(lr){3-9}
& & \multirow{2}{*}{BG Consistency ($\uparrow$)} & GPT
& 0.320 & 0.508 & -0.083 & 0.073 & 0.071 \\
& & & Gemini
& 0.312 & 0.494 & 0.103 & 0.041 & 0.032 \\
\cmidrule(lr){3-9}
& & \multirow{2}{*}{Avg} & GPT
& 0.275 & 0.453 & 0.064 & 0.095 & 0.085 \\
\bottomrule
\end{tabular}
\end{table*}

\begin{table*}[t]
\centering
\caption{
Results of Image Preservation using semantic similarity metrics
(CLIP Text, CLIP Image, DINO Image) (OFFLINE setting, refer to Figure \ref{fig:prompt-mllm-as-a-judge-image-editing, main} for prompt)).
Evaluation measures score prediction (MSE, MAE)
and ranking quality (Pearson, Spearman, Kendall $\tau$).
Scores are reported for both GPT and Gemini.
}
\tiny
\label{tab:clip_dino_gpt_gemini_image_preservation_offline}
\begin{tabular}{@{}c l l c *{5}{c}@{}}
\toprule
& \textbf{Factor} & \textbf{Metric} & \textbf{Model} &
\methead{MSE $\downarrow$} &
\methead{MAE $\downarrow$} &
\methead{Pearson $\uparrow$} &
\methead{Spearman $\uparrow$} &
\methead{Kendall $\tau$ $\uparrow$} \\
\midrule
\multirow{21}{*}{\rotatebox[origin=c]{90}{\textcolor{GoogleGreen}{\textsc{\bfseries Image Preservation}}}}
& \multirow{7}{*}{\textcolor{GoogleGreen}{\textbf{Unchanged Regions}}}
& \multirow{2}{*}{CLIP Text ($\uparrow$)} & GPT
& \textbf{0.102} & 0.292 & \textbf{0.241} & \textbf{0.224} & \textbf{0.171} \\
& & & Gemini
& 0.172 & 0.398 & 0.245 & 0.203 & 0.144 \\
\cmidrule(lr){3-9}
& & \multirow{2}{*}{CLIP Image ($\uparrow$)} & GPT
& 0.109 & \textbf{0.230} & -0.121 & -0.070 & -0.051 \\
& & & Gemini
& 0.219 & 0.335 & -0.161 & -0.080 & -0.085 \\
\cmidrule(lr){3-9}
& & \multirow{2}{*}{DINO Image ($\uparrow$)} & GPT
& 0.128 & 0.302 & 0.130 & -0.065 & -0.062 \\
& & & Gemini
& 0.240 & 0.403 & -0.062 & -0.078 & -0.062 \\
\cmidrule(lr){3-9}
& & \multirow{2}{*}{Avg} & GPT
& 0.113 & 0.275 & 0.083 & 0.030 & 0.019 \\
\bottomrule
\end{tabular}
\end{table*}

\begin{table*}[t]
\centering
\caption{
Results of Edit Quality using semantic similarity metrics
(CLIP Text, CLIP Image, DINO Image) (OFFLINE setting, refer to Figure \ref{fig:prompt-mllm-as-a-judge-image-editing, main} for prompt)).
Evaluation measures score prediction (MSE, MAE)
and ranking quality (Pearson, Spearman, Kendall $\tau$).
Scores are reported for both GPT and Gemini.
}
\tiny
\label{tab:clip_dino_gpt_gemini_edit_quality_offline}

\begin{tabular}{@{}c l l c *{5}{c}@{}}
\toprule
& \textbf{Factor} & \textbf{Metric} & \textbf{Model} &
\methead{MSE $\downarrow$} &
\methead{MAE $\downarrow$} &
\methead{Pearson $\uparrow$} &
\methead{Spearman $\uparrow$} &
\methead{Kendall $\tau$ $\uparrow$} \\
\midrule

\multirow{42}{*}{\rotatebox[origin=c]{90}{\textcolor{GoogleBlue}{\textsc{\bfseries Edit Quality}}}}

& \multirow{7}{*}{\textcolor{GoogleBlue}{\textbf{Scale Realism}}}
& \multirow{2}{*}{CLIP Text ($\uparrow$)} & GPT
& \textbf{0.114} & 0.293 & 0.180 & 0.137 & 0.118 \\
& & & Gemini
& 0.157 & 0.377 & -0.273 & -0.146 & -0.123 \\
\cmidrule(lr){3-9}

& & \multirow{2}{*}{CLIP Image ($\uparrow$)} & GPT
& 0.207 & 0.361 & -0.156 & -0.120 & -0.082 \\
& & & Gemini
& 0.107 & 0.220 & 0.282 & 0.396 & 0.334 \\
\cmidrule(lr){3-9}

& & \multirow{2}{*}{DINO Image ($\uparrow$)} & GPT
& 0.128 & \textbf{0.278} & \textbf{0.326} & \textbf{0.321} & \textbf{0.250} \\
& & & Gemini
& 0.111 & 0.255 & 0.463 & 0.455 & 0.375 \\
\cmidrule(lr){3-9}

& & \multirow{2}{*}{Avg} & GPT
& 0.150 & 0.311 & 0.117 & 0.113 & 0.095 \\
\bottomrule
\end{tabular}
\end{table*}

\begin{table*}[t]
\centering
\caption{
Results of Instruction Fidelity using semantic similarity metrics
(CLIP Text, CLIP Image, DINO Image) (OFFLINE setting, refer to Figure \ref{fig:prompt-mllm-as-a-judge-image-editing, main} for prompt)).
Evaluation measures score prediction (MSE, MAE)
and ranking quality (Pearson, Spearman, Kendall $\tau$).
Scores are reported for both GPT and Gemini.
}
\tiny
\label{tab:clip_dino_gpt_gemini_instruction_fidelity_offline}
\begin{tabular}{@{}c l l c *{5}{c}@{}}
\toprule
& \textbf{Factor} & \textbf{Metric} & \textbf{Model} &
\methead{MSE $\downarrow$} &
\methead{MAE $\downarrow$} &
\methead{Pearson $\uparrow$} &
\methead{Spearman $\uparrow$} &
\methead{Kendall $\tau$ $\uparrow$} \\
\midrule
\multirow{21}{*}{\rotatebox[origin=c]{90}{\textcolor{GoogleRed}{\textsc{\bfseries Instruction Fidelity}}}}
& \multirow{7}{*}{\textcolor{GoogleRed}{\textbf{Alignment}}}
& \multirow{2}{*}{CLIP Text ($\uparrow$)} & GPT
& 0.127 & 0.319 & -0.133 & -0.000 & 0.014 \\
& & & Gemini
& 0.218 & 0.412 & -0.247 & -0.190 & -0.141 \\
\cmidrule(lr){3-9}
& & \multirow{2}{*}{CLIP Image ($\uparrow$)} & GPT
& \textbf{0.117} & \textbf{0.253} & \textbf{0.561} & \textbf{0.456} & \textbf{0.368} \\
& & & Gemini
& 0.504 & 0.670 & 0.220 & 0.142 & 0.130 \\
\cmidrule(lr){3-9}
& & \multirow{2}{*}{DINO Image ($\uparrow$)} & GPT
& 0.130 & 0.288 & 0.245 & 0.181 & 0.149 \\
& & & Gemini
& 0.298 & 0.479 & 0.311 & 0.364 & 0.279 \\
\cmidrule(lr){3-9}
& & \multirow{2}{*}{Avg} & GPT
& 0.125 & 0.287 & 0.224 & 0.212 & 0.177 \\
\bottomrule
\end{tabular}
\end{table*}

\begin{table*}[t]
\centering
\caption{
Results of Image Preservation using semantic similarity metrics (CLIP Text, CLIP Image, DINO Image) (ONLINE setting, refer to Figure \ref{fig:prompt-mllm-as-a-judge-image-editing, main} for prompt).
Evaluation measures score prediction (MSE, MAE) and ranking quality (Pearson, Spearman, Kendall $\tau$).
Scores are reported for both GPT and Gemini.
}
\label{tab:clip_dino_gpt_gemini_image_preservation_online}
\tiny
\begin{tabular}{@{}c c l c *{5}{c}@{}}
\toprule
& \textbf{Factor} &
\textbf{Metric} &
\textbf{Model} &
MSE $\downarrow$ & MAE $\downarrow$ &
Pearson $\uparrow$ & Spearman $\uparrow$ & Kendall $\tau$ $\uparrow$ \\
\midrule
\multirow{21}{*}{\rotatebox[origin=c]{90}{\textcolor{GoogleGreen}{\textsc{\bfseries Image Preservation}}}}
& \multirow{7}{*}{\textcolor{GoogleGreen}{\textbf{Unchanged Regions}}}
& CLIP Text & GPT & 0.197 & 0.434 & -0.076 & -0.116 & -0.096 \\
& & & Gemini & 0.168 & 0.397 & -0.057 & -0.103 & -0.076 \\
\cmidrule(lr){3-9}
& & CLIP Image & GPT & \textbf{0.166} & \textbf{0.258} & 0.147 & 0.000 & 0.000 \\
& & & Gemini & 0.137 & 0.241 & -0.037 & -0.050 & -0.040 \\
\cmidrule(lr){3-9}
& & DINO Image & GPT & 0.301 & 0.464 & \textbf{-0.275} & \textbf{-0.318} & \textbf{-0.264} \\
& & & Gemini & 0.208 & 0.365 & 0.044 & 0.006 & 0.011 \\
\cmidrule(lr){3-9}
& & Avg & GPT & 0.221 & 0.385 & -0.068 & -0.145 & -0.120 \\
\bottomrule
\end{tabular}
\end{table*}

\begin{table*}[t]
\centering
\caption{
Results of Edit Quality using semantic similarity metrics (CLIP Text, CLIP Image, DINO Image) (ONLINE setting, refer to Figure \ref{fig:prompt-mllm-as-a-judge-image-editing, main} for prompt).
Evaluation measures score prediction (MSE, MAE) and ranking quality (Pearson, Spearman, Kendall $\tau$).
Scores are reported for both GPT and Gemini.
}
\label{tab:clip_dino_gpt_gemini_edit_quality_online}
\tiny
\begin{tabular}{@{}c c l c *{5}{c}@{}}
\toprule
& \textbf{Factor} & \textbf{Metric} & \textbf{Model} &
MSE $\downarrow$ & MAE $\downarrow$ &
Pearson $\uparrow$ & Spearman $\uparrow$ & Kendall $\tau$ $\uparrow$ \\
\midrule
\multirow{42}{*}{\rotatebox[origin=c]{90}{\textcolor{GoogleBlue}{\textsc{\bfseries Edit Quality}}}}

& \multirow{7}{*}{\textcolor{GoogleBlue}{\textbf{Scale Realism}}}
& CLIP Text & GPT & 0.127 & 0.337 & 0.063 & -0.042 & -0.020 \\
& & & Gemini & 0.154 & 0.369 & 0.000 & -0.034 & -0.025 \\
\cmidrule(lr){3-9}
& & CLIP Image & GPT & \textbf{0.069} & \textbf{0.175} & \textbf{0.169} & \textbf{0.333} & \textbf{0.254} \\
& & & Gemini & 0.097 & 0.203 & 0.277 & 0.298 & 0.249 \\
\cmidrule(lr){3-9}
& & DINO Image & GPT & 0.157 & 0.311 & 0.124 & 0.276 & 0.209 \\
& & & Gemini & 0.184 & 0.338 & 0.104 & 0.214 & 0.162 \\
\cmidrule(lr){3-9}
& & Avg & GPT & 0.118 & 0.274 & 0.119 & 0.189 & 0.148 \\
\bottomrule
\end{tabular}
\end{table*}

\begin{table*}[t]
\centering
\caption{
Results of Instruction Fidelity using semantic similarity metrics (CLIP Text, CLIP Image, DINO Image) (ONLINE setting, refer to Figure \ref{fig:prompt-mllm-as-a-judge-image-editing, main} for prompt).
Evaluation measures score prediction (MSE, MAE) and ranking quality (Pearson, Spearman, Kendall $\tau$).
Scores are reported for both GPT and Gemini.
}
\label{tab:clip_dino_gpt_gemini_instruction_fidelity_online}
\tiny
\begin{tabular}{@{}c c l c *{5}{c}@{}}
\toprule
& \textbf{Factor} & \textbf{Metric} & \textbf{Model} &
MSE $\downarrow$ & MAE $\downarrow$ &
Pearson $\uparrow$ & Spearman $\uparrow$ & Kendall $\tau$ $\uparrow$ \\
\midrule
\multirow{21}{*}{\rotatebox[origin=c]{90}{\textcolor{GoogleRed}{\textsc{\bfseries Instruction Fidelity}}}}
& \multirow{7}{*}{\textcolor{GoogleRed}{\textbf{Alignment}}}
& CLIP Text & GPT & 0.152 & 0.376 & \textbf{0.141} & 0.116 & 0.087 \\
& & & Gemini & 0.154 & 0.390 & 0.035 & -0.053 & -0.044 \\
\cmidrule(lr){3-9}
& & CLIP Image & GPT & \textbf{0.075} & \textbf{0.173} & 0.008 & 0.064 & 0.054 \\
& & & Gemini & 0.043 & 0.140 & 0.251 & 0.163 & 0.133 \\
\cmidrule(lr){3-9}
& & DINO Image & GPT & 0.190 & 0.346 & 0.042 & \textbf{0.189} & \textbf{0.146} \\
& & & Gemini & 0.174 & 0.342 & 0.182 & -0.001 & -0.006 \\
\cmidrule(lr){3-9}
& & Avg & GPT & 0.139 & 0.298 & 0.064 & 0.123 & 0.096 \\
\bottomrule
\end{tabular}
\end{table*}

\end{document}